\documentclass[journal]{vgtc}
\onlineid{0}
\vgtccategory{Research}
\newcommand{\chartanno}{\textsc{ChartAnno}}
\newcommand{\chartannotitle}{\chartanno{}}
\newcommand{\bpstart}[1]{\smallskip\noindent\textbf{#1.}}

\title{\chartannotitle{}: Benchmarking Multimodal Large Language Models for Chart Annotation Generation}
\author{
  \authororcid{Zhenghan Chen}{0009-0007-1607-4732},
  \authororcid{Zekai Shao}{0000-0003-2014-5293},
  Lidan Tan,
  Xin Lin,
  Xingchen Zeng,
  Yi Shan,
  Ziyue Lin,\\
  Xiaoliang Fu,
  Xinyuan Liu,
  Yuetong Guo,
  Fen Wang,
  \authororcid{Bongshin Lee}{0000-0002-4217-627X}, and
  \authororcid{Siming Chen}{0000-0002-2690-3588}
}
\authorfooter{
  \item Zhenghan Chen, Zekai Shao, Lidan Tan, Yi Shan, Ziyue Lin, Xiaoliang Fu, Xinyuan Liu, Yuetong Guo, Fen Wang, and Siming Chen are with Fudan University. E-mail: \{chenzh26, zkshao23\}@m.fudan.edu.cn; simingchen@fudan.edu.cn.
  \item Xin Lin is with Sun Yat-sen University. E-mail: linx225@mail2.sysu.edu.cn.
  \item Xingchen Zeng is with The Hong Kong University of Science and Technology (Guangzhou). E-mail: xzeng159@connect.hkust-gz.edu.cn.
  \item Bongshin Lee is with the Department of Computational Science and Engineering and the Yonsei Institute of Digital Health, Yonsei University. E-mail: b.lee@yonsei.ac.kr.
  \item Zhenghan Chen and Zekai Shao are co-first authors, and Siming Chen and Bongshin Lee are co-corresponding authors.
}
\abstract{%
Annotations are essential to communicative visualization, helping explain data, emphasize key findings, and guide attention. While multimodal large language models (MLLMs) offer new opportunities for automatic chart annotation authoring, their capabilities in this task remain underexplored. To address this gap, we introduce \chartanno{}, a comprehensive benchmark for evaluating MLLMs on chart annotation generation. \chartanno{} contains 1,200 real-world charts with paired annotated and unannotated executable code, along with 3,600 annotation instructions spanning three levels of specificity. We also develop a multidimensional evaluation framework combining rule-based and LLM-judged metrics to assess execution, structural compliance, semantic consistency, and design effectiveness. We evaluate 10 representative MLLMs under two primary chart input settings: (1) chart code alone and (2) both code and chart image. Results reveal that proprietary models lead overall, though open-source models narrow the gap. While higher instruction specificity improves annotation quality, inferring abstract communicative intent remains difficult across all models. Providing chart images yields marginal benefit when code is available.
We also examine the effect of chart code through an image-only ablation and analyze the effects of multiple task complexity indicators and instruction-level transitions.
Further analyses characterize common failure modes and validate the reliability of the LLM-based judge.
Experiments with D3 and SVG demonstrate the generalizability of \chartanno{} beyond its primary Python setting.
}
\keywords{Chart annotation, benchmark, evaluation framework, multimodal large language models.}
\teaser{%
    \centering
    \includegraphics[width=\linewidth]{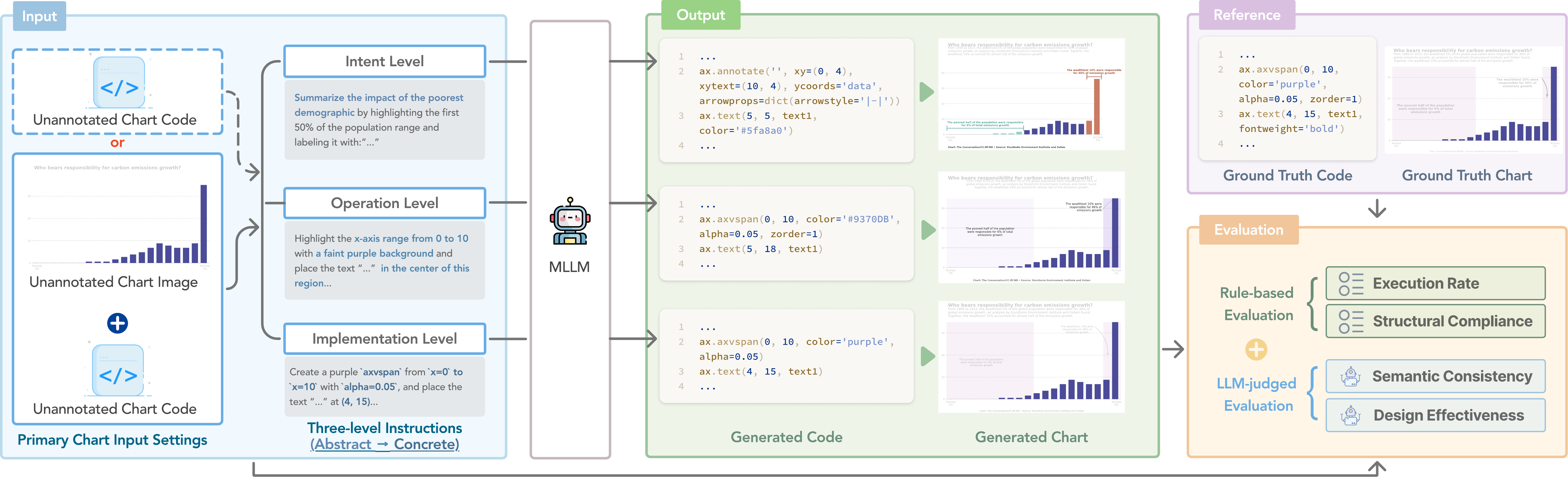}
    \caption{Task design of \chartanno{}.
Inputs are Code or Code + Chart Image, paired with annotation instructions at three levels of specificity: abstract Intent, Operation, and concrete Implementation.
For each chart input--instruction combination, the MLLM generates annotated code and a rendered chart, evaluated with reference to the unannotated code, instruction, and annotated ground truth.}
    \label{fig:task_examples}
}

\usepackage{colortbl}

\usepackage{latexsym}
\usepackage{inconsolata}
\usepackage{booktabs}
\usepackage{pifont}
\usepackage{tabularx}
\usepackage{makecell}
\usepackage[most]{tcolorbox}
\usepackage{multirow}
\usepackage{float}
\usepackage{titletoc}
\usepackage{amsmath}
\usepackage{amssymb}
\usepackage{listings}
\tcbuselibrary{listings,breakable}
\providecommand{\citep}[1]{\cite{#1}}
\providecommand{\citet}[1]{\cite{#1}}
\newcommand{\cmark}{\ding{51}}
\newcommand{\xmark}{\ding{55}}
\newcommand{\pmark}{\(\sim\)}

\newtcblisting{codebox}[2][]{
  enhanced jigsaw,
  breakable,
  pad at break*=0.5mm,
  colback=promptgreen,
  colframe=promptgreenframe,
  coltitle=promptgreentitle,
  colbacktitle=promptgreenframe,
  title={#2},
  fonttitle=\bfseries,
  boxrule=0.6pt,
  arc=2mm,
  left=1mm,
  right=1mm,
  top=1mm,
  bottom=1mm,
  listing only,
  listing options={
    language=Python,
    basicstyle=\ttfamily\scriptsize,
    breaklines=true,
    breakatwhitespace=false,
    columns=fullflexible,
    keepspaces=true,
    showstringspaces=false,
    tabsize=4
  },
  #1
}

\definecolor{chartheat}{HTML}{1FA7C9}
\definecolor{groupgreen}{RGB}{229, 241, 222}
\definecolor{chartblue}{RGB}{36, 102, 160}
\definecolor{annogreen}{RGB}{35, 140, 110}
\definecolor{promptblue}{RGB}{240,247,255}
\definecolor{promptblueframe}{RGB}{218,237,255}
\definecolor{promptbluetitle}{RGB}{61,123,184}

\definecolor{promptgreen}{RGB}{247,250,239}
\definecolor{promptgreenframe}{RGB}{237,244,221}
\definecolor{promptgreentitle}{RGB}{53,132,13}
\definecolor{promptorange}{RGB}{251,249,246}
\definecolor{promptorangeframe}{RGB}{255,242,215}
\definecolor{promptorangetitle}{RGB}{236,161,56}
\definecolor{purple}{RGB}{94,129,198}
\tcbset{
  promptbox/.style={
    enhanced,
    breakable,
    colback=promptblue,
    colframe=promptblueframe,
    coltitle=promptbluetitle,
    colbacktitle=promptblueframe,
    fonttitle=\bfseries,
    boxrule=0.8pt,
    arc=2mm,
    left=1mm,
    right=1mm,
    top=1mm,
    bottom=1mm,
  },
  constraintbox/.style={
    enhanced,
    breakable,
    colback=promptorange,
    colframe=promptorangeframe,
    coltitle=promptorangetitle,
    colbacktitle=promptorangeframe,
    fonttitle=\bfseries,
    boxrule=0.8pt,
    arc=2mm,
    left=1mm,
    right=1mm,
    top=1mm,
    bottom=1mm,
  }
}
\graphicspath{{image/}{./}}

\begin{document}
\firstsection{Introduction}
\maketitle
\renewcommand{\thefootnote}{\arabic{footnote}}
\setcounter{footnote}{0}

Chart annotations are an important component of communicative visualization and data-driven storytelling, helping explain data, emphasize key findings, guide attention, and convey intended messages~\citep{ren2017chartaccent}.
Widely used in real-world visualizations, annotations encompass diverse textual and graphical elements that augment existing charts~\citep{rahman2025annotationdesignspace,rahman2025annotationsurvey}.
Their importance has motivated recent authoring systems and structured frameworks, including mixed-initiative systems for jointly authoring text and charts~\citep{srinivasan2025pluto} and declarative approaches for specifying annotations~\citep{annogram_2025,chartmark_2025}.
Despite these advances, creating effective annotations can still require manual effort.

Recent multimodal large language models (MLLMs) provide new opportunities to further support annotation authoring.
Advances in MLLM capabilities~\citep{openai2026gpt54,google2026gemini31pro,anthropic2026claude46,qwen2026qwen35} have driven progress in chart understanding~\citep{wang2024charxiv,masry2025chartqapro,xie2025infochartqa,mukherjee2026encqa,zhu2025multichartqa}, generation~\citep{viseval_2025,tian2024chartgpt,shi2025chartmimic,zhang2026realchart2code}, and editing~\citep{zhao2025chartedit,charteditbench_2026,li2026chartsarenotimages}.
Chart annotation generation, however, presents a distinct challenge.
Whereas chart editing often begins with a relatively explicit specification of the desired modification, annotation generation may begin with only a high-level communicative intent.
A model must infer what information to convey, associate it with relevant chart elements, select an appropriate annotation design, and generate executable code that integrates the annotation without disrupting the existing chart.

Despite this challenge, systematic evaluation of MLLMs for chart annotation generation remains underexplored.
Existing chart benchmarks largely focus on chart understanding, generation, or editing and rarely consider varying levels of instruction specificity in annotation authoring.
Moreover, evaluating annotation generation requires accounting for multiple aspects of quality, including successful execution, structural compliance, semantic consistency, and design effectiveness.
These considerations motivate a benchmark that captures both instruction specificity and multidimensional annotation quality.

To address this need, we present \chartanno{},\footnote{Code and dataset are available at \url{https://chartanno.github.io/}.} a comprehensive benchmark for evaluating MLLMs on chart annotation generation.
In chart authoring scenarios, annotations are often added to existing code-generated charts, making chart code a natural primary input representation for this task.
Accordingly, as illustrated in Fig.~\ref{fig:task_examples}, \chartanno{} considers two primary chart input settings: (1) chart code alone and (2) chart code and the corresponding chart image.
\chartanno{} contains 1,200 public-facing and scientific charts curated from public chart datasets and arXiv papers, with manually refined chart representations and executable code.
For each chart, we construct instructions at three levels of specificity: (1) Intent describes the communicative goal, (2) Operation specifies annotation actions and target relations, and (3) Implementation further provides concrete rendering parameters.
Together, these yield 3,600 chart--instruction instances and 7,200 tasks across the two primary chart input settings.
We additionally include an image-only setting with 3,600 tasks as an auxiliary ablation to assess model performance when chart code is unavailable.

We further develop an evaluation framework grounded in visualization knowledge.
The framework combines rule-based analysis with LLM-based judgment across four complementary dimensions: \emph{Execution Rate}, \emph{Structural Compliance}, \emph{Semantic Consistency}, and \emph{Design Effectiveness}.
The rule-based component adopts a differential analysis strategy, using the unannotated reference to assess preservation of the original chart and the difference between annotated and unannotated references to identify expected annotation elements for structured matching.
For aspects that are difficult to capture with predefined rules, the LLM-based component evaluates whether the annotations faithfully communicate the intended information and whether their visual design effectively supports that communication.

Our evaluation with \chartanno{} covers 10 recent MLLMs across three instruction levels, two primary chart input settings, and an auxiliary image-only ablation.
Spanning both proprietary and open-source models, we evaluate GPT-5.4~\citep{openai2026gpt54}, Gemini 3.1 Pro Preview~\citep{google2026gemini31pro}, Gemini 3 Flash Preview~\citep{google2025gemini3flash}, Claude Sonnet 4.6~\citep{anthropic2026claude46}, Kimi K2.5~\citep{kimi25_2026}, Gemma 4 31B~\citep{google2026gemma4}, and four Qwen3.5 variants ranging from 9B to 397B parameters~\citep{qwen2026qwen35}.
Our main results show that more detailed annotation instructions generally improve model performance, while Intent-level generation remains challenging.
Providing chart images offers limited additional benefit when code is available.
The Image-only ablation shows substantially lower performance, especially for open-source models and more detailed instructions.
Performance also decreases systematically as chart and annotation complexity increase.
We also validate our evaluation framework, finding strong agreement with human judgments and high consistency across repeated judging runs.
Finally, we extend \chartanno{} beyond Python to two additional representation formats, D3 and SVG, on a 120-chart subset, demonstrating the generalizability of our dataset and evaluation framework.
The main trends observed with Python largely persist with D3 and SVG, while the representations exhibit performance trade-offs across instruction levels, suggesting that code length alone does not fully explain performance.

In summary, our main contributions are as follows:
\begin{itemize}[itemsep=1pt, topsep=2pt, parsep=0pt]
    \item We formulate chart annotation generation as an MLLM benchmark task that spans from high-level communicative intents to concrete implementation specifications.
    \item We construct a high-quality and large-scale benchmark dataset of 1,200 real-world charts with paired unannotated and annotated executable code and 3,600 instructions at three levels of specificity, providing reusable resources for tasks such as chart generation.
    \item We introduce a multidimensional evaluation framework combining rule-based and LLM-judged metrics to assess execution, structural compliance, semantic consistency, and design effectiveness.
    \item We conduct a large-scale evaluation of 10 MLLMs across three instruction levels, two primary chart input settings, and an ablation setting, while validating the evaluation framework through strong human agreement and demonstrating its generalizability across different representation formats.
\end{itemize}

\section{Related Work}
\label{sec:related_work}

\subsection{Chart Annotation}

\textbf{Annotation roles and design.}
Annotations have long been used as additional visual layers that augment charts with textual and graphical elements~\cite{kong2012graphicaloverlays}.
Prior work has systematically characterized how annotations are used in practice.
Studies of real-world visualizations identify diverse annotation forms, targets, and communicative functions, and develop taxonomies and design spaces that capture common annotation practices~\cite{rahman2025annotationdesignspace,rahman2025annotationsurvey}.
These studies show that annotations can combine textual and graphical elements, refer to different components of a visualization, and support different authoring goals.
Other work examines structures formed by user-authored annotations~\cite{annotation_graphs_2017} and the broader communicative functions of textual content in visualizations~\cite{stokes2025textfunctions}.

\bpstart{Annotation authoring and automation}
Building on this design knowledge, visualization research has developed a range of tools and frameworks to support annotation creation.
Contextifier automatically generates contextual annotations for stock visualizations~\cite{hullman2013contextifier}, while ChartAccent supports interactive authoring of manual and data-driven annotations for data-driven storytelling~\cite{ren2017chartaccent}.
Automatic graphical annotation generation has also been explored by deriving strategies from human annotation practices~\cite{shi2024automating}.
More recent approaches extend annotation authoring through mixed-initiative assistance and structured specifications.
Pluto supports coordinated authoring of textual content and charts~\cite{srinivasan2025pluto}, while AnnoGram and ChartMark provide structured, grammar-based representations for specifying chart annotations~\cite{annogram_2025,chartmark_2025}.

\bpstart{Annotations in communication and interpretation}
Annotations play broader roles in visual analysis and communication, where they can externalize insights and connect visual evidence with narrative explanations.
For example, annotations have been integrated into visual dashboards to support analytical exploration~\cite{badam2022facetnotes} and into data-driven storytelling systems to communicate findings alongside visualizations~\cite{islam2024datanarrative}.
Empirical studies further show that the amount, semantic content, and placement of textual annotations can affect readers' preferences, takeaways, and interpretation of visualized information~\cite{stokes2023striking,stokes2024roletext}.
An empirical study evaluates in-situ annotations generated using a single VLM on eight charts, finding improved accuracy for basic factual reading but no significant gains for simple interpretation or response time~\cite{haraguchi2026automating}.

Despite substantial research on annotation design, authoring, and application, systematic evaluation of executable chart annotation generation by MLLMs remains underexplored.
Building on these foundations, our work evaluates annotation generation across different levels of instruction specificity, considering whether generated annotations preserve the underlying chart while remaining structurally compliant, semantically faithful, and visually effective.

\subsection{MLLMs for Chart Generation, Editing, and Annotation}

Recent MLLMs have advanced chart generation and editing, motivating benchmarks of their ability to translate natural language, visual inputs, and existing chart representations into executable visualizations.

\bpstart{Chart generation}
Natural-language-to-visualization research has progressed from NL-to-visualization mapping toward more advanced generation, evaluation, and agentic workflows.
Early benchmarks such as nvBench~\cite{nvbench_2021} provide large-scale natural-language and visualization pairs, while nvBench 2.0 extends this setting with an ambiguity-aware benchmark, multiple valid visualizations, and reasoning paths~\cite{nvbench2_2025}.
ChartGPT constructs an instruction-chart dataset for fine-tuning LLMs to generate charts~\cite{tian2024chartgpt}, while VisEval establishes visualization-specific criteria for assessing generated charts~\cite{viseval_2025}.
Text2Vis further extends text-to-visualization generation toward analytical tasks through an agentic workflow with automated evaluation~\cite{text2vis_2025}.

A related line of work investigates chart-to-code generation, extending chart generation toward executable visualization programs.
Plot2Code introduces image-to-code chart reconstruction~\cite{wu2025plot2code}, while ChartMimic and RealChart2Code extend this setting toward instruction-driven and real-world chart reconstruction scenarios~\cite{shi2025chartmimic,zhang2026realchart2code}.
Chart2Code further considers broader chart authoring scenarios~\cite{tang2026chart2code}.
Recent approaches also investigate iterative refinement and agent-based workflows for chart authoring beyond one-shot generation~\cite{matplotagent_2024,goswami2025plotgen,koh-etal-2025-c2,metal_2025}.
Together, these studies primarily focus on creating or reconstructing charts, whereas \chartanno{} investigates augmenting existing charts with communicative annotations.

\bpstart{Chart editing}
Chart editing focuses on modifying an existing visualization according to user instructions and has recently been studied under increasingly diverse settings.
ChartEdit evaluates code-based chart editing from natural-language instructions~\cite{zhao2025chartedit}, while ChartM$^3$ combines textual instructions with visual indicators for fine-grained multimodal editing~\cite{yang2025chartm3}.
ChartEditBench studies grounded multi-turn editing across successive modifications~\cite{charteditbench_2026}.
ChartEditor instead considers image-based chart editing without access to the original chart code~\cite{chen2026charteditor}, while FigEdit focuses on structured editing of scientific charts~\cite{li2026chartsarenotimages}.
Recent work also explores end-to-end chart editing across both local visual changes and more global transformations~\cite{li2026charte3}.

\chartanno{} complements existing chart generation and chart editing benchmarks by systematically evaluating executable chart annotation generation.
Beyond execution, it evaluates whether generated annotations preserve the underlying chart and are structurally compliant, semantically faithful, and visually effective.
It further covers both public-facing and scientific charts.
Some chart editing tasks partially overlap with annotation generation because they involve adding textual or graphical elements to existing visualizations.
However, existing editing benchmarks generally assume that desired modifications are explicitly specified through editing instructions.
In contrast, \chartanno{} considers instructions ranging from abstract communicative intent to concrete implementation, requiring models to infer appropriate annotations rather than simply apply predefined edits.

\bpstart{Chart annotation}
Chart annotation generation remains largely unexplored in published benchmarks.
During the preparation of this work, we became aware of a concurrent preprint, AnnoBench~\cite{rahatzaman2026annobench}, which also studies chart annotation generation, but in a different setting.
AnnoBench contains 342 charts across six visualization representations, two instruction levels, and multiple chart-description conditions. It uses reference-free LLM judging as its automated evaluation and reports experiments on sampled subsets, with inconsistent alignment between LLM and human judgments across settings.
Among them, the annotation tasks associated with 58 professional charts are derived from the original real-world charts, whereas those for the remaining 284 Vega/Vega-Lite charts are constructed through an LLM-assisted pipeline rather than derived from existing real-world annotations.
In comparison, \chartanno{} contains 1,200 real-world charts with paired executable references, three instruction levels, two primary chart input settings (Code and Code + Image), and an Image-only ablation, enabling large-scale evaluation across the full benchmark with rule-based and human-aligned LLM-judged metrics in Python, with further extensions to D3 and SVG.
We examined whether a controlled comparison could be conducted using AnnoBench's 58 professional charts, whose associated annotation tasks are closest to our setting.
However, only 28 samples support a matched reference-based comparison at the Operation and Implementation levels, which we consider insufficient for a reliable benchmark-level comparison.

Tab.~\ref{tab:benchmark_comparison} summarizes the differences between \chartanno{} and representative benchmarks. The Supplementary Material further discusses differences from chart editing tasks and provides cross-benchmark comparison.

\begin{table}[t]
\caption{Comparison of \chartanno{} with representative visualization generation, chart editing, and annotation benchmarks.
\pmark{} indicates partial overlap or coverage.
Diverse Sources denotes coverage of both public-facing and scientific charts.
Real-world denotes charts collected from existing real-world visualizations rather than charts constructed within the benchmark.
Hybrid Eval. denotes evaluation combining rule-based or programmatic metrics with model-based judgment.
Vis.-grounded Eval. denotes criteria derived from visualization domain knowledge rather than generic execution, similarity, or model-based measures.}
\centering
\small
\setlength{\tabcolsep}{4pt}
\resizebox{\columnwidth}{!}{
\begin{tabular}{lccccc}
\toprule
\textbf{Benchmark}
& \textbf{Anno. Gen.}
& \textbf{Diverse Sources}
& \textbf{Real-world}
& \textbf{Hybrid Eval.}
& \textbf{Vis.-grounded Eval.} \\
\midrule
VisEval         & \xmark & \xmark & \xmark & \cmark & \cmark \\
ChartMimic      & \xmark & \xmark & \cmark & \cmark & \xmark \\
RealChart2Code  & \xmark & \xmark & \xmark & \cmark & \xmark \\
\midrule
ChartEdit       & \pmark & \xmark & \cmark & \xmark & \xmark \\
ChartEditBench  & \pmark & \xmark & \xmark & \cmark & \xmark \\
FigEdit         & \pmark & \xmark & \xmark & \xmark & \xmark \\
\midrule
AnnoBench       & \cmark & \xmark & \pmark & \xmark & \cmark \\
\midrule
\textbf{\chartanno{}}
& \cmark & \cmark & \cmark & \cmark & \cmark \\
\bottomrule
\end{tabular}
}
\label{tab:benchmark_comparison}
\vspace{-10pt}
\end{table}

\section{\chartanno{}}
\label{sec:chartanno}
\begin{figure*}[t]
    \centering
    \includegraphics[width=1\textwidth]{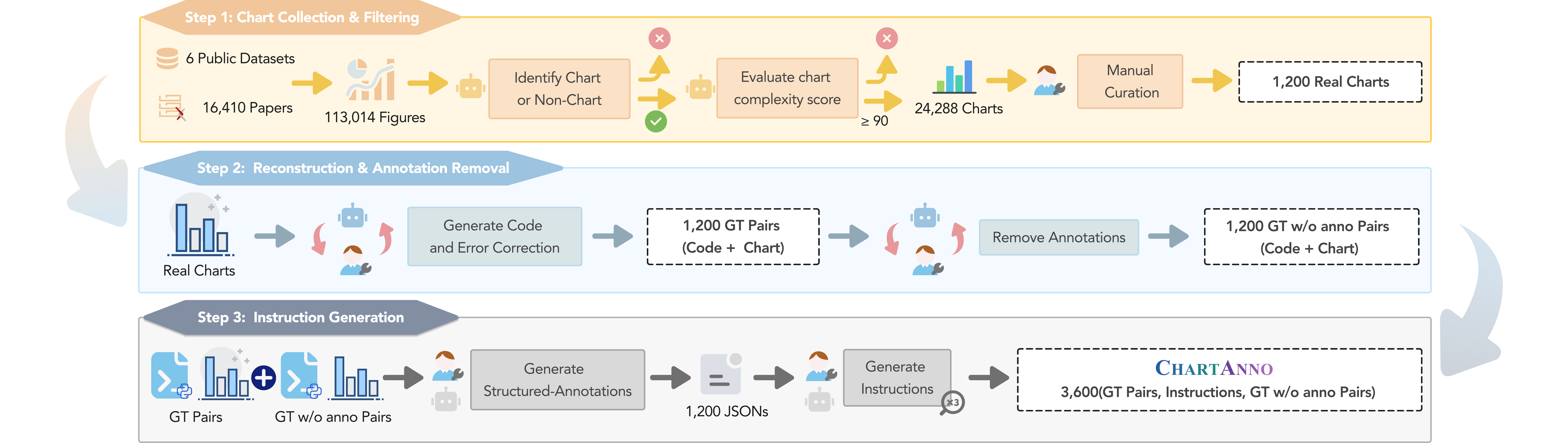}
    \caption{Construction pipeline of \chartanno{}, including collection and filtering, reconstruction and annotation removal, and instruction generation.}
    \label{fig:dataset_construction}
\end{figure*}

We develop \chartanno{} to systematically evaluate MLLMs for chart annotation generation.
We describe its task formulation, benchmark construction, and multidimensional evaluation framework.
Detailed construction procedures, metric implementation, scoring rubrics, and evaluation prompt are provided in the Supplementary Material.

\subsection{Task Formulation}
\label{sec:task_definition}

\textbf{Executable Visualization Environment.}
We instantiate \chartanno{} in a controlled static-chart environment using Python-based executable code.
similar executable settings have been adopted by recent chart generation and editing benchmarks, including ChartMimic~\cite{shi2025chartmimic}, ChartEdit~\cite{zhao2025chartedit}, VisEval~\cite{viseval_2025}, MatPlotAgent~\cite{matplotagent_2024}, and Text2Vis~\cite{text2vis_2025}, enabling reproducible rendering, direct code execution, and programmatic inspection of chart elements.
We further construct D3 and SVG representations for a subset of \chartanno{} and evaluate them in Sec.~\ref{sec:representation_generalization} to validate generalizability beyond the Python setting.

\bpstart{Chart Input Settings and Task Objective}
In chart authoring workflows, authors often refine an existing chart by editing its underlying code to add annotations that communicate a specific message more clearly.
Accordingly, \chartanno{} primarily considers chart code as the input representation and examines whether the corresponding chart image provides complementary visual information.
As illustrated in Fig.~\ref{fig:task_examples}, we consider two primary chart input settings:
(1) chart code alone and
(2) chart code together with the corresponding chart image.
We additionally include an image-only setting as an auxiliary ablation to assess model performance when chart code is unavailable.

\bpstart{Instruction Specificity}
Prior work on chart annotation distinguishes communicative purposes, annotation forms and targets, and concrete visual properties~\cite{ren2017chartaccent,chartmark_2025}.
These distinctions motivate our three instruction levels of increasing specificity.
(1) \textit{\textbf{Intent-level instructions}} describe what should be communicated, including necessary information not inferable from the chart, while leaving the annotation strategy to the model.
(2) \textit{\textbf{Operation-level instructions}} additionally specify annotation operations, target chart elements, and placement relationships, while leaving concrete rendering parameters unspecified.
(3) \textit{\textbf{Implementation-level instructions}} further provide rendering parameters that can be directly translated into executable annotation code.
Together, these levels progressively reduce the annotation design decisions left to the model.

\subsection{Benchmark Construction}
\label{sec:benchmark_construction}

We construct \chartanno{} through a three-stage pipeline, as shown in Fig.~\ref{fig:dataset_construction}.
The pipeline consists of chart collection and filtering, reconstruction and annotation removal, and instruction generation.

\bpstart{Chart Collection and Filtering}
To ensure diversity in chart types, annotation forms, and application contexts, we construct a large candidate pool from existing chart benchmarks, including ChartQAPro~\cite{masry2025chartqapro}, CharXiv~\cite{wang2024charxiv}, ChartMimic~\cite{shi2025chartmimic}, and MatPlotBench~\cite{matplotagent_2024}, as well as visualization studies focused on chart annotations and textual content~\cite{rahman2025annotationdesignspace,stokes2025textfunctions}.
We further supplement these sources with figures from CC BY 4.0 arXiv papers~\cite{arxiv} released between February 2025 and February 2026 and accepted at leading peer-reviewed venues.
We use Semantic Scholar~\cite{kinney2023semantic} to retrieve publication metadata and MinerU~\cite{wang2024mineru} to parse the PDFs and extract figures, following the pipeline used in ChartFI~\cite{wang2026chartfibenchmarkingfaithfulnessinsightfulness}.
Together, these sources yield 113,014 candidate figures.

\begin{figure}[H]
\centering
\includegraphics[width=\columnwidth]{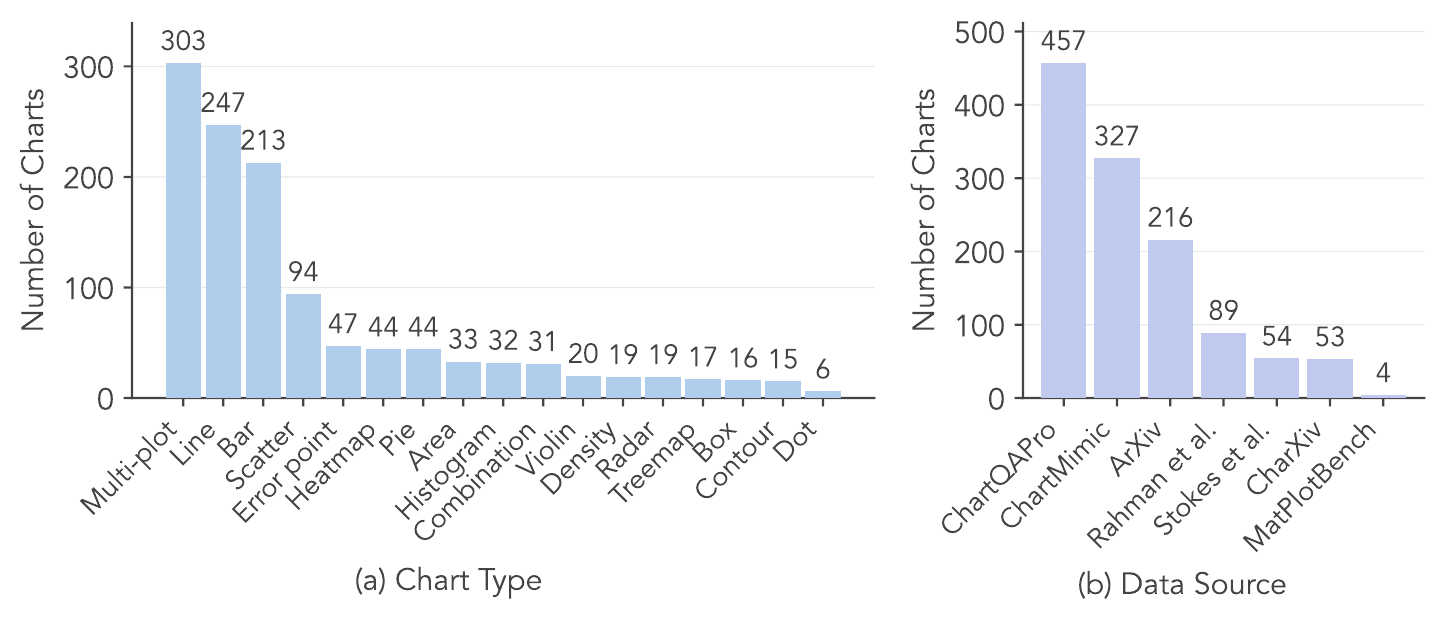}
\vspace{-18pt}
\caption{Distribution of \chartanno{} by chart type and data source.}
\label{fig:chart_distribution}
\vspace{-5pt}
\end{figure}

We then apply a two-stage MLLM-based screening procedure.
The first stage identifies and removes non-chart figures, while the second scores the remaining charts based on chart completeness, information richness, and visual complexity.
Charts with scores below 90 are removed, leaving 24,288 candidates for manual review.
Three authors further review the screened candidates for readability, information sufficiency, diversity in chart types and visual structures, and redundancy, while removing low-quality, incomplete, ambiguous, or overly simple charts and checking for obvious sensitive or inappropriate content.
This process results in 1,200 annotated charts, including 653 public-facing and 547 scientific charts. As shown in Fig.~\ref{fig:chart_distribution}, the dataset covers 17 chart types from seven data sources.

\bpstart{Chart Reconstruction and Annotation Removal}
We reconstruct each selected chart in Python, obtaining an annotated ground truth (GT) pair (Code + Chart) consisting of executable code and its rendered image.
We use LLMs (GPT-5.2~\cite{openai2025gpt52} and Gemini 3 Pro~\cite{google2026gemini3pro}) to generate initial reconstruction code for each selected chart.
We evaluate reconstruction quality using a 100-point rubric adapted from ChartMimic~\cite{shi2025chartmimic}, which assesses consistency with the source image in chart type, layout, text content, data, style, and clarity.
The initial reconstructions achieve an average score of 92.34 on charts from sources other than ChartMimic.
Based on this assessment, five authors manually refine the reconstructed charts for consistency with the source images and further correct three types of source-chart issues:
(1) \textit{ambiguous annotation intent}, where the intended target, direction, or connection of an annotation is unclear or inconsistent with the chart content;
(2) \textit{factual inconsistencies}, where annotation text conflicts with the underlying data or derived statistics, such as incorrect percentage changes or summary values; and
(3) \textit{visual presentation issues}, where annotations are difficult to interpret because of truncated text, occluded labels, or unclear placement.
To construct the corresponding unannotated references, we categorize annotation elements by both annotation type and information source following prior annotation design-space research~\cite{rahman2025annotationdesignspace}.
These labels guide LLMs in identifying and removing annotation-specific code, and five authors manually review and refine the outputs to produce the final unannotated GT pairs (Code + Chart), which preserve the underlying chart content and visual structure.

\begin{figure}[t]
    \centering
    \includegraphics[width=\columnwidth]{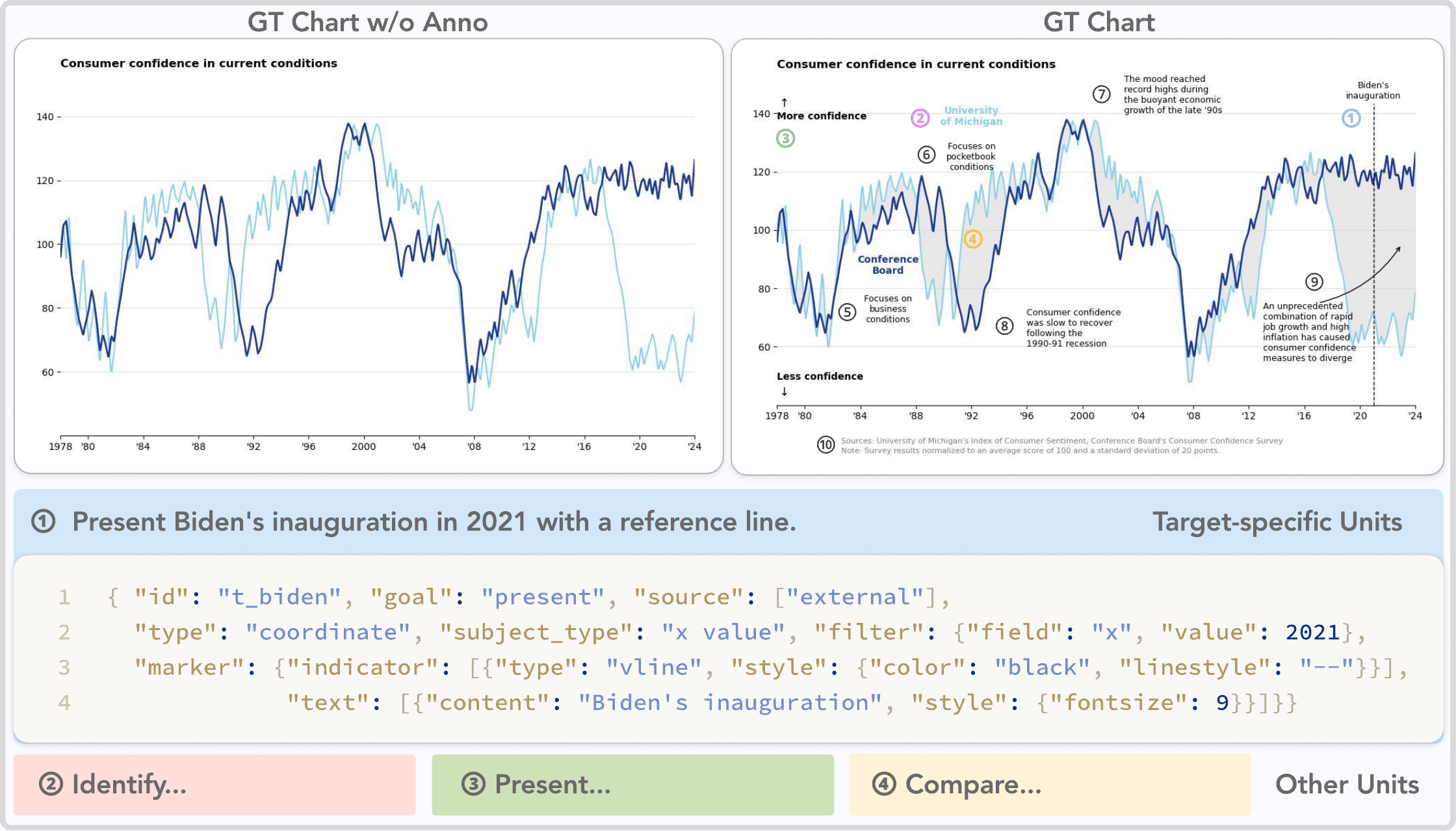}
    \caption{Example of the structured annotation representation with target-specific units. One unit is expanded in JSON.}
    \label{fig:json-simplified}
    \vspace{-10pt}
\end{figure}

\bpstart{Instruction Generation}
Public chart datasets generally do not provide the annotation rationales or original user requests that motivated individual annotations.
Rather than attempting to recover their exact original wording, we aim to capture the underlying communicative intent and task requirements expressed by the annotations.
Guided by prior work on chart annotation design spaces~\cite{rahman2025annotationdesignspace} and annotation grammars~\cite{chartmark_2025,annogram_2025}, we represent the annotations in each chart using a structured schema for instruction construction, recording their communication goal (identify, compare, summarize, or present), target, source, content, annotation type, and markers~\cite{rahman2025annotationdesignspace}.
Annotations associated with different targets are represented as separate target-specific units, as shown in Fig.~\ref{fig:json-simplified}.
Based on these structured representations, we construct three instruction levels that progressively describe the annotation from communicative intent to annotation operations and concrete implementation details, as formally defined in Sec.~\ref{sec:task_definition}.

We first prompt Gemini 3 Pro~\cite{google2026gemini3pro} to generate an initial structured annotation representation for each chart from the annotated and unannotated GT pairs.
Three authors then review and refine these representations assisted by GPT-5.2~\cite{openai2025gpt52}.
Based on the representations, we follow the same procedure to generate the three-level instructions, with Gemini 3 Pro producing the initial drafts and GPT-5.2 assisting in their refinement, yielding 3,600 instructions in total.

\begin{figure}[t]
\centering
\includegraphics[width=\columnwidth]{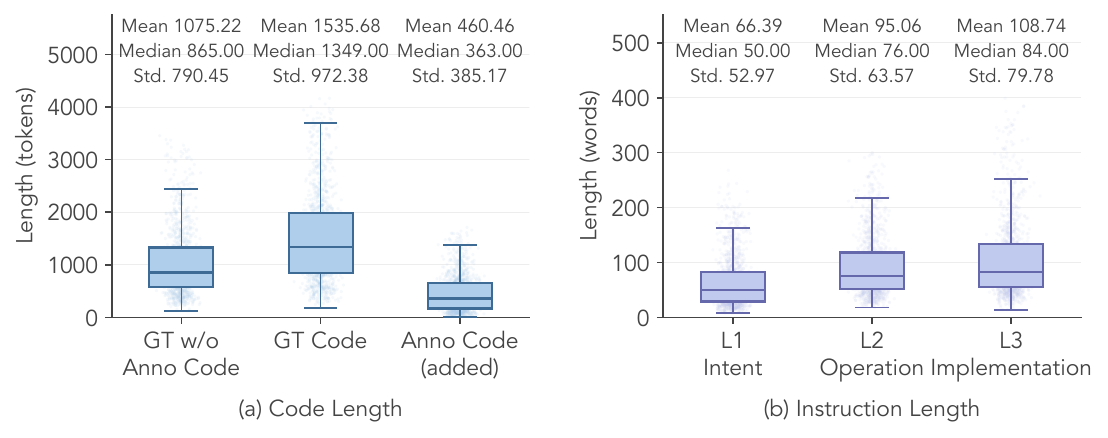}
\caption{Code and instruction length distributions in \chartanno{}.}
\label{fig:dataset_statistics}
\vspace{-10pt}
\end{figure}

\bpstart{Dataset Statistics}
\chartanno{} contains 1,200 annotated GT pairs, 1,200 corresponding unannotated GT pairs, and 3,600 instructions.
Across the GT charts, we identify 25,772 annotation elements, averaging 21.48 elements and 2.17 annotation types per chart.
Fig.~\ref{fig:dataset_statistics} summarizes the code and instruction length distributions.
Using the Llama 2 tokenizer~\cite{touvron2023llama2}, annotated GT code has a median length of 1,349 tokens, compared with 865 tokens for unannotated GT code, with annotations adding a median of 363 tokens.
Instruction length increases with specificity, with median lengths of 50, 76, and 84 words for the Intent, Operation, and Implementation levels, respectively, with the Intent-level median comparable to user instructions reported in $C^2$~\cite{koh-etal-2025-c2}.
The annotation extraction procedure is detailed in Sec.~\ref{sec:rule_based_metrics}.

\subsection{Evaluation Framework}
\label{sec:evaluation_framework}

Evaluating executable chart annotation generation requires both directly verifiable constraints and higher-level judgments of semantic and design quality.
Low-level automatic metrics provide reproducible checks of properties such as executability and structural correctness, but cannot fully capture semantic correctness or overall design quality; visually similar chart outputs may still contain incorrect or unintended transformations~\cite{li2026chartsarenotimages}.
Conversely, model- or human-based judgments can assess semantic alignment and visual quality more holistically, but may introduce evaluator variability and subjective bias~\cite{goswami-etal-2025-charteval}.
Existing visualization benchmarks therefore combine complementary low- and high-level evaluation signals~\cite{viseval_2025,shi2025chartmimic,zhao2025chartedit}.
Following this strategy, we use rule-based metrics for properties that can be explicitly verified from rendered outputs and LLM judgment for qualities that require contextual interpretation of annotation meaning and visual design.

Accordingly, we evaluate annotations along four dimensions: \emph{Execution Rate} and \emph{Structural Compliance} (rule-based metrics), and \emph{Semantic Consistency} and \emph{Design Effectiveness} (LLM-judged metrics).

\subsubsection{Rule-based Metrics}
\label{sec:rule_based_metrics}

Rule-based evaluation focuses on properties that can be explicitly verified from executable and rendered chart outputs.
Prior chart-generation benchmarks assess executability and rendered properties such as text, layout, chart type, and color~\cite{viseval_2025,shi2025chartmimic}, while chart-editing benchmarks further consider preservation of unmodified content and correctness of graphical and textual changes~\cite{zhao2025chartedit,charteditbench_2026,chen2026charteditor,li2026chartsarenotimages}.
Building on these evaluations, we assess execution validity, preservation of the underlying chart, required annotation elements, and their visual properties, through annotation-specific graphical-element comparisons between paired annotated and unannotated GT charts.

\bpstart{Execution Rate}
To measure whether the generated code executes and renders successfully, we assign 1 to instances that produce a chart without runtime errors and 0 otherwise.
We report the proportion of successful instances, with failed executions receiving zero for all downstream quality metrics.
We also summarize common execution failure modes to characterize model errors in executable chart generation.

\begin{figure}[!htbp]
    \centering
    \includegraphics[width=\columnwidth]{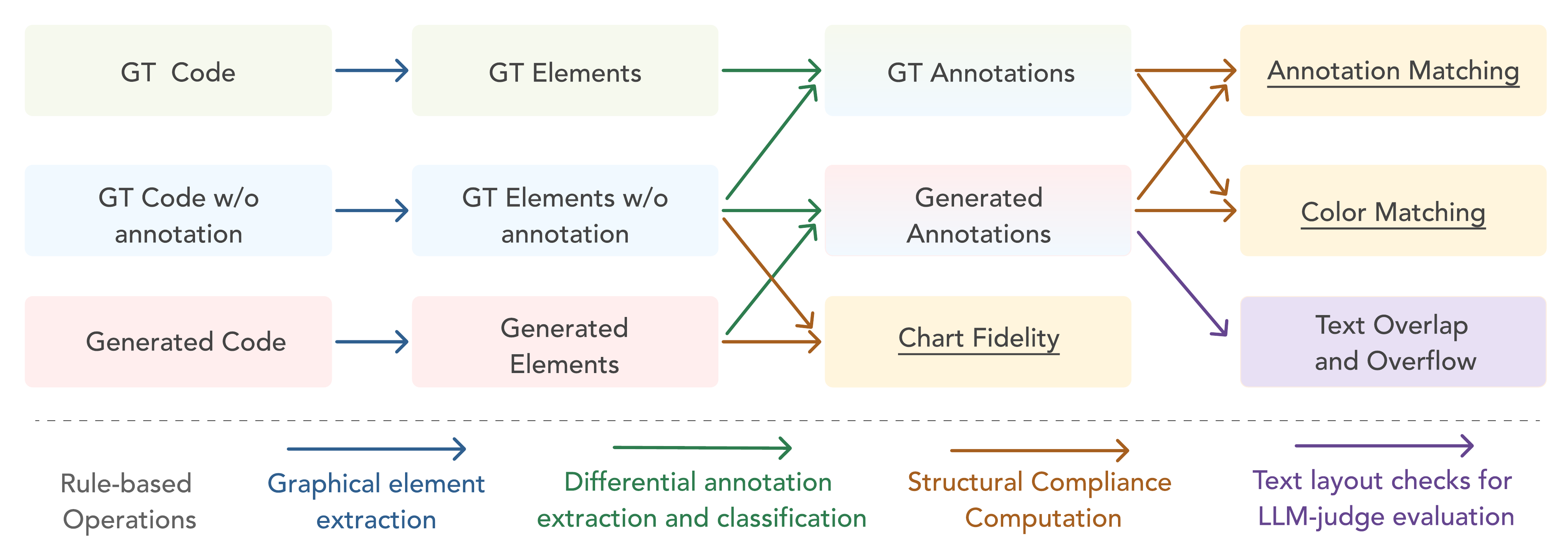}
    \caption{Workflow for computing \emph{Structural Compliance} in \chartanno{}.}
    \label{fig:struc_workflow}
    \vspace{-10pt}
\end{figure}

\bpstart{Structural Compliance}
For successfully rendered outputs, \emph{Structural Compliance} evaluates whether annotation generation preserves the underlying chart and correctly realizes the specified annotation elements and their visual properties.
For Operation- and Implementation-level instructions, we define \emph{Structural Compliance} as the average of \emph{Chart Fidelity}, \emph{Annotation Matching}, and \emph{Color Matching}.
For Intent-level instructions, it consists only of \emph{Chart Fidelity}, because the same communicative intent may be validly realized through different annotation structures and color designs.
We operationalize these requirements by comparing rendered chart elements to assess chart preservation, annotation matching, and color matching.
Fig.~\ref{fig:struc_workflow} summarizes the workflow based on generated code and annotated and unannotated GT code, with the computation procedure and individual metrics described below.

\underline{\textit{Graphical element extraction.}}
Our evaluation operates on the graphical object representations underlying the rendered charts rather than source code, since equivalent visualizations may be implemented differently and expressed in different code representations; evaluating their graphical elements therefore reduces dependence on the underlying representation and supports evaluation across visualization languages.
We first programmatically traverse the underlying graphical object hierarchy of the GT chart, GT chart w/o annotation, and generated chart to extract visible graphical and textual elements and their properties.

\textit{\textbf{Chart Fidelity}} evaluates whether annotation generation preserves the original chart structure and data representation, which prior chart-editing work commonly assesses through human or model judgment~\cite{zhao2025chartedit,li2026chartsarenotimages}.
We instead operationalize preservation as a rule-based comparison between the generated chart and the GT chart w/o annotation, considering figure aspect ratio, axes layout and aspect settings, and the preservation of data-carrying marks and their encoded values.
The metric is binary:
\begin{equation}
\mathrm{ChartFidelity}=
\begin{cases}
1, & \text{if all protected chart properties are preserved},\\
0, & \text{otherwise}.
\end{cases}
\label{eq}
\end{equation}

\underline{\textit{Differential annotation extraction and classification.}}
To identify newly introduced annotations, we perform element-level differencing using the GT chart w/o annotation as a common reference.
We separately match elements in the GT chart and generated chart against those in the GT chart w/o annotation; unmatched elements are treated as newly introduced annotation candidates.
Element matching is based on element type, textual content, spatial position and extent, shape, and data-related characteristics, with criteria adapted to different graphical elements.
We classify the resulting candidates using heuristic rules into seven categories: enclosure, connector, text, glyph, color, indicator, and geometric annotations, following prior annotation taxonomies~\cite{rahman2025annotationdesignspace,rahman2025annotationsurvey}.

\textit{\textbf{Annotation Matching}} evaluates the correspondence between annotations in the generated and GT charts.
We perform this comparison separately within each of the seven annotation categories.
Unlike text annotations, which provide textual content as a relatively stable basis for correspondence, non-text annotations generally lack a unique element-level identity.
The same annotation function may be realized with different graphical primitives, sizes, or placements (e.g., an enclosure highlighting the same chart region may vary in its exact extent and position while remaining semantically equivalent).
We therefore adopt a relaxed matching criterion for non-text annotations, allowing variations in their exact graphical realization and assessing correspondence at the category level, where the intersection and union are defined as the minimum and maximum of the GT and generated annotation counts, respectively.
For text annotations, content provides a stronger basis for correspondence, but equivalent text may still differ in representation (e.g., line wrapping, text-block splitting or merging, capitalization, or whitespace).
We therefore normalize textual content and perform multiset matching, so that correspondence is determined primarily by text content rather than its exact rendered representation.
Let $I_k$ and $U_k$ denote the resulting intersection and union counts for category $k$.
\emph{Annotation Matching} is computed using a Jaccard-style coefficient:
\begin{equation}
\mathrm{AnnotationMatching}
=
\frac{\sum_k I_k}{\sum_k U_k}.
\label{eq:annotation_matching}
\end{equation}

\textit{\textbf{Color Matching}} evaluates how closely the colors of generated annotations match those of the GT annotations.
We extract a primary color from each annotation element and perform matching separately within each annotation category, avoiding matches between colors that serve different annotation functions.
For each GT--generated color pair that can be represented numerically, we convert both colors to the CIELAB color space and compute their perceptual color difference $\Delta E_{00}$ using CIEDE2000~\cite{luo2001ciede2000}, while colors without a numeric representation are matched exactly.
We convert the resulting difference into a normalized similarity:
\begin{equation}
s(c_i,c_j)
=
\max\left(
0,\,
1-\frac{\Delta E_{00}(c_i,c_j)}{100}
\right).
\label{eq:color_similarity}
\end{equation}
Within each annotation category, we construct a pairwise color-similarity matrix and apply the Hungarian algorithm~\cite{kuhn1955hungarian} to obtain the maximum-similarity one-to-one assignment between GT and generated colors.
Let $M$ denote the summed similarity of all matched pairs, and $N_{\mathrm{GT}}$ and $N_{\mathrm{gen}}$ denote the total numbers of GT and generated annotation colors, respectively.
Precision, recall, and \emph{Color Matching} are then computed as
\begin{equation}
P=\frac{M}{N_{\mathrm{gen}}},
\qquad
R=\frac{M}{N_{\mathrm{GT}}},
\qquad
\mathrm{ColorMatching}=\frac{2PR}{P+R}.
\label{eq:color_matching}
\end{equation}

\underline{\textit{Text layout checks for LLM-judged evaluation.}}
VisEval shows that fine-grained layout issues such as text overlap and overflow are difficult to assess reliably using GPT-4V alone and therefore supplements model-based readability assessment with explicit layout checks~\cite{viseval_2025}.
In chart annotation, however, overlap between non-text graphical elements is not necessarily undesirable, as annotations such as enclosures, highlights, and connectors may intentionally overlap existing chart content.
We therefore restrict these geometric checks to newly added annotation text, computing text-overlap and off-canvas statistics as quantitative evidence for the subsequent LLM-based assessment of visual clarity rather than as standalone Structural Compliance scores.

\subsubsection{LLM-Judged Metrics}
\label{sec:llm_judged_metrics}

Rule-based metrics capture executable and structural properties, but cannot assess whether annotations communicate the intended meaning or support effective visual design.
We therefore develop an LLM-based evaluation rubric grounded in visualization knowledge on chart annotations.
The judge evaluates the added annotations with reference to the instruction and annotated GT chart across five complementary aspects, aggregated into \emph{Semantic Consistency} and \emph{Design Effectiveness}.

\bpstart{Semantic Consistency}
Chart annotation semantics involve multiple aspects, including association with relevant chart targets~\cite{ren2017chartaccent}, contextual relevance~\cite{hullman2013contextifier}, and diverse communicative purposes such as identifying, comparing, summarizing, and presenting information~\cite{rahman2025annotationdesignspace}.
Beyond semantic correctness, annotations should also communicate their intended meaning clearly, as ambiguous annotation content can affect viewers' interpretations~\cite{stokes2024roletext,rahman2025annotationsurvey}.
We define \emph{Semantic Consistency} as the average of \emph{Semantic Faithfulness} and \emph{Semantic Clarity}.

\textit{\textbf{Semantic Faithfulness}} evaluates whether the generated annotations correctly realize the meaning specified by the instruction.
The judge considers whether annotations refer to the intended chart targets and accurately express required text, trends, values, relations, conclusions, and visual encodings, while checking for omissions, factual deviations, misreferences, and unsupported additions.
A semantically correct annotation may still be difficult to interpret if its referent or relation to the chart is ambiguous.
\textit{\textbf{Semantic Clarity}} therefore evaluates whether annotation content forms a clear relation with the corresponding visual objects, such that the intended referent and meaning can be identified without competing interpretations or unnecessary inference.

\bpstart{Design Effectiveness}
Beyond semantic quality, annotations should integrate with the underlying visualization and support its visual communication.
Annotation research emphasizes readable placement, effective organization, and visual guidance without excessive clutter~\cite{rahman2025annotationdesignspace,rahman2025annotationsurvey}.
We define \emph{Design Effectiveness} as the average of \emph{Visual Clarity}, \emph{Annotation Organization Quality}, and \emph{Attention Guidance}.

\textit{\textbf{Visual Clarity}} evaluates whether annotations remain readable without interfering with important chart content.
The judge considers overlap, clipping, crowding, off-canvas placement, and occlusion of titles, axes, legends, labels, and data-carrying marks.
The LLM judge receives both the rendered chart and quantitative rule-based evidence, including text-overlap and off-canvas statistics, and jointly considers them when assessing visual clarity.
Annotations may remain readable while still being poorly arranged.
\textit{\textbf{Annotation Organization Quality}} evaluates how effectively annotation elements are placed, grouped, spaced, attached to their targets, and coordinated through color to form a coherent composition integrated with the underlying chart.
It also considers redundancy, complexity, imbalance, and weak coordination among annotation elements.
\textit{\textbf{Attention Guidance}}
evaluates whether annotations make the intended target or target set visually salient, establish a clear focus, and avoid competing or misleading emphasis that increases the effort required to locate the intended information.

Each submetric is scored on an integer scale from 1 to 5, with 0 for failed or missing cases.
The same evaluation criteria and scoring rubrics are applied across all three instruction levels; only the role of the GT chart varies with instruction specificity.
For Intent-level evaluation, the LLM judge is instructed to use the GT chart only to interpret the intended message and chart content, rather than as a reference for annotation designs.
For Operation- and Implementation-level evaluation, the judge additionally references the GT chart when assessing compliance with the specified annotation requirements.

\bpstart{Judge Validation and Selection}
Before the large-scale evaluation, we conduct a pilot study on 90 randomly sampled outputs generated by Gemini 3 Flash Preview and independently rated by three coauthors to validate the rubric and select the LLM judge.
Human ratings show high reliability, with ICCs of 0.869 for \emph{Semantic Consistency}, 0.903 for \emph{Design Effectiveness}, and 0.918 overall.
Among GPT-5.4, Claude Sonnet 4.6, and Gemini 3.1 Pro Preview, GPT-5.4 shows the strongest alignment with aggregated human ratings, as shown in Tab.~\ref{tab:judge_validation}.
Given that GPT-5.4 is also an evaluated model, we additionally examine cross-judge rank consistency and find strong correlations between GPT-5.4 and the other candidate judges.
We therefore use GPT-5.4 as the judge for the large-scale evaluation.

\begin{table}[t]
\caption{Judge validation results. Human-rating reliability is measured by ICC, while judge--human and cross-judge consistency are measured by Spearman's $\rho$. Best results within each comparison group are \textbf{bolded}.}
\centering
\small
\setlength{\tabcolsep}{4pt}
\resizebox{\columnwidth}{!}{
\begin{tabular}{lccc}
\toprule
\textbf{Comparison}
& \textbf{Semantic Consistency}
& \textbf{Design Effectiveness}
& \textbf{Overall} \\
\midrule
Human 1 - Human 2 - Human 3
& 0.869 & 0.903 & 0.918 \\
\midrule
GPT-5.4--Human Avg.
& \textbf{0.812} & \textbf{0.873} & \textbf{0.867} \\
Claude Sonnet 4.6--Human Avg.
& 0.808 & 0.805 & 0.832 \\
Gemini 3.1 Pro--Human Avg.
& 0.714 & 0.732 & 0.748 \\
\midrule
GPT-5.4--Claude Sonnet 4.6
& \textbf{0.730} & \textbf{0.783} & \textbf{0.789} \\
GPT-5.4--Gemini 3.1 Pro
& 0.716 & 0.772 & 0.782 \\
Gemini 3.1 Pro--Claude Sonnet 4.6
& 0.664 & 0.652 & 0.668 \\
\bottomrule
\end{tabular}
}
\vspace{-10pt}
\label{tab:judge_validation}
\end{table}

\section{Experiments}

\begin{table*}[t]
\caption{Main evaluation results of 10 MLLMs under Code Input (left) and Code + Image Input (right) settings across three instruction levels (Intent, Operation, Implementation). Exec., Struct., Sem., and Design denote \emph{Execution Rate}, \emph{Structural Compliance}, \emph{Semantic Consistency}, and \emph{Design Effectiveness}, respectively. \colorbox{gray!12}{Gray} Struct.$^\ast$ columns report \emph{Chart Fidelity} only for Intent-level instructions and are not directly comparable with \emph{Structural Compliance} at the other levels. Within each model category, best results are \textbf{bolded} and second-best results are \underline{underlined}.}
\centering
\small
\setlength{\tabcolsep}{2.5pt}
\resizebox{\textwidth}{!}{
\begin{tabular}{
l|
c>{\columncolor{gray!12}}ccc|
cccc|
cccc||
c>{\columncolor{gray!12}}ccc|
cccc|
cccc
}
\toprule
\textbf{Model}
& \multicolumn{12}{c||}{\textbf{Code Input}}
& \multicolumn{12}{c}{\textbf{Code + Image Input}} \\
\cmidrule(lr){2-13} \cmidrule(lr){14-25}
& \multicolumn{4}{c|}{Intent-level}
& \multicolumn{4}{c|}{Operation-level}
& \multicolumn{4}{c||}{Implementation-level}
& \multicolumn{4}{c|}{Intent-level}
& \multicolumn{4}{c|}{Operation-level}
& \multicolumn{4}{c}{Implementation-level} \\
\cmidrule(lr){2-5} \cmidrule(lr){6-9} \cmidrule(lr){10-13}
\cmidrule(lr){14-17} \cmidrule(lr){18-21} \cmidrule(lr){22-25}
& Exec. & Struct.$^\ast$ & Sem. & Design
& Exec. & Struct. & Sem. & Design
& Exec. & Struct. & Sem. & Design
& Exec. & Struct.$^\ast$ & Sem. & Design
& Exec. & Struct. & Sem. & Design
& Exec. & Struct. & Sem. & Design \\
\midrule
\rowcolor{groupgreen}
\multicolumn{25}{c}{\textit{Proprietary Models}} \\
\midrule
GPT-5.4
& \underline{0.992} & \underline{0.870} & 3.562 & 3.429
& \textbf{0.996} & \underline{0.854} & \underline{3.674} & 3.767
& \underline{0.991} & \underline{0.903} & \underline{4.045} & 4.126
& 0.987 & 0.856 & 3.553 & 3.421
& \underline{0.988} & 0.841 & 3.603 & 3.697
& \underline{0.993} & 0.897 & \underline{4.057} & 4.137 \\
Gemini 3.1 Pro Preview
& \textbf{0.994} & \textbf{0.922} & \textbf{3.688} & \textbf{3.641}
& \underline{0.991} & \textbf{0.873} & \textbf{3.777} & \textbf{3.907}
& \textbf{0.998} & \textbf{0.919} & \textbf{4.128} & \textbf{4.206}
& \underline{0.995} & \textbf{0.917} & \textbf{3.680} & \textbf{3.639}
& \textbf{0.993} & \textbf{0.876} & \textbf{3.780} & \textbf{3.909}
& \textbf{0.997} & \textbf{0.916} & \textbf{4.148} & \textbf{4.230} \\
Gemini 3 Flash Preview
& 0.990 & 0.730 & \underline{3.590} & \underline{3.556}
& 0.985 & 0.816 & 3.612 & \underline{3.801}
& 0.990 & 0.887 & 4.005 & \underline{4.134}
& \textbf{0.996} & 0.696 & \underline{3.600} & \underline{3.596}
& \underline{0.988} & 0.815 & \underline{3.643} & \underline{3.816}
& 0.992 & 0.885 & 4.006 & \underline{4.144} \\
Claude Sonnet 4.6
& 0.975 & 0.864 & 3.493 & 3.385
& 0.963 & 0.828 & 3.461 & 3.571
& 0.984 & \underline{0.903} & 3.999 & 4.098
& 0.984 & \underline{0.885} & 3.512 & 3.421
& 0.974 & \underline{0.843} & 3.498 & 3.646
& 0.988 & \underline{0.906} & 4.020 & 4.112 \\
\midrule
\rowcolor{groupgreen}
\multicolumn{25}{c}{\textit{Open-Source Models}} \\
\midrule
Kimi K2.5
& \textbf{0.973} & \textbf{0.860} & \textbf{3.315} & \textbf{3.267}
& \textbf{0.964} & \textbf{0.812} & \textbf{3.307} & \textbf{3.422}
& \textbf{0.969} & \textbf{0.877} & \textbf{3.869} & \textbf{3.959}
& \textbf{0.983} & \textbf{0.877} & \textbf{3.310} & \textbf{3.280}
& \textbf{0.965} & \textbf{0.818} & \textbf{3.333} & \textbf{3.455}
& \textbf{0.972} & \textbf{0.878} & \textbf{3.878} & \textbf{3.961} \\
Gemma 4 31B
& \underline{0.948} & 0.806 & \underline{3.200} & \underline{3.108}
& \underline{0.929} & \underline{0.796} & \underline{3.206} & \underline{3.292}
& 0.922 & 0.831 & \underline{3.624} & \underline{3.737}
& \underline{0.959} & 0.787 & \underline{3.159} & \underline{3.107}
& 0.940 & \underline{0.796} & \underline{3.220} & \underline{3.334}
& 0.938 & \underline{0.840} & \underline{3.681} & \underline{3.777} \\
Qwen3.5-397B-A17B
& 0.933 & 0.802 & 3.082 & 3.012
& 0.905 & 0.752 & 3.000 & 3.086
& \underline{0.948} & \underline{0.841} & 3.617 & 3.736
& 0.953 & \underline{0.832} & 3.035 & 2.988
& \underline{0.945} & 0.780 & 3.051 & 3.201
& \underline{0.945} & 0.838 & 3.656 & 3.770 \\
Qwen3.5-122B-A10B
& 0.919 & 0.800 & 2.928 & 2.833
& 0.909 & 0.757 & 2.886 & 2.964
& 0.914 & 0.818 & 3.442 & 3.561
& 0.929 & 0.808 & 2.843 & 2.822
& 0.918 & 0.768 & 2.885 & 3.012
& 0.921 & 0.819 & 3.458 & 3.599 \\
Qwen3.5-27B
& 0.935 & \underline{0.828} & 2.966 & 2.896
& 0.896 & 0.750 & 2.830 & 2.934
& 0.904 & 0.813 & 3.432 & 3.552
& 0.939 & 0.828 & 2.959 & 2.900
& 0.894 & 0.746 & 2.837 & 2.960
& 0.914 & 0.821 & 3.440 & 3.550 \\
Qwen3.5-9B
& 0.828 & 0.676 & 2.311 & 2.287
& 0.764 & 0.601 & 2.158 & 2.263
& 0.828 & 0.719 & 2.865 & 3.017
& 0.865 & 0.732 & 2.285 & 2.330
& 0.810 & 0.639 & 2.238 & 2.366
& 0.832 & 0.715 & 2.841 & 2.991 \\
\bottomrule
\end{tabular}
}
\label{tab:main_results}
\end{table*}

\subsection{Experimental Setup}
\label{sec:experimental_setup}
We use the proposed evaluation framework (Sec.~\ref{sec:evaluation_framework}) to evaluate various models with prompt protocols, as detailed below.

\bpstart{Models}
We evaluate 10 MLLMs spanning proprietary and open-source models.
The proprietary models include GPT-5.4~\cite{openai2026gpt54}, Gemini 3.1 Pro Preview~\cite{google2026gemini31pro}, Gemini 3 Flash Preview~\cite{google2025gemini3flash}, and Claude Sonnet 4.6~\cite{anthropic2026claude46}.
The open-source models include Kimi K2.5~\cite{kimi25_2026}, Gemma 4 31B~\cite{google2026gemma4}, and four Qwen3.5 variants (9B, 27B, 122B-A10B, and 397B-A17B)~\cite{qwen2026qwen35}.
The proprietary models and Kimi K2.5 are accessed through APIs, while Gemma 4 31B and the Qwen3.5 models are locally deployed in BF16 precision on eight NVIDIA H100 80GB GPUs.
We use deterministic decoding whenever supported, setting temperature to 0 and top-$p$ to 1 where available.

\bpstart{Prompts}
Following the chart input settings and instruction levels defined in Sec.~\ref{sec:task_definition}, we use consistent prompt templates across all evaluated models.
Under Code Input, models receive the chart code together with an Intent-, Operation-, or Implementation-level instruction and are asked to return executable Python code while preserving the content and style of the unannotated chart.
For Code + Image Input, the chart image is additionally provided.
For the Image-only ablation, we remove the chart code while retaining the chart image and instruction.

\subsection{Results}
\label{sec:main_results}

Tab.~\ref{tab:main_results} summarizes the performance of 10 MLLMs on \chartanno{}.
We first highlight the main findings and describe the effects of chart code, task complexity, and instruction specificity.
We then report error analyses and assess the validity of the LLM-based judge.

\subsubsection{Main Findings}

\textbf{Gemini 3.1 Pro Preview leads proprietary models, while Kimi K2.5 leads open-source models.}
Across three instruction levels, two chart input settings, and four metrics, Gemini 3.1 Pro Preview ranks first in 22 out of 24 comparisons within the proprietary group, with the only exceptions on \emph{Execution Rate}.
Among open-source models, Kimi K2.5 ranks first in all 24 comparisons.
This shows that the leading models are consistently strong across execution, structure, semantics, and design, rather than excelling on a single metric.

\bpstart{A gap remains between proprietary and open-source models, but large open-source models narrow it}
Proprietary models remain stronger overall, especially on \emph{Semantic Consistency} and \emph{Design Effectiveness}.
However, Kimi K2.5 approaches proprietary models in some settings.
For example, under Implementation-level Code Input, Kimi K2.5 reaches 3.869 in \emph{Semantic Consistency} and 3.959 in \emph{Design Effectiveness}, approaching Claude Sonnet 4.6's 3.999 and 4.098.
At the Intent level, Gemini 3 Flash Preview shows relatively low \emph{Structural Compliance}, falling below many evaluated open-source models.
This suggests that strong semantic and design scores do not necessarily imply better preservation of the base chart, as analyzed in Sec.~\ref{sec:error_analysis}.

\bpstart{More detailed instructions improve annotation quality, with Intent-level generation being the most challenging}
Performance generally improves as instructions become more detailed, especially in \emph{Semantic Consistency} and \emph{Design Effectiveness}.
Although models achieve high \emph{Execution Rate} under Intent-level instructions, their semantic and design scores remain lower than under Implementation-level instructions.
For example, Gemini 3.1 Pro Preview with Code Input improves from 3.688 to 4.128 in \emph{Semantic Consistency} and from 3.641 to 4.206 in \emph{Design Effectiveness} from Intent to Implementation level.
This indicates that current MLLMs can often produce executable code, but still struggle to infer suitable annotation targets and designs from abstract Intent-level instructions.
See Sec.~\ref{sec:instruction_specificity_discussion} for further analysis.

\begin{figure}[t]
    \centering
    \includegraphics[width=\columnwidth]{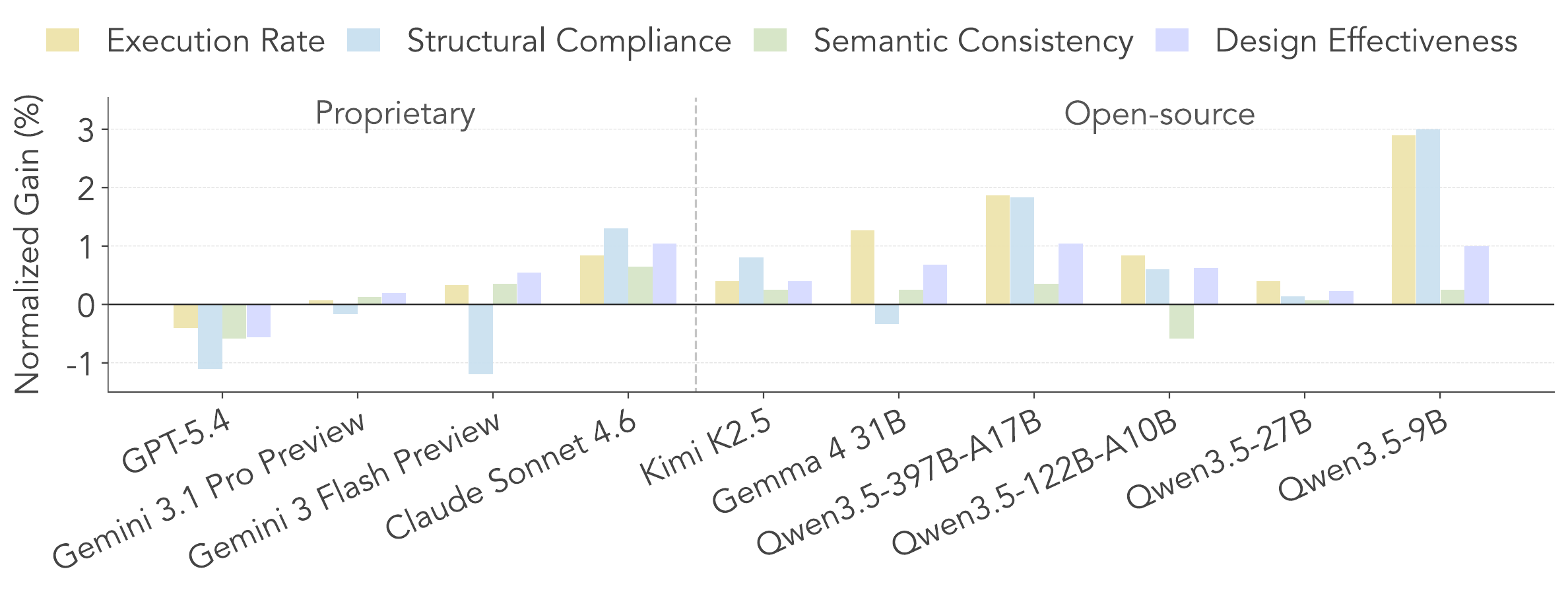}
    \vspace{-15pt}
    \caption{Normalized score gain from adding chart image input.}
    \label{fig:overall_image_input_gain}
\end{figure}

\bpstart{Chart image input brings marginal gains across evaluation metrics}
As shown in Fig.~\ref{fig:overall_image_input_gain}, the normalized gain for any individual model and metric remains within 3\%, with little change to the overall model ranking.
Averaged across all models and instruction levels, the normalized gains are 0.85\%, 0.49\%, 0.11\%, and 0.52\% in \emph{Execution Rate}, \emph{Structural Compliance}, \emph{Semantic Consistency}, and \emph{Design Effectiveness}, respectively.
All but \emph{Semantic Consistency} show significant improvements (all $p<.001$), yet effect sizes are negligible (Cohen's $g$ or $d_z<.10$).
Image input has larger effects on rule-based metrics, especially for open-source models, while LLM-judged gains mainly appear in \emph{Design Effectiveness}.

\subsubsection{Effects of Code, Complexity, and Instruction Specificity}
\label{sec:image_only}
\begin{figure}[t]
    \centering
    \includegraphics[width=\columnwidth]{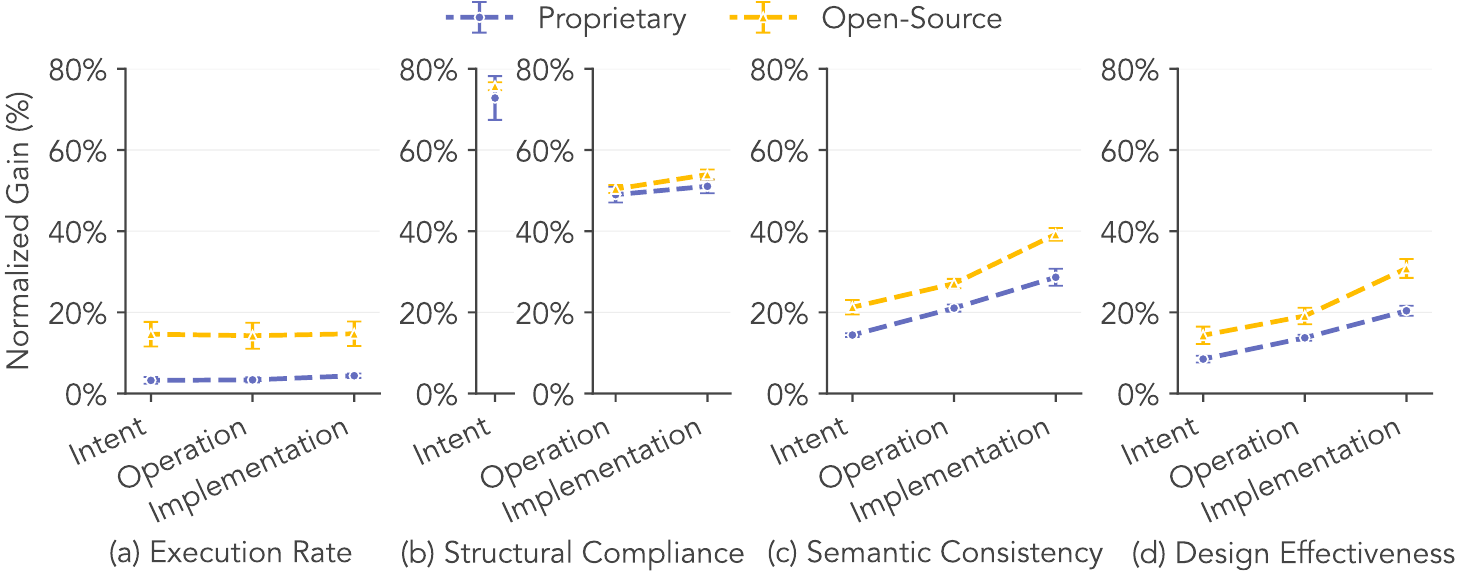}
    \caption{Normalized gains from adding chart code to Image-only input for proprietary and open-source models. Intent-level \emph{Structural Compliance} is shown separately because it measures \emph{Chart Fidelity} only.}
    \label{fig:code_gain}
    \vspace{-15pt}
\end{figure}

\textbf{Image-Only Ablation.} To quantify the contribution of chart code, we compare the Image-only setting with the corresponding Code + Image setting.
Overall, adding chart code substantially improves performance across evaluation metrics.
As shown in Fig.~\ref{fig:code_gain}, adding chart code yields larger gains for open-source models across most metrics, except for \emph{Structural Compliance}, where the two groups show comparable gains.
For the three quality metrics, gains generally increase with instruction specificity, with \emph{Structural Compliance} showing this pattern from Operation to Implementation levels, whereas gains in \emph{Execution Rate} remain relatively stable.
\emph{Structural Compliance} shows the largest gains overall. In particular, the substantial Intent-level gain, which reflects \emph{Chart Fidelity} alone, highlights the difficulty of reconstructing the unannotated chart from image input alone.
Together, these results show that chart code provides important grounding for annotation generation, particularly for open-source models and more detailed instructions.
Complete Image-only results are provided in Supplementary Material.

\begin{figure}[t]
    \centering
    \includegraphics[width=\columnwidth]{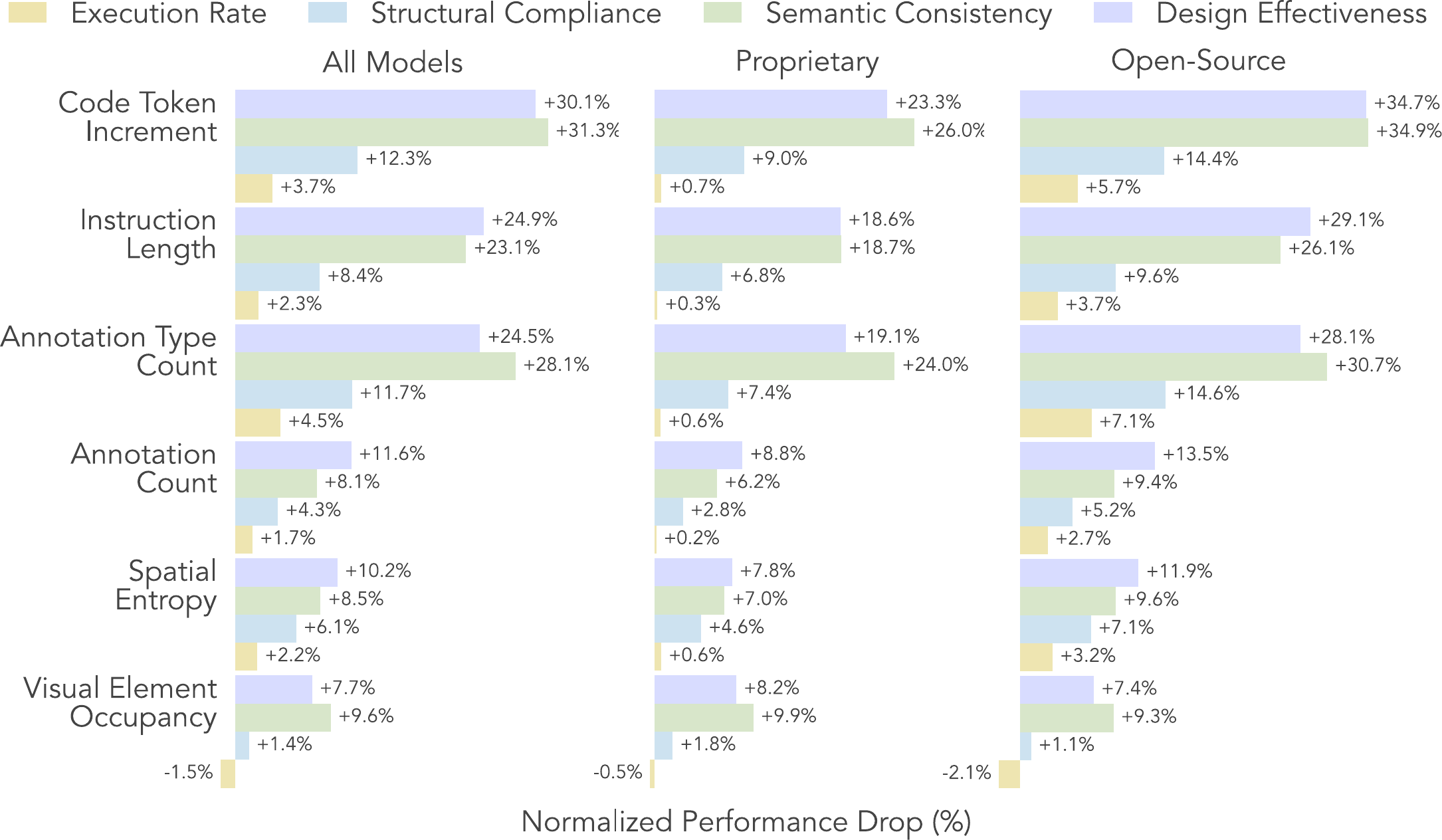}
    \caption{Normalized performance drops from simple to complex cases across complexity indicators and evaluation metrics.}
    \label{fig:complexity_factor_metric_drop}

\end{figure}

\bpstart{Effect of Task Complexity}
\label{sec:complexity_effect}
We examine how different aspects of complexity affect annotation generation using six indicators.
\underline{\textit{Code Token Increment}} measures the additional code tokens introduced by annotations relative to the unannotated chart, reflecting the amount of code modification required.
\underline{\textit{Instruction Length}} measures the amount of information provided to the model.
\underline{\textit{Annotation Type Count}} records the number of distinct annotation types in a chart, while \underline{\textit{Annotation Count}} measures the total number of annotation elements.
\underline{\textit{Annotation Spatial Distribution Entropy}} measures how broadly annotations are distributed across the chart, while \underline{\textit{Visual Element Occupancy}} measures the proportion of the chart area occupied by visual elements.
Detailed definitions and computation procedures are provided in the Supplementary Material.

For each indicator, we sort charts by the corresponding value and split them into three equal-sized groups: simple, medium, and complex.
We then compute the normalized performance drop from simple to complex cases. As shown in Fig.~\ref{fig:complexity_factor_metric_drop}, larger code token increments correspond to the largest drops, especially in \emph{Semantic Consistency} and \emph{Design Effectiveness}.
Higher annotation type counts and longer instructions also correspond to substantial performance declines, particularly in \emph{Semantic Consistency} and \emph{Design Effectiveness}.
The drops are generally more pronounced for open-source models, especially for larger code token increments.
These results show that \chartanno{} exposes model limitations associated with both task complexity, such as larger code modifications and longer instructions, and annotation complexity, such as a greater number of annotation types.
We also conduct complementary heatmap and regression analyses, which support these trends and are included in the Supplementary Material.

\begin{figure}[t]
    \centering
    \includegraphics[width=\columnwidth]{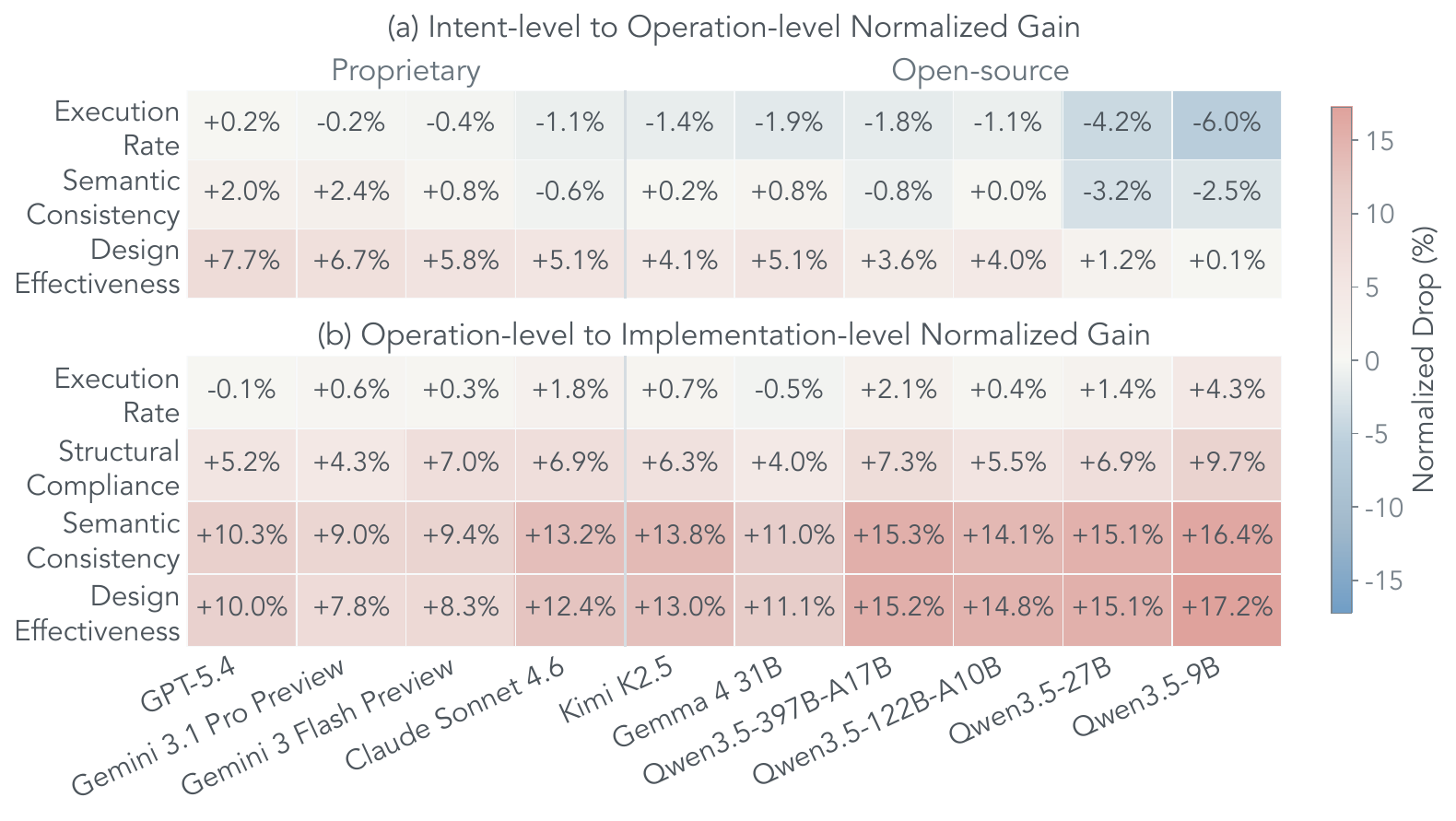}
    \vspace{-10pt}
    \caption{Normalized gains across instruction-level transitions. \emph{Structural Compliance} is omitted from the Intent-to-Operation comparison because its Intent-level definition includes only \emph{Chart Fidelity}.}
    \label{fig:instruction_specificity_gain}
    \vspace{-10pt}
\end{figure}

\bpstart{Effects of Instruction-Level Transitions}
\label{sec:instruction_specificity_discussion}
As reported in Sec.~\ref{sec:main_results}, performance generally improves as instructions become more detailed.
We further examine how these improvements differ across the two instruction-level transitions.
Fig.~\ref{fig:instruction_specificity_gain} shows distinct patterns across the two transitions.
From Intent-level to Operation-level instructions, stronger models generally benefit more, with the largest gains in \emph{Design Effectiveness}, whereas changes in \emph{Semantic Consistency} and \emph{Execution Rate} are smaller and even negative for some weaker models.
In contrast, the transition from Operation-level to Implementation-level instructions yields more consistent gains, particularly in \emph{Semantic Consistency} and \emph{Design Effectiveness}, with larger improvements for weaker models.
These patterns suggest that Operation-level instructions provide annotation strategies that stronger models are better able to translate into concrete implementations, whereas Implementation-level instructions specify annotation parameters more directly, reducing the implementation burden and thereby yielding larger gains for weaker models.

\subsubsection{Further Analyses}
\begin{figure}[t]
    \centering
    \includegraphics[width=\columnwidth]{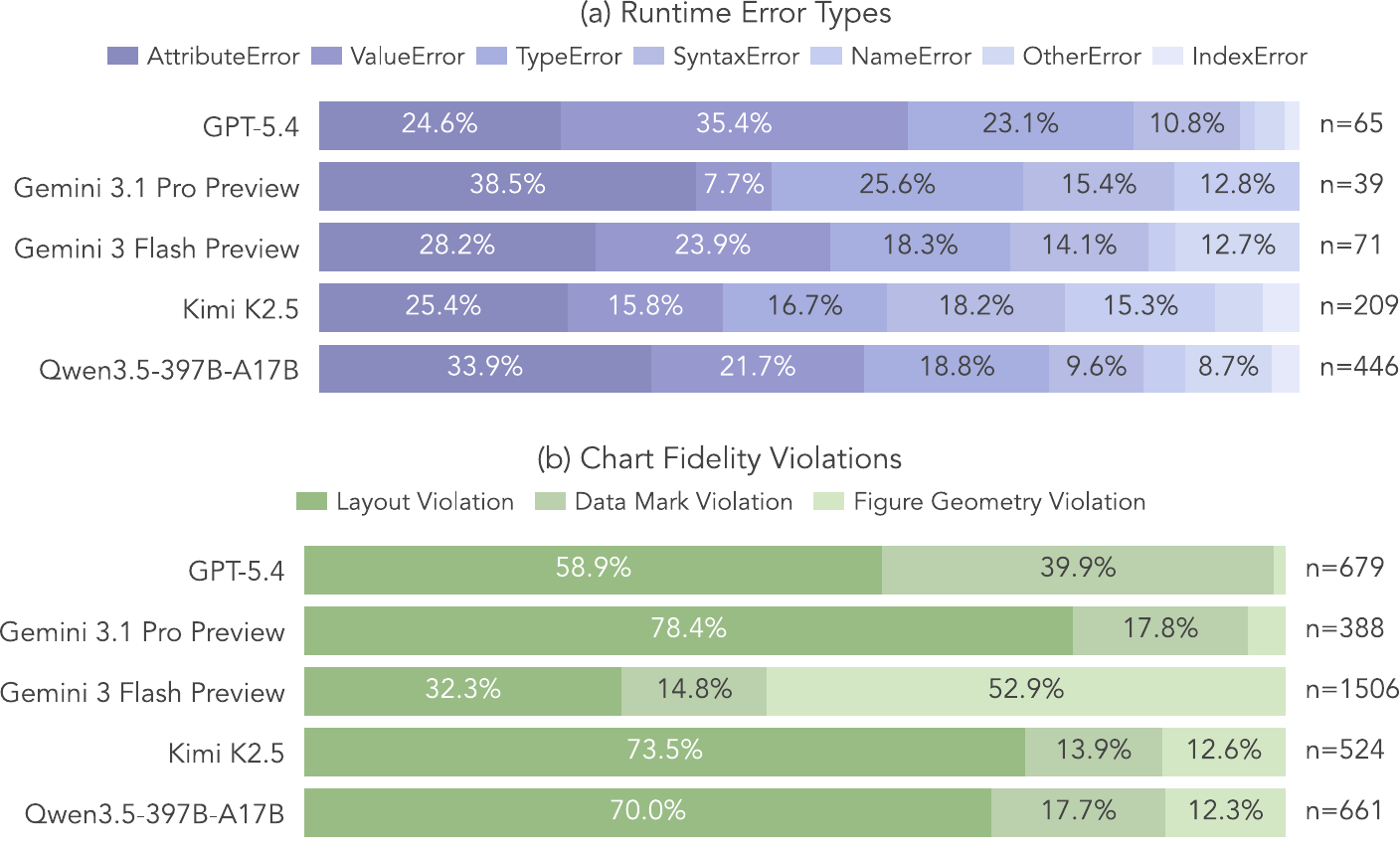}
    \caption{Error distributions for representative models. Bars show percentages; labels show counts.}
    \label{fig:error_analysis_representative}
    \vspace{-10pt}
\end{figure}

\textbf{Error Analysis.}
\label{sec:error_analysis}
We analyze model failures using distributional statistics and representative cases, focusing on two measurable binary failures: runtime errors and chart fidelity violations.
As shown in Fig.~\ref{fig:error_analysis_representative}, runtime errors are mainly caused by AttributeError, TypeError, and ValueError, suggesting failures in object references, value settings, and plotting API usage.
For chart fidelity violations, layout, data mark, and figure geometry violations are common across models, with layout violations dominating in most cases.
Gemini 3 Flash Preview is an exception, as its violations are dominated by figure geometry changes, which helps explain its low Intent-level \emph{Structural Compliance} despite strong semantic and design scores in Sec.~\ref{sec:main_results}.
Complete statistics and representative error cases are provided in Supplementary Material.

\bpstart{Validity of the LLM-based Judge}
\label{sec:judge_validation}
We further validate the GPT-5.4 judge on 600 randomly sampled outputs spanning models of different capability levels, including 300 outputs from Gemini 3.1 Pro Preview and 300 from weaker open-source models (150 each from Qwen3.5-9B and Qwen3.5-27B).
The samples cover both primary input settings and all instruction levels.
Three external visualization researchers (one PostDoc and two PhD students), each with at least two first-author IEEE TVCG publications, independently rated all 600 outputs.
Each rater spent approximately 3.5 hours and received \$50 in compensation; the human-evaluation protocol was approved by our institution's internal ethics review, and all raters provided informed consent.
GPT-5.4 judgments are then compared with the human ratings and repeated runs.
As shown in Tab.~\ref{tab:gpt_judge_validation}, GPT-5.4 shows strong overall alignment with human ratings and high repeat stability.
Supplementary analyses further show consistently high correlations across instruction levels and for both stronger and weaker models.

To assess potential judge-model bias arising from GPT-5.4's dual role as an evaluated model and the primary judge, we conduct a cross-judge analysis following the cross-validation strategy used in Plot2Code~\cite{wu2025plot2code}.
The analysis covers 1,500 outputs from five models, including GPT-5.4 itself, evaluated by GPT-5.4, Claude Sonnet 4.6, and Gemini 3.1 Pro Preview.
The three judges show strong overall consistency, with Cronbach's $\alpha$ values of 0.885, 0.898, and 0.913 for \emph{Semantic Consistency}, \emph{Design Effectiveness}, and Overall scores, respectively, and pairwise Spearman correlations ranging from 0.712 to 0.842 across the metrics.
Moreover, removing GPT-5.4 from the judge set leaves the ranking of the five evaluated models unchanged, providing further evidence against substantial GPT-5.4-specific judging bias.
Additional details are provided in Supplementary Material.

\begin{table}[t]
\caption{GPT-5.4 judge validation on 600 outputs. Human Alignment is measured by Spearman's $\rho$, and Repeat Stability by ICC(3,1).}
\centering
\small
\setlength{\tabcolsep}{8pt}
\begin{tabular}{lcc}
\toprule
\textbf{Metric} & \textbf{Human Alignment} & \textbf{Repeat Stability} \\
\midrule
\emph{Semantic Consistency} & 0.8194 & 0.9109 \\
\emph{Design Effectiveness} & 0.8378 & 0.9474 \\
Overall & 0.8593 & 0.9427 \\
\bottomrule
\end{tabular}
\label{tab:gpt_judge_validation}
\end{table}

\section{Generalization and External Validation}
\label{sec:representation_generalization}
We further examine the generalizability of \chartanno{} at both the dataset and evaluation, providing a basis for future extensions to additional visualization languages and representations.
To this end, we evaluate two representative extensions, D3 and SVG, on a random 10\% subset of \chartanno{} (120 charts).
For D3, we adopt a two-stage LLM-assisted conversion process.
We first use GPT-5.4 to convert the GT Code w/o annotation from Python to D3, followed by manual refinement.
We then construct the annotated D3 code by providing GPT-5.4 with both the unannotated D3 code and the annotated Python GT code.
For SVG, we execute each D3 implementation and serialize the resulting SVG DOM.
To reduce input tokens and LLM context burden, we simplify raw SVGs through structural compression, numerical precision reduction, and Ramer--Douglas--Peucker path simplification~\cite{douglas1973algorithms}, while manually verifying that chart structure and annotation content are preserved, reducing token count by 53.6\%.
Overall, the average GT code token counts for D3 and SVG are 3.48$\times$ and 6.61$\times$ that of Python, respectively.
For the instructions, we retain the Intent- and Operation-level instructions, while converting the Implementation-level instructions to match the target representation because they contain representation-dependent details.

\begin{figure}[!htbp]
    \centering
    \includegraphics[width=\columnwidth]{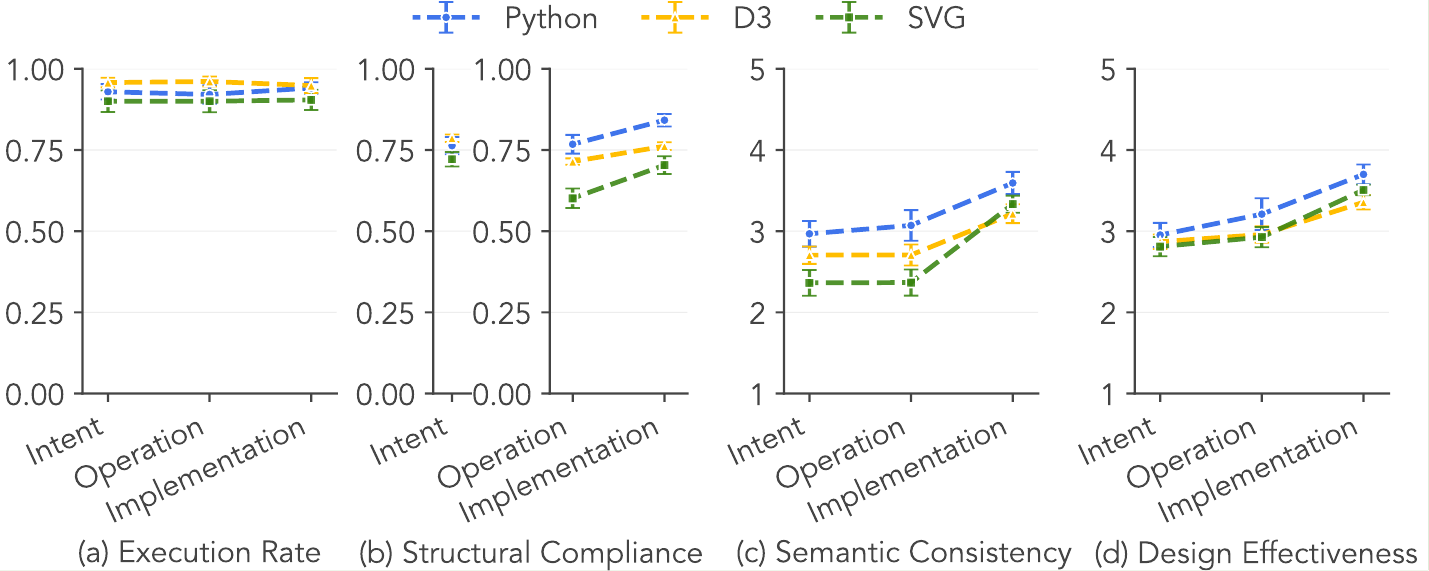}
    \caption{Average performance across Python, D3, and SVG representations over three instruction levels. Intent-level \emph{Structural Compliance} is shown separately, as in Fig.~\ref{fig:code_gain}.}
    \label{fig:representation_comparison}
\end{figure}

To examine the generalizability of our evaluation framework, we apply the same framework to D3 and SVG, adapting Graphical Element Extraction and Differential Annotation Extraction and Classification in the rule-based evaluation (Sec.~\ref{sec:rule_based_metrics}) to each representation.
We evaluate all models on the 120-chart subset under both representations.
Fig.~\ref{fig:representation_comparison} summarizes performance across representations and instruction levels. Detailed results are provided in Supplementary Material.

\bpstart{Key performance trends under Python generalize to D3 and SVG}
More specific instructions generally improve \emph{Semantic Consistency} and \emph{Design Effectiveness}, and the overall model ranking remains similar to that observed in the Python setting.

\bpstart{Representations exhibit different performance trade-offs across instruction levels}
Model performance is generally highest with Python in \emph{Semantic Consistency} and \emph{Design Effectiveness}.
For the rule-based metrics, models consistently achieve higher scores with D3 than with SVG across instruction levels, with smaller gaps at the Implementation level.
For the LLM-judged metrics, average model scores are higher with D3 than with SVG at the Intent and Operation levels, but slightly higher with SVG at the Implementation level in \emph{Semantic Consistency} (3.333 vs.\ 3.215) and \emph{Design Effectiveness} (3.507 vs.\ 3.356).
This reversal in average performance suggests that SVG benefits more from explicit Implementation-level guidance and that input length alone may not fully explain performance.

\section{Discussion}

\textbf{Implications for Chart Annotation Authoring.}
Our findings suggest several directions for future MLLM-based annotation authoring systems.
First, the difficulty of Intent-level generation indicates that current models still struggle to translate abstract communicative goals into appropriate annotation targets and designs.
Interactive and mixed-initiative authoring can help users progressively articulate and refine such design intentions~\cite{ren2017chartaccent,srinivasan2025pluto}.
Rather than relying on one-shot generation, future systems could support progressive specification and refinement, allowing users to move from high-level intent toward more explicit annotation operations or implementation details when needed.
Extending prior annotation grammars such as AnnoGram and ChartMark~\cite{annogram_2025,chartmark_2025}, our structured representation could support progressive refinement from abstract intents to concrete annotation specifications.

Second, the strong benefit of chart code and the limited additional gain from chart images suggest that executable chart representations provide semantic and structural grounding for annotation generation, while images offer complementary visual information.
Future authoring systems could therefore combine code-based reasoning with visual feedback, using the former to preserve chart semantics and structure and the latter to support annotation layout and visual integration.

Finally, recent work has improved chart generation and editing through specialized data construction and post-training~\cite{zhao2025chartcoder,tang2026mmrecoder,chen2026charteditor}.
For chart annotation authoring, such improvement can be guided by evaluation signals that identify failures in conveying intended information, associating annotations with chart elements, and providing appropriate contextual and visual emphasis~\cite{ren2017chartaccent,hullman2013contextifier,kong2012graphicaloverlays,stokes2023striking}.
Our multidimensional evaluation provides such signals across structural compliance, semantic communication, and visual design, supporting targeted regeneration and failure-focused data construction for post-training MLLMs.

\bpstart{Limitations and Future Work}
Our primary evaluation focuses on Python-based static charts, while the D3 and SVG study on a benchmark subset demonstrates the generalizability of both the dataset and evaluation framework.
Future work could extend \chartanno{} to additional representations and interactive or animated charts.
Although \chartanno{} includes public-facing and scientific visualizations from diverse real-world sources, future work could further strengthen coverage of specific application domains through targeted sampling, including domains such as public health~\cite{hines2026chartingpublichealth} and professional data journalism~\cite{rahatzaman2026annobench}.
Finally, as MLLMs evolve, we will continue updating the \chartanno{} GitHub leaderboard to track emerging models.

\section{Conclusion}

We introduced \chartanno{}, a benchmark for evaluating MLLMs on chart annotation generation across three instruction levels and two primary chart input settings.
Our evaluation of 10 MLLMs shows that proprietary models remain stronger overall, although large open-source models narrow the performance gap.
More detailed instructions improve annotation quality, suggesting that models still struggle with abstract Intent-level generation.
Providing chart images beyond chart code brings limited gains.
The Image-only ablation performs substantially worse than the code-based settings.
Additional analyses show that performance varies with multiple task complexity indicators and across instruction-level transitions.
Further analyses characterize common failure modes and validate the reliability of the LLM-based judge.
Experiments with D3 and SVG demonstrate the generalizability of \chartanno{}.
We hope \chartanno{} can support research on more reliable chart annotation and visual communication.

\bibliographystyle{abbrv-doi-hyperref}

\bibliography{custom}

\clearpage
\appendix

\clearpage
\setcounter{page}{1}
\renewcommand{\thepage}{\arabic{page}}

\section*{Supplementary Material}
\addcontentsline{toc}{section}{Supplementary Material}

\startcontents[supplementary]
\printcontents[supplementary]{}{1}{\setcounter{tocdepth}{2}}

\clearpage
\section{Comparison with Related Benchmarks}
\label{app:benchmark_comparison}

We compare \chartanno{} with representative benchmarks from two related task families: chart editing tasks and annotation tasks.

\subsection{Comparison with Chart Editing Tasks}

\bpstart{Instruction and Code Complexity.}
We further compare \chartanno{} with ChartEdit, the closest code-based chart editing benchmark, in terms of instruction and target code complexity. As shown in Tab.~\ref{tab:chartanno_chartedit_length}, \chartanno{} contains longer instructions and target code, suggesting additional complexity in annotation generation.

\begin{table}[htbp]
\caption{Instruction and code length comparison.}
\centering
\small
\setlength{\tabcolsep}{5pt}
\begin{tabular}{lcc}
\toprule
\textbf{Benchmark} &
\textbf{Instr.} &
\textbf{Code} \\
\midrule
ChartEdit
& 20.08 & 758.12 \\
\chartanno{} (Operation)
& 95.06 & 1535.68 \\
\bottomrule
\end{tabular}
\label{tab:chartanno_chartedit_length}
\end{table}

\bpstart{Annotation Source Analysis.}
Following prior studies on visualization annotation practices and design spaces~\citep{rahman2025annotationsurvey,rahman2025annotationdesignspace}, we categorize annotation sources into three non-exclusive types: Internal (information available from the chart), Derived (information computed or inferred from data), and External (information obtained outside the chart).

Tab.~\ref{tab:annotation_source_analysis} summarizes the distribution of annotation sources in \chartanno{}.

\begin{table}[htbp]
\caption{Distribution of annotation sources. Categories are non-exclusive.}
\centering
\small
\setlength{\tabcolsep}{8pt}
\begin{tabular}{lcc}
\toprule
\textbf{Source} &
\textbf{Charts} &
\textbf{Ratio} \\
\midrule
Internal
& 595 & 49.58\% \\
Derived
& 543 & 45.25\% \\
External
& 918 & 76.50\% \\
\bottomrule
\end{tabular}
\label{tab:annotation_source_analysis}
\end{table}

We further estimate the proportion of instruction instances requiring reasoning beyond direct visual modification. At least 63.5\% of annotation instances involve non-trivial reasoning requirements:

\[
\frac{1200 + 2 \times 543}{3600}=63.5\%.
\]

This conservative estimate considers two sources of reasoning. First, all Intent-level instructions require inferring communication goals and annotation strategies. Second, derived annotations in Operation-level and Implementation-level instructions require identifying relevant values or patterns from the underlying data before generating annotations.
Therefore, a substantial portion of \chartanno{} requires reasoning about why and how annotations should be introduced, beyond applying predefined visual edits.

\subsection{Comparison with Annotation Tasks}
\label{app:annobench_comparison}

We further examine AnnoBench~\cite{rahatzaman2026annobench}, focusing on its 58 professional charts whose associated annotation tasks are derived from the original real-world visualizations.
We reconstruct the corresponding annotated GT references through LLM-assisted conversion and manual refinement, and manually review the instructions, released code or data, and reconstructed GT references for all 58 samples.

We find that 30 samples (51.7\%) cannot be directly evaluated under our GT-referenced Operation- and Implementation-level protocol for one or more of three reasons, although most remain compatible with AnnoBench's reference-free evaluation protocol.
As these categories overlap, 28 samples remain for direct comparison under our protocol.
The examples below illustrate these cases together with the corresponding annotation outputs generated by Claude Sonnet 4.6.

\begin{figure}[!htbp]
    \centering
    \includegraphics[width=\columnwidth]{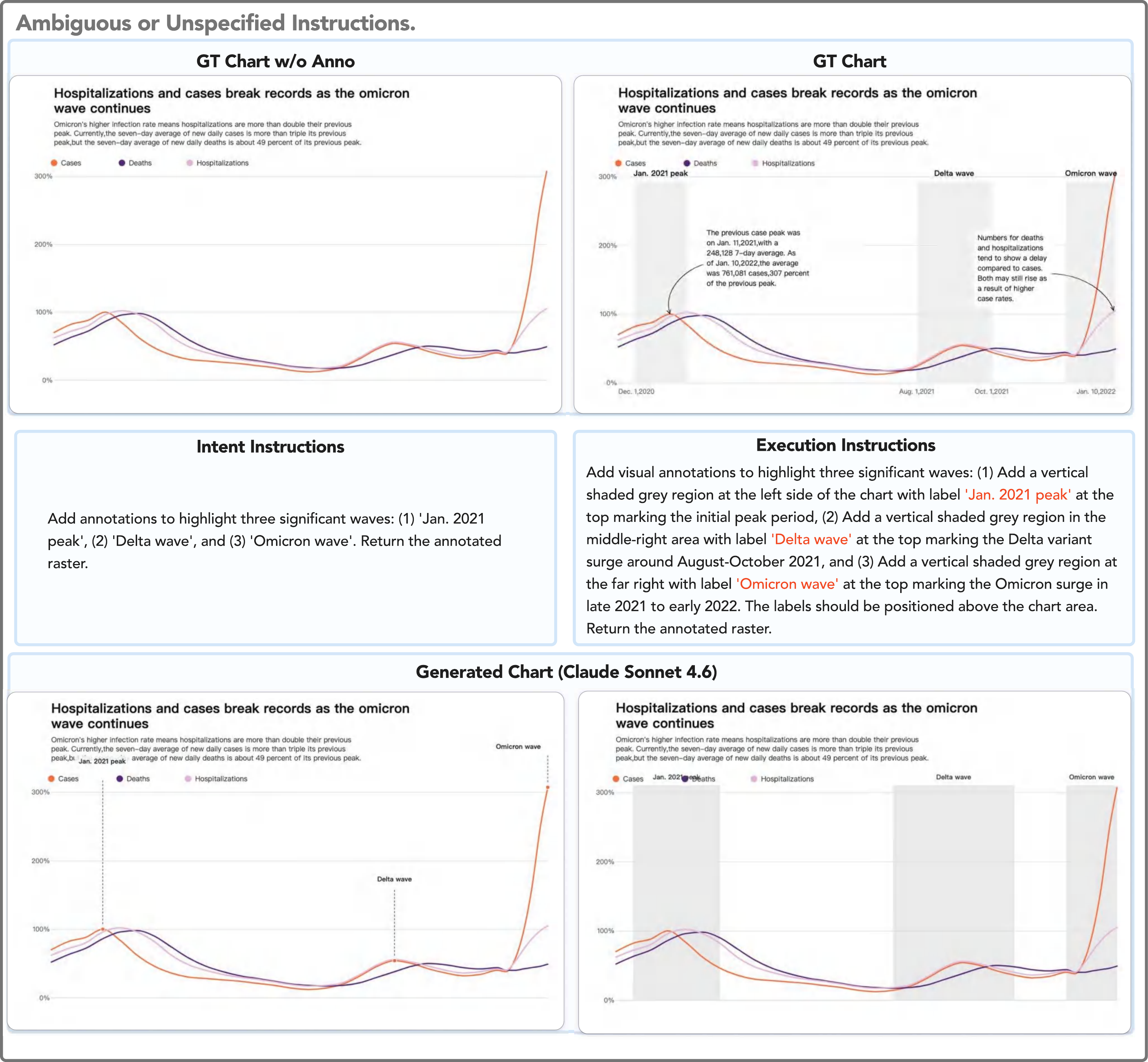}
    \caption{Example of limited specification in AnnoBench. The instruction omits the required annotation text, making it impossible to reproduce the original chart.}
    \label{fig:annobench_limited_specification}
\end{figure}

\begin{figure}[!htbp]
    \centering
    \includegraphics[width=\columnwidth]{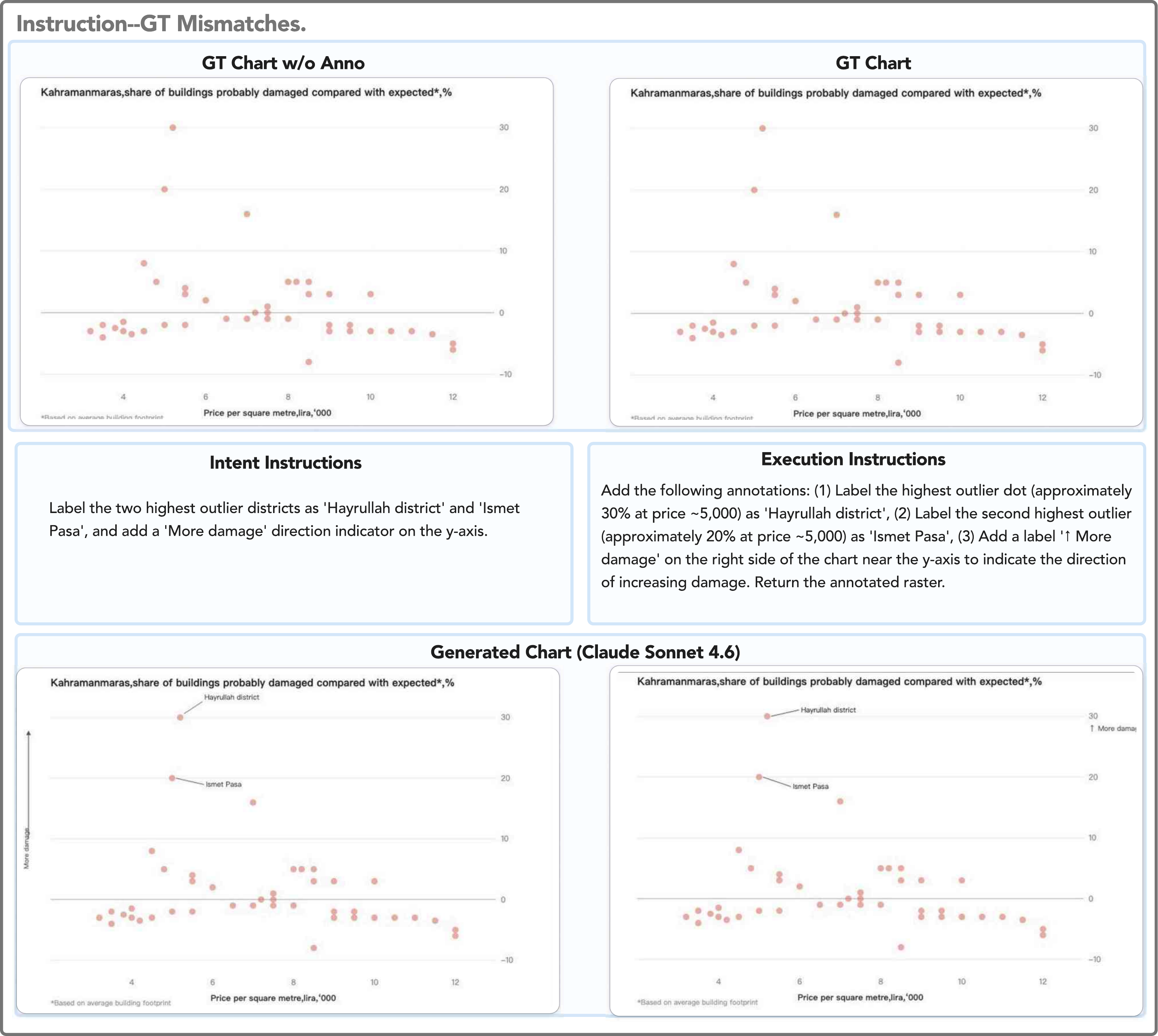}
    \caption{Example of an instruction--GT mismatch in AnnoBench. The instruction requires reproducing annotation elements that are absent from the original chart.}
    \label{fig:annobench_instruction_gt_mismatch}
\end{figure}

\begin{figure}[!htbp]
    \centering
    \includegraphics[width=\columnwidth]{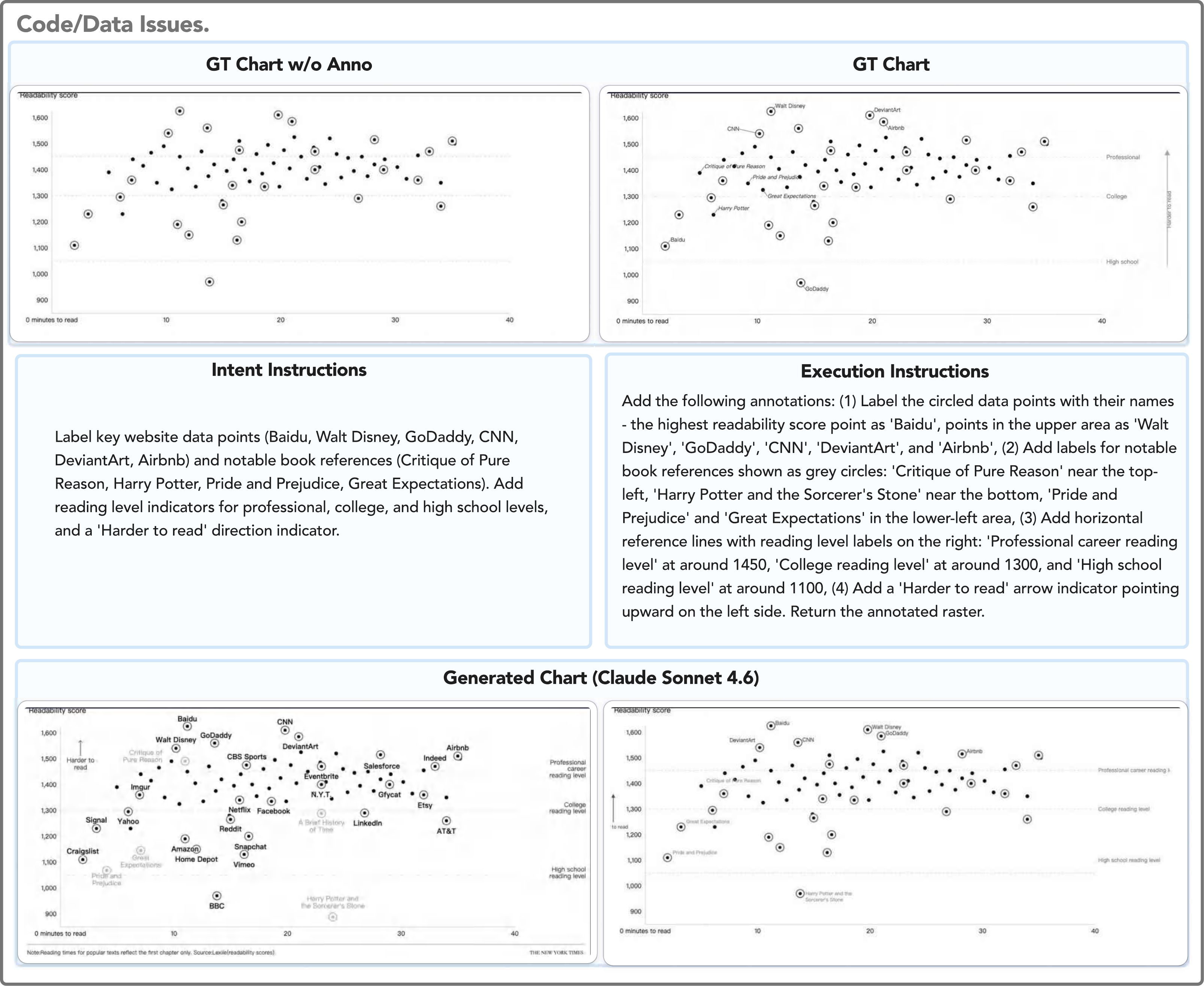}
    \caption{Example of a code/data limitation in AnnoBench. The released code lacks the required labels, making it impossible to reproduce the original chart.}
    \label{fig:annobench_code_data_limitation}
\end{figure}

\begin{itemize}
    \item \emph{Ambiguous or Unspecified Instructions} ($n=21$): annotation details such as components, text, color, position, or range are not specified.
    These samples remain compatible with reference-free Intent-level evaluation, but not with our GT-referenced Operation- and Implementation-level evaluation, as shown in Fig.~\ref{fig:annobench_limited_specification}.

    \item \emph{Instruction--GT Mismatches} ($n=7$): the instruction content does not directly correspond to the reconstructed GT or source chart, so these samples cannot be evaluated against the GT reference, as illustrated in Fig.~\ref{fig:annobench_instruction_gt_mismatch}.

    \item \emph{Code/Data Issues} ($n=4$): required information, such as labels referenced by the instruction, is absent from the released code or data, preventing faithful reproduction of the requested annotation, as shown in Fig.~\ref{fig:annobench_code_data_limitation}.
\end{itemize}

\bpstart{Complexity Comparison with \chartanno{}.}
We compare these 28 AnnoBench samples with the 120-chart \chartanno{} subset from our cross-representation experiments.
Compared with AnnoBench, \chartanno{} has 1.69$\times$ longer instructions (86.6 vs.\ 51.2 words) and 2.79$\times$ more annotation elements per chart (20.55 vs.\ 7.36).
Its average code token increment is also 2.19$\times$ larger for D3 (1,374.2 vs.\ 627.1) and 3.87$\times$ larger for SVG (2,612.7 vs.\ 674.8).
These differences indicate greater task complexity in \chartanno{}, particularly in annotation density and required code modification.

\section{Data Construction Details}
\label{app:data_construction_details}

This section provides implementation details for the data construction process described in Sec.~3.2 of the main paper.

\subsection{Source Licenses and Intended Use}
\label{app:source_licenses}

Existing chart sources are used as references for manual reconstruction and annotation revision, rather than directly redistributed as original images.
Our use is limited to research and evaluation and is compatible with the public benchmark or scholarly use contexts of the source artifacts.
The released artifact is intended for research and evaluation.
Tab.~\ref{tab:source_licenses} summarizes the licenses and access conditions of the source benchmarks and datasets.

\begin{table}[!htbp]
\caption{Licenses and access conditions of source benchmarks and datasets.}
\centering
\small
\setlength{\tabcolsep}{4pt}
\begin{tabularx}{\columnwidth}{p{0.38\columnwidth}X}
\toprule
\textbf{Source} & \textbf{License / Access Condition} \\
\midrule
ChartQAPro~\citep{masry2025chartqapro}
& MIT license. \\
CharXiv~\citep{wang2024charxiv}
& Data CC BY-SA 4.0; code Apache 2.0. \\
ChartMimic~\citep{shi2025chartmimic}
& Data and codebase Apache-2.0. \\
MatPlotBench~\citep{matplotagent_2024}
& Public benchmark; license not explicitly specified. \\
Rahman et al.~\citep{rahman2025annotationdesignspace}
& Paper and supplemental materials under CC BY 4.0. \\
Stokes et al.~\citep{stokes2025textfunctions}
& Paper and supplemental materials under CC BY 4.0. \\
Recent arXiv papers
& CC BY 4.0. \\
\bottomrule
\end{tabularx}
\label{tab:source_licenses}
\end{table}

\subsection{Manual Selection and Dataset Coverage}
\label{app:manual_selection}

The review is conducted in multiple rounds.
We then review candidates by chart type to reduce visual and structural redundancy.
The final round verifies that the selected charts satisfy the above criteria.

Tab.~\ref{tab:chart_distribution} reports the distribution of chart types across data sources.
CM, R, CQA, CX, MPB, and S denote ChartMimic, Rahman et al., ChartQAPro, CharXiv, MatPlotBench, and Stokes et al., respectively.
Darker cells indicate more samples.

\begin{table}[t]
\caption{Distribution of chart types across data sources. Darker cells indicate more samples.}
\centering
\small
\setlength{\tabcolsep}{5pt}
\renewcommand{\arraystretch}{1.08}
\resizebox{\columnwidth}{!}{
\begin{tabular}{lrrrrrrrr}
\toprule
\textbf{Chart Type}
& \textbf{CM}
& \textbf{R}
& \textbf{CQA}
& \textbf{CX}
& \textbf{MPB}
& \textbf{S}
& \textbf{ArXiv}
& \textbf{Total} \\
\midrule
Area
& \cellcolor{blue!0}0
& \cellcolor{blue!8}13
& \cellcolor{blue!9}15
& \cellcolor{blue!0}0
& \cellcolor{blue!0}0
& \cellcolor{blue!2}4
& \cellcolor{blue!1}1
& \cellcolor{blue!11}33 \\

Bar
& \cellcolor{blue!19}41
& \cellcolor{blue!10}19
& \cellcolor{blue!29}114
& \cellcolor{blue!4}6
& \cellcolor{blue!0}0
& \cellcolor{blue!8}12
& \cellcolor{blue!11}21
& \cellcolor{blue!37}213 \\

Box
& \cellcolor{blue!8}13
& \cellcolor{blue!0}0
& \cellcolor{blue!0}0
& \cellcolor{blue!1}1
& \cellcolor{blue!0}0
& \cellcolor{blue!0}0
& \cellcolor{blue!1}2
& \cellcolor{blue!8}16 \\

Combination
& \cellcolor{blue!0}0
& \cellcolor{blue!0}0
& \cellcolor{blue!12}25
& \cellcolor{blue!0}0
& \cellcolor{blue!0}0
& \cellcolor{blue!1}1
& \cellcolor{blue!3}5
& \cellcolor{blue!11}31 \\

Contour
& \cellcolor{blue!9}15
& \cellcolor{blue!0}0
& \cellcolor{blue!0}0
& \cellcolor{blue!0}0
& \cellcolor{blue!0}0
& \cellcolor{blue!0}0
& \cellcolor{blue!0}0
& \cellcolor{blue!8}15 \\

Density
& \cellcolor{blue!10}19
& \cellcolor{blue!0}0
& \cellcolor{blue!0}0
& \cellcolor{blue!0}0
& \cellcolor{blue!0}0
& \cellcolor{blue!0}0
& \cellcolor{blue!0}0
& \cellcolor{blue!9}19 \\

Dot
& \cellcolor{blue!0}0
& \cellcolor{blue!0}0
& \cellcolor{blue!2}4
& \cellcolor{blue!0}0
& \cellcolor{blue!0}0
& \cellcolor{blue!1}2
& \cellcolor{blue!0}0
& \cellcolor{blue!6}6 \\

Error point
& \cellcolor{blue!19}46
& \cellcolor{blue!0}0
& \cellcolor{blue!0}0
& \cellcolor{blue!0}0
& \cellcolor{blue!0}0
& \cellcolor{blue!1}1
& \cellcolor{blue!0}0
& \cellcolor{blue!13}47 \\

Heatmap
& \cellcolor{blue!13}28
& \cellcolor{blue!0}0
& \cellcolor{blue!1}1
& \cellcolor{blue!4}6
& \cellcolor{blue!0}0
& \cellcolor{blue!0}0
& \cellcolor{blue!6}9
& \cellcolor{blue!12}44 \\

Histogram
& \cellcolor{blue!12}24
& \cellcolor{blue!4}6
& \cellcolor{blue!1}1
& \cellcolor{blue!0}0
& \cellcolor{blue!0}0
& \cellcolor{blue!0}0
& \cellcolor{blue!1}1
& \cellcolor{blue!11}32 \\

Line
& \cellcolor{blue!12}25
& \cellcolor{blue!9}17
& \cellcolor{blue!34}144
& \cellcolor{blue!6}9
& \cellcolor{blue!1}1
& \cellcolor{blue!9}17
& \cellcolor{blue!15}34
& \cellcolor{blue!42}247 \\

Multi-plot
& \cellcolor{blue!14}33
& \cellcolor{blue!0}0
& \cellcolor{blue!32}133
& \cellcolor{blue!12}27
& \cellcolor{blue!1}1
& \cellcolor{blue!3}5
& \cellcolor{blue!28}104
& \cellcolor{blue!50}303 \\

Pie
& \cellcolor{blue!10}20
& \cellcolor{blue!8}13
& \cellcolor{blue!4}6
& \cellcolor{blue!0}0
& \cellcolor{blue!1}2
& \cellcolor{blue!0}0
& \cellcolor{blue!2}3
& \cellcolor{blue!12}44 \\

Radar
& \cellcolor{blue!5}8
& \cellcolor{blue!3}5
& \cellcolor{blue!0}0
& \cellcolor{blue!0}0
& \cellcolor{blue!0}0
& \cellcolor{blue!0}0
& \cellcolor{blue!4}6
& \cellcolor{blue!9}19 \\

Scatter
& \cellcolor{blue!11}21
& \cellcolor{blue!9}15
& \cellcolor{blue!9}14
& \cellcolor{blue!2}4
& \cellcolor{blue!0}0
& \cellcolor{blue!6}10
& \cellcolor{blue!14}30
& \cellcolor{blue!20}94 \\

Treemap
& \cellcolor{blue!9}15
& \cellcolor{blue!1}1
& \cellcolor{blue!0}0
& \cellcolor{blue!0}0
& \cellcolor{blue!0}0
& \cellcolor{blue!1}1
& \cellcolor{blue!0}0
& \cellcolor{blue!9}17 \\

Violin
& \cellcolor{blue!10}19
& \cellcolor{blue!0}0
& \cellcolor{blue!0}0
& \cellcolor{blue!0}0
& \cellcolor{blue!0}0
& \cellcolor{blue!1}1
& \cellcolor{blue!0}0
& \cellcolor{blue!9}20 \\

\midrule
\textbf{Total}
& \cellcolor{blue!50}\textbf{327}
& \cellcolor{blue!18}\textbf{89}
& \cellcolor{blue!50}\textbf{457}
& \cellcolor{blue!12}\textbf{53}
& \cellcolor{blue!2}\textbf{4}
& \cellcolor{blue!12}\textbf{54}
& \cellcolor{blue!34}\textbf{216}
& \cellcolor{blue!50}\textbf{1200} \\
\bottomrule
\end{tabular}
}
\label{tab:chart_distribution}
\end{table}

\subsection{Reconstruction Quality and Correction Process}
\label{app:reconstruction_quality}

We provide additional details on the reconstruction quality evaluation and correction process described in Sec.~3.2 of the main paper.
For charts from sources other than ChartMimic, we evaluate the initial reconstructions using GPT-5.4~\cite{openai2026gpt54} as the judge and a 100-point rubric adapted from ChartMimic~\cite{shi2025chartmimic}.
The rubric assesses consistency with the source image across six dimensions: Chart Type, Layout, Text Content, Data, Style, and Clarity.

Tab.~\ref{tab:reconstruction_quality_overall} reports the per-dimension average scores of the initial reconstructions.

\begin{table}[!htbp]
\caption{Average reconstruction quality scores for initial reconstructions of charts from sources other than ChartMimic. The total score is computed out of 100.}
\centering
\small
\setlength{\tabcolsep}{6pt}
\resizebox{0.5\columnwidth}{!}{
\begin{tabular}{lc}
\toprule
\textbf{Dimension} & \textbf{Average Score} \\
\midrule
Total & 92.34 \\
Chart Type & 19.99 \\
Layout & 9.88 \\
Text Content & 17.78 \\
Data & 17.95 \\
Style & 16.91 \\
Clarity & 9.83 \\
\bottomrule
\end{tabular}
}
\label{tab:reconstruction_quality_overall}
\end{table}

Following this evaluation, we manually refine the reconstructed charts to improve their consistency with the corresponding source images.
We additionally inspect the source annotations and correct three types of issues that could compromise the quality of the resulting references.

\begin{figure}[!htbp]
    \centering
    \includegraphics[width=\columnwidth]{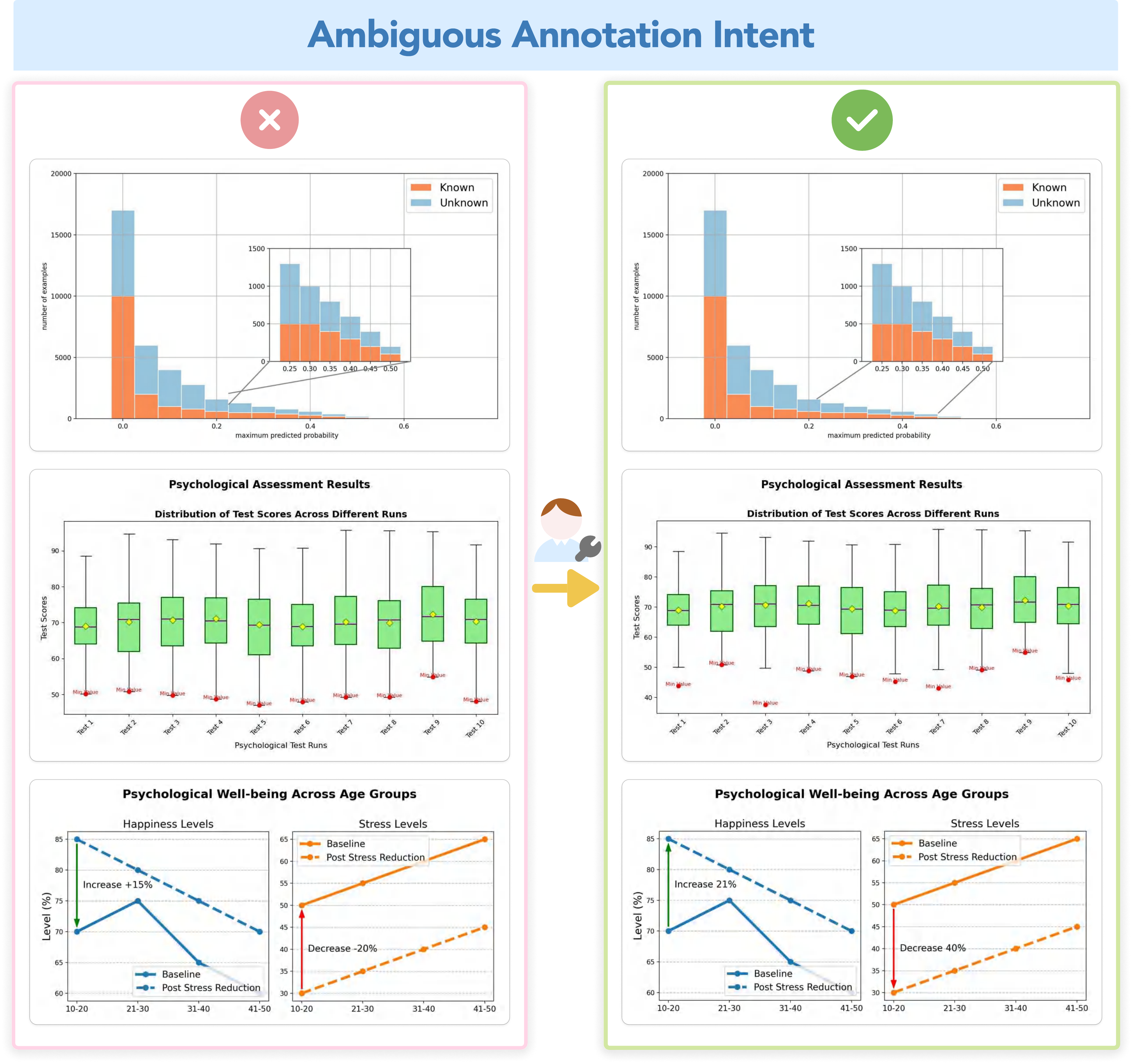}
    \caption{Examples of correcting ambiguous annotation intent. The left side shows annotations with unclear or inconsistent targets, directions, or connections; the right side shows the corrected versions.}
    \label{fig:ambiguous_annotation_intent}
\end{figure}

\textbf{Ambiguous Annotation Intent.}
Some annotations have targets, directions, or connections that are unclear or inconsistent with the surrounding chart content.
We inspect the chart content and underlying data to recover the intended relation, and revise the annotation text, anchor position, direction, or connection accordingly.
Fig.~\ref{fig:ambiguous_annotation_intent} shows representative examples.

\begin{figure}[!htbp]
    \centering
    \includegraphics[width=\columnwidth]{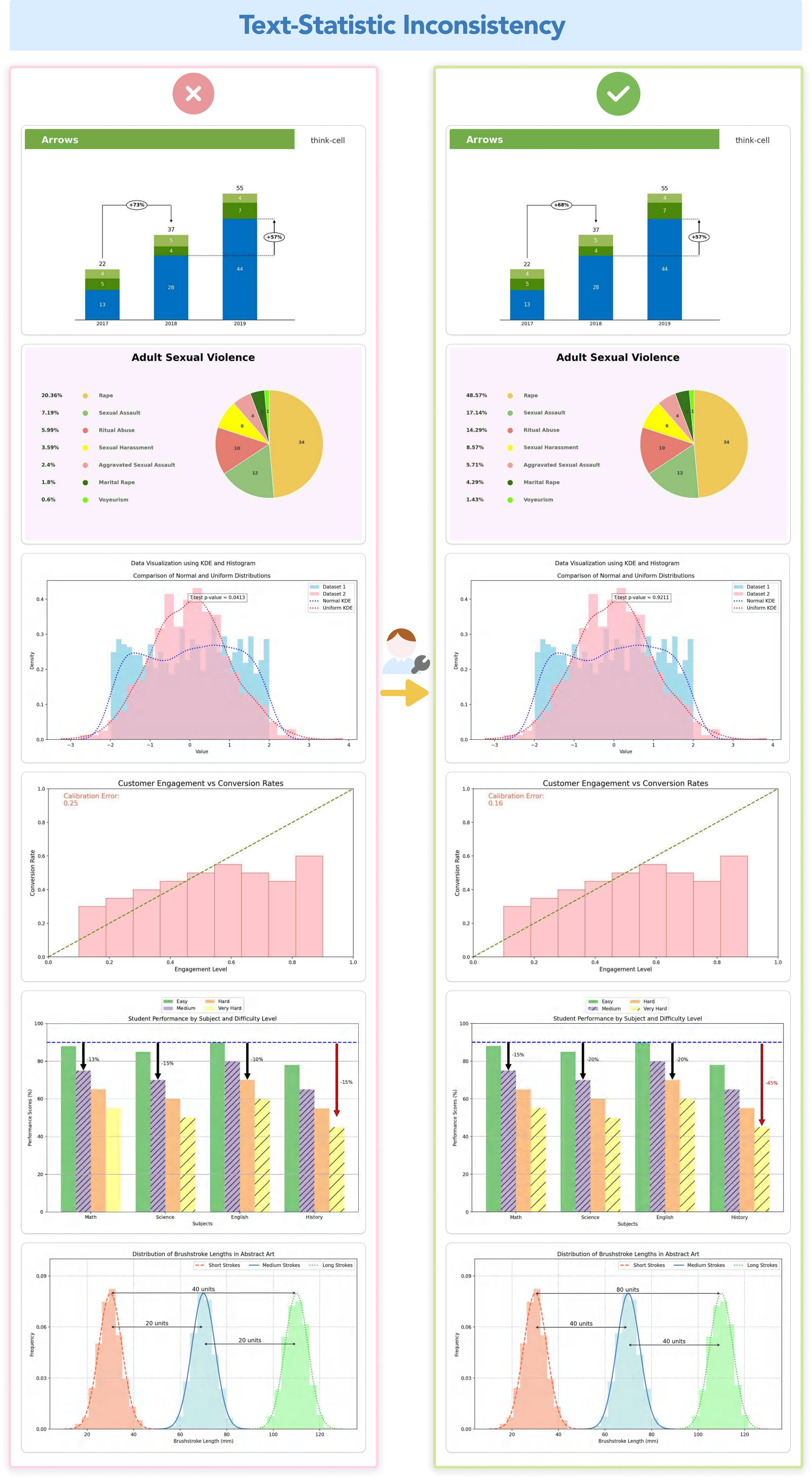}
    \caption{Examples of correcting factual inconsistencies. The left side shows annotation text that conflicts with the underlying data or derived statistics; the right side shows the corrected versions.}
    \label{fig:text_statistic_inconsistency}
\end{figure}

\textbf{Factual Inconsistencies.}
Some annotation text conflicts with the underlying data or derived statistics, such as incorrect percentage changes or summary values; we recompute the relevant quantities and revise the text accordingly, as shown in Fig.~\ref{fig:text_statistic_inconsistency}.

\begin{figure}[!htbp]
    \centering
    \includegraphics[width=\columnwidth]{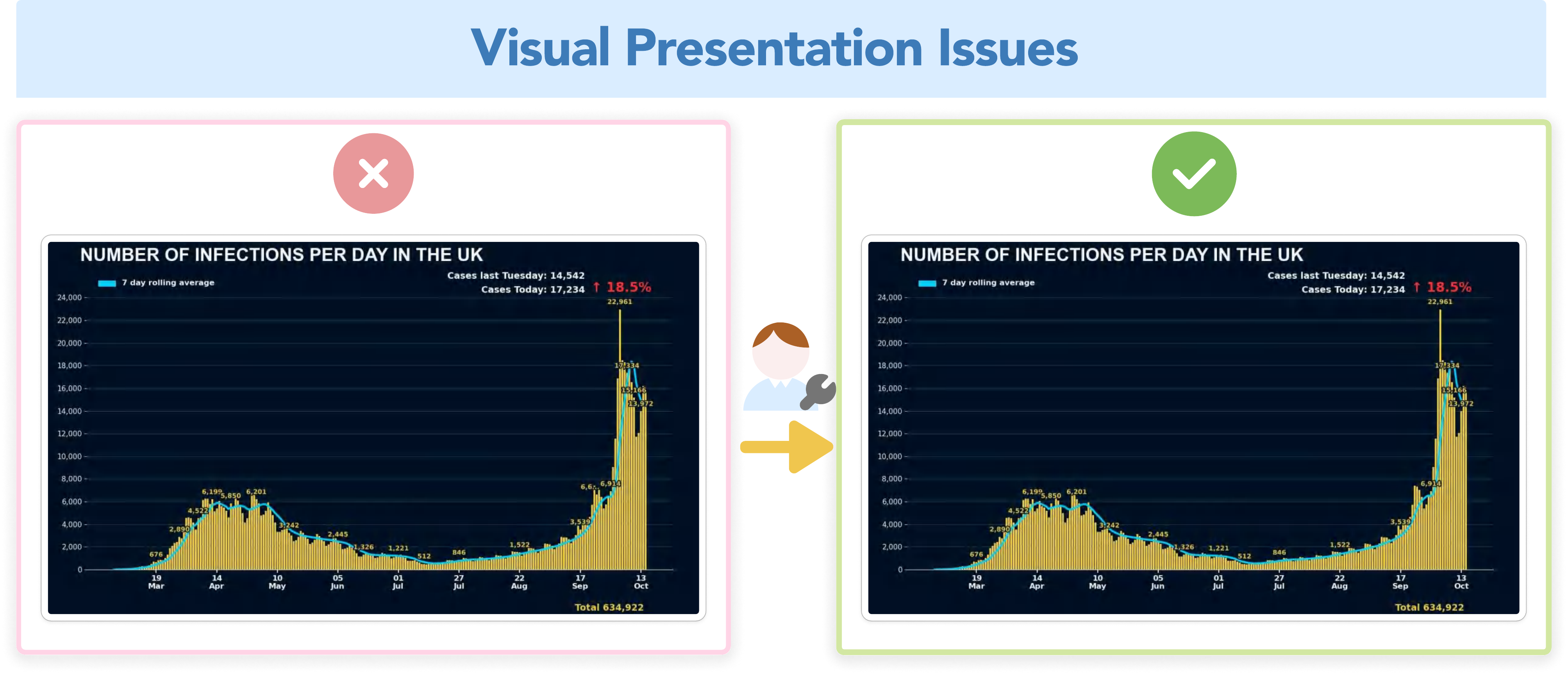}
    \caption{Examples of correcting visual presentation issues. The left side shows truncated, occluded, or otherwise unclear annotations; the right side shows the corrected versions.}
    \label{fig:visual_presentation_issues}
\end{figure}

\textbf{Visual Presentation Issues.}
Some annotations are difficult to interpret because of presentation problems such as truncated text, occluded labels, or unclear placement.
When the intended meaning is recoverable, we correct these issues by adjusting the annotation position, style, or text.
If an annotation cannot be corrected without introducing ambiguity, we remove it from the reference chart.
Fig.~\ref{fig:visual_presentation_issues} shows representative examples.

\section{Rule-based Evaluation Details}
\label{app:rule_based_metrics}

\subsection{Execution Environment}
\label{app:execution_environment}

Generated programs are executed in a sandboxed Python environment with Python 3.12.13, Matplotlib 3.10.8, NumPy 2.2.6, and Pandas 3.0.2.
Each program is executed with a timeout of 120 seconds.
Rendered charts are saved as JPG images using the output settings specified by each generated program.
Programs that fail to execute or exceed the timeout receive zero for all downstream quality metrics.

\subsection{Rendered Element Representation}
\label{app:rule_based_representation}

The rule-based metrics build on the differential annotation extraction and classification procedure described in Sec.~3.3.1 of the main paper, which matches elements in the annotated and generated charts against the unannotated counterpart and classifies unmatched elements into seven annotation categories.
The following subsections present the corresponding implementations: annotation element extraction, chart fidelity, annotation matching, and color matching.

\subsection{Annotation Element Extraction}
\label{app:annotation_extraction}

We summarizes the annotation extraction process.
The extractor identifies elements from rendered charts, removes elements already present in the unannotated chart, and groups the remaining elements into seven annotation categories.

\begin{codebox}{Listing 1: Annotation Element Extraction}
def extract_annotations(annotated_fig, unannotated_fig):
    base_elements = extract_visible_elements(unannotated_fig)
    anno_elements = extract_visible_elements(annotated_fig)

    added_elements = diff_elements(anno_elements, base_elements)

    annotations = {
        "enclosure": [],
        "connector": [],
        "text": [],
        "glyph": [],
        "color": [],
        "indicator": [],
        "geometric": [],
    }

    for element in added_elements:
        if is_enclosure(element):
            annotations["enclosure"].append(record(element))
        elif is_connector(element):
            annotations["connector"].append(record(element))
        elif is_text(element):
            annotations["text"].append(record(element))
        elif is_glyph(element):
            annotations["glyph"].append(record(element))
        elif is_color(element):
            annotations["color"].append(record(element))
        elif is_indicator(element):
            annotations["indicator"].append(record(element))
        elif is_geometric(element):
            annotations["geometric"].append(record(element))

    return annotations
\end{codebox}
\label{lst:annotation_extraction}

We identify the seven annotation categories using rules based on element type, visibility, geometry, color, and position relative to chart regions or text.

\textbf{Enclosure.}
Enclosure refers to visual regions that surround, group, or emphasize chart content.
We extract highlighted regions, background areas, framed regions, and other visual boundaries.
To avoid confusing data marks with annotations, we exclude ordinary data representations and retain elements that provide additional emphasis beyond the original chart encoding.

\textbf{Connector.}
Connector captures visual links between annotations and their targets.
We extract arrows, leader lines, curved connectors, and guide lines associated with annotation elements.
Long lines spanning the full chart range are not treated as connectors because they usually represent reference structures rather than annotation links.

\textbf{Text.}
Text includes newly added semantic annotation text.
We extract visible text elements with non-empty content and record their text content, bounding regions, chart regions, and color.
For text annotations with attached connectors, we separate the text region from the connecting component to avoid counting them as a single element.
Text elements already present in the unannotated chart, such as axis labels and legends, are excluded by comparison with the unannotated baseline.

\textbf{Glyph.}
Glyph captures local symbols that mark specific data points or regions.
Examples include added markers, highlighted points, symbols, and emphasis marks.
The extractor avoids treating dense data marks as annotations and retains only sparse or visually distinctive added elements.

\textbf{Color.}
Color captures sparse color changes that carry annotation meaning.
We do not record every color used in the chart.
Instead, we identify colors that are newly introduced or selectively applied compared with the original chart.
Default colors, background colors, and ordinary data-series colors are filtered out when they do not function as annotation highlights.

\textbf{Indicator.}
Indicator refers to auxiliary structures that mark values, ranges, thresholds, or statistical references.
We extract reference lines, threshold markers, baselines, bracket-like structures, and other visual references.
Unlike connectors, indicators typically do not link a text label to a specific target but instead highlight values, intervals, or structural properties.

\textbf{Geometric.}
Geometric annotation captures structural components that change the visual organization of the chart.
This includes inset regions, zoom-related structures, and exploded chart components.
These elements are treated separately because they are not well represented by simple text, line, or region categories.

\subsection{Chart Fidelity}
\label{app:chart_fidelity}

\begin{codebox}{Listing 2: Chart Fidelity}
def chart_fidelity(unannotated_fig, predicted_fig):
    base_spec = extract_protected_chart_spec(unannotated_fig)
    pred_spec = extract_protected_chart_spec(predicted_fig)

    if not match_figure_ratio(base_spec, pred_spec):
        return 0

    if not match_chart_layout(base_spec, pred_spec):
        return 0

    if not preserve_data_marks(base_spec, pred_spec):
        return 0

    return 1
\end{codebox}

\subsection{Annotation Matching}
\label{app:annotation_matching}

\begin{codebox}{Listing 3: Annotation Matching}
def annotation_matching(gt_annotations, pred_annotations):
    total_intersection = 0
    total_union = 0

    for category in ANNOTATION_CATEGORIES:
        gt_items = gt_annotations[category]
        pred_items = pred_annotations[category]

        if category == "text":
            intersection, union = match_text_items(gt_items, pred_items)
        else:
            intersection = min(len(gt_items), len(pred_items))
            union = max(len(gt_items), len(pred_items))

        total_intersection += intersection
        total_union += union

    return safe_divide(total_intersection, total_union, empty_value=1.0)
\end{codebox}

\subsection{Color Matching}
\label{app:color_matching}

\begin{codebox}{Listing 4: Color Matching}
def color_matching(gt_annotations, pred_annotations):
    gt_colors = extract_colors_by_category(gt_annotations)
    pred_colors = extract_colors_by_category(pred_annotations)

    total_similarity = 0.0
    total_gt = 0
    total_pred = 0

    for category in ANNOTATION_CATEGORIES:
        gt_group = gt_colors[category]
        pred_group = pred_colors[category]

        pairs = maximum_bipartite_matching(
            gt_group,
            pred_group,
            score_fn=color_similarity,
        )

        total_similarity += sum_pair_scores(pairs)
        total_gt += len(gt_group)
        total_pred += len(pred_group)

    precision = safe_divide(total_similarity, total_pred, empty_value=1.0)
    recall = safe_divide(total_similarity, total_gt, empty_value=1.0)
    return f1_score(precision, recall)
\end{codebox}

\subsection{Text Relationship Statistics}
\label{app:text_spatial_statistics}

As described in Sec.~3.3 of the main paper, text-overlap and off-canvas statistics are provided to the LLM judge as quantitative evidence for visual clarity rather than as standalone scores.
For each newly added text element, we compute its maximum overlap ratio with other visible text elements.
The overlap ratio is measured relative to the area of the added text element, reflecting how much of the annotation is visually obstructed.
We also compute the off-canvas ratio of each added text element, measuring the portion of its bounding region that falls outside the figure canvas.

We further aggregate these quantities into global overlap and off-canvas statistics.
The global overlap statistic measures the total area affected by text conflicts relative to the figure canvas.
The global off-canvas statistic measures the total area of added text that falls outside the canvas.
These statistics help the judge identify text crowding, severe overlap, and off-canvas annotations, while the final score remains based on the full rendered chart.

\section{LLM-Judged Evaluation Rubric}
\label{app:evaluation_rubric}

This appendix provides the detailed scoring rubric used by the LLM-based judge (Sec.~3.3.2 of the main paper).
Beyond the integer scale from 1 to 5, with 0 for failed or missing cases, a score of 1 indicates a severe failure; 2 indicates clear problems below the baseline acceptable level; 3 indicates baseline acceptability; 4 indicates clearly above-baseline quality; and 5 indicates the highest quality.
The judge is instructed to score strictly, evaluate each metric independently, and choose the lower score when uncertain.

\subsection{Semantic Faithfulness}

\begin{table}[H]
\caption{Scoring rubric for Semantic Faithfulness.}
\centering
\small
\setlength{\tabcolsep}{4pt}
\renewcommand{\arraystretch}{1.15}
\begin{tabularx}{\columnwidth}{p{0.11\columnwidth}X}
\toprule
\textbf{Score} & \textbf{Criterion} \\
\midrule
0 & Required annotation is missing or cannot be judged. \\
1 & Severe semantic failure, including wrong target, incorrect trend/value/relation/conclusion, missing required text, factual deviation, misreference, or over-generation. \\
2 & Noticeable semantic problems; the main meaning is partly preserved but contains local errors, incomplete execution, inaccurate expression, or partial omission. \\
3 & Baseline acceptable; target, text, trend, value, relation, and conclusion are basically correct, with only minor imperfections. \\
4 & Above baseline; key targets, text, relations, scope, and structure are correct, with no substantive omission, misreference, or factual deviation. \\
5 & Fully satisfies all semantic requirements accurately and rigorously, consistent with the instruction and reference when applicable. \\
\bottomrule
\end{tabularx}
\label{tab:rubric_sf}
\end{table}

\subsection{Semantic Clarity}

\begin{table}[H]
\caption{Scoring rubric for Semantic Clarity.}
\centering
\small
\setlength{\tabcolsep}{4pt}
\renewcommand{\arraystretch}{1.15}
\begin{tabularx}{\columnwidth}{p{0.11\columnwidth}X}
\toprule
\textbf{Score} & \textbf{Criterion} \\
\midrule
0 & Required annotation is missing or cannot be judged. \\
1 & Severe clarity failure, including unclear referents, wrong or unstable text-object relations, misleading encoding, or conflicting interpretations. \\
2 & Noticeable ambiguity in target, relation, or meaning, requiring extra inference, pause, or confirmation from the reader. \\
3 & Baseline acceptable; referents are basically clear, the text-object relation is followable, and only minor local ambiguity remains. \\
4 & Above baseline; object, relation, and meaning are clear, stable, and non-misleading, requiring almost no extra confirmation. \\
5 & Meaning is immediately clear, with a natural and unambiguous relation between annotation text and visual objects. \\
\bottomrule
\end{tabularx}
\label{tab:rubric_sc}
\end{table}

\subsection{Visual Clarity}

\begin{table}[H]
\caption{Scoring rubric for Visual Clarity.}
\centering
\small
\setlength{\tabcolsep}{4pt}
\renewcommand{\arraystretch}{1.15}
\begin{tabularx}{\columnwidth}{p{0.11\columnwidth}X}
\toprule
\textbf{Score} & \textbf{Criterion} \\
\midrule
0 & Required annotation is missing or cannot be judged. \\
1 & Severe clarity failure, including overlap with essential chart text, occlusion of key data content, severe overlap, off-canvas or clipped text above 10\%, or moderate overlap above 50\%. \\
2 & Noticeable clarity problems, including crowding, moderate overlap, partial blocking, non-severe clipping/off-canvas issues, 20\%--50\% moderate overlap, or \texttt{anno\_overlap\_pct} $\geq$ 40\%. \\
3 & Baseline acceptable; the chart is basically readable, with only minor local crowding or readable overlap in non-critical areas. \\
4 & Above baseline; annotations and chart content are clearly separated, with no visible overlap, clipping, occlusion, or intrusion into important information. \\
5 & Reading is smooth, with almost no perceptible clarity issue and no intrusion into important chart information. \\
\bottomrule
\end{tabularx}
\label{tab:rubric_vc}
\end{table}
\subsection{Annotation Organization Quality}

\begin{table}[H]
\caption{Scoring rubric for Annotation Organization Quality.}
\centering
\small
\setlength{\tabcolsep}{4pt}
\renewcommand{\arraystretch}{1.15}
\begin{tabularx}{\columnwidth}{p{0.11\columnwidth}X}
\toprule
\textbf{Score} & \textbf{Criterion} \\
\midrule
0 & Required annotation is missing or cannot be judged. \\
1 & Severe organization failure, including chaotic placement, wrong grouping, severe target detachment, out-of-body annotations, colors that harm recognition, or more than 50\% of annotation groups containing four or more elements. \\
2 & Noticeable organization problems, including loose placement, weak target attachment, redundancy, overlap, imbalance, poor spacing, weak contrast, confusing color use, or more than 20\% of annotation groups containing four or more elements. \\
3 & Baseline acceptable; layout is generally reasonable, annotations are integrated with the chart body, and colors are visible and acceptable. \\
4 & Above baseline; grouping, target attachment, spacing, hierarchy, and colors are well coordinated, with no obvious redundancy or over-complexity. \\
5 & Excellent organization; composition is balanced, natural, easy to parse, and supported by effective color use. \\
\bottomrule
\end{tabularx}
\label{tab:rubric_org}
\end{table}

\subsection{Attention Guidance}

\begin{table}[H]
\caption{Scoring rubric for Attention Guidance.}
\centering
\small
\setlength{\tabcolsep}{4pt}
\renewcommand{\arraystretch}{1.15}
\begin{tabularx}{\columnwidth}{p{0.11\columnwidth}X}
\toprule
\textbf{Score} & \textbf{Criterion} \\
\midrule
0 & Required annotation is missing or cannot be judged. \\
1 & Severe guidance failure, including attention drawn to the wrong region, buried main target, or strongly misleading visual focus. \\
2 & Noticeable guidance problems, including weak salience, competing highlights, hierarchy confusion, unstable focus, or substantial search effort. \\
3 & Baseline acceptable; the main target can be found and the guidance basically works, though some browsing or brief pause may be needed. \\
4 & Above baseline; the main target is salient, hierarchy is clear and stable, and the focus can usually be grasped quickly. \\
5 & The main target is extremely clear, with efficient and natural guidance that can be identified almost immediately. \\
\bottomrule
\end{tabularx}
\label{tab:rubric_attn}
\end{table}

\section{LLM-Judged Evaluation Prompt}
\label{app:prompt_templates}

We use the same evaluation prompt template across all models and input settings to score \emph{Semantic Consistency} and \emph{Design Effectiveness}.
The scoring rubric is fixed.
Only the placeholder \texttt{\{stage\_constraints\}} is replaced according to the three instruction levels: Intent, Operation, and Implementation.

\subsection{Evaluation Prompt Template}

\begin{tcolorbox}[promptbox,title={Prompt for Semantic and Design Evaluation}]
\small

\textbf{Role.} You are an expert judge for chart annotation quality.

\medskip
\textbf{Task.} Evaluate only the added annotations in the AI-generated chart, not the base chart design.
Use the instruction, AI image, Text\_Relation\_Results, and GT image to judge whether the annotations are semantically correct, visually clear, well organized, and effective in guiding attention.
Use the GT image according to the instruction-level constraints specified below.
Text\_Relation\_Results provides text-overlap and off-canvas statistics as supporting evidence.

\medskip
\textbf{Scoring Policy.}
\begin{itemize}
    \item Use a 0--5 rating scale for each metric: 0 = completely missing or extremely poor; 1 = very poor; 2 = below the acceptable level; 3 = acceptable; 4 = clearly above acceptable; 5 = highest quality.
    \item Score strictly. If uncertain, choose the lower score.
    \item Score each metric independently.
    \item Return JSON only.
\end{itemize}

\medskip
\textbf{Semantic Consistency.} Judge the following two metrics based on the instruction and the generated annotations, using the GT image as a reference when needed.

\medskip
\textbf{1. Semantic Faithfulness (sf)}

\textbf{Description:} Evaluate whether the annotations follow the intended meaning of the instruction.

\textbf{Check:} target object, specified text, trend, value, relation, conclusion, missing required content, wrong reference, factual error, over-generation, wrong visual encoding, and misleading grouping.

\textbf{Level:}
\begin{itemize}
    \item \textbf{1:} Clear semantic failure, such as wrong target, wrong trend/value/relation/conclusion, missing key content, major factual error, wrong reference, over-generation, or wrong encoding.
    \item \textbf{2:} Noticeable semantic problems. The main meaning is partly preserved, but local errors, incomplete execution, inaccurate value/trend/relation/text, or partial omission remain.
    \item \textbf{3:} Acceptable. The target object and main meaning are basically correct, with only minor non-critical problems.
    \item \textbf{4:} Clearly above acceptable. Key targets, text, meaning, relations, scope, and structure are correct, with at most trivial problems.
    \item \textbf{5:} All semantic requirements of the instruction are fully and accurately satisfied, with correct targets, content, relations, scope, and visual encoding.
\end{itemize}

\medskip
\textbf{2. Semantic Clarity (sc)}

\textbf{Description:} Evaluate whether the relation between annotation text and visual objects is clear.

\textbf{Check:} clear referent, text-object relation, ambiguity, competing interpretations, and need for extra inference.

\textbf{Level:}
\begin{itemize}
    \item \textbf{1:} Clear failure, including unclear referents, wrong or unstable text-object relations, obvious ambiguity, misleading encoding, or competing interpretations.
    \item \textbf{2:} Noticeable problems, including locally unclear object, relation, or meaning that requires extra inference, pause, or confirmation.
    \item \textbf{3:} Acceptable. Referents are clear and the text-object relation is basically followable, with only minor ambiguity.
    \item \textbf{4:} Clearly above acceptable. Object, relation, and meaning are clear, stable, and not misleading.
    \item \textbf{5:} Meaning is immediately clear; the relation between text and visual objects is natural and easy to understand.
\end{itemize}

\medskip
\textbf{Design Effectiveness.}

\medskip
\textbf{3. Visual Clarity (vc)}

\textbf{Description:} Evaluate whether annotations interfere with reading the chart.
Focus on overlap, clipping, crowding, and blocking between annotations and chart content.
Text\_Relation\_Results provides evidence, but direct visual judgment from the image has priority when visible blocking is present.

\textbf{Check:} overlap or clipping among annotations; blocking between annotations and chart content; blocked titles, subtitles, axis labels, legends, tick labels, data marks, or key labels; and readability of key chart information.

\textbf{Level:}
\begin{itemize}
    \item \textbf{1:} Severe clarity failure, including severe overlap, intrusion into essential chart text, visible blocking of key chart content, obvious clipped/off-canvas text, or unreadable placement. Typical triggers include any severe overlap that makes text unreadable, off-canvas or clipped text $>$ 10\%, or moderate overlap $>$ 50\%.
    \item \textbf{2:} Noticeable clarity problems, including visible crowding, conflict, moderate overlap, partial blocking, competition with chart content, partial blocking of non-critical data marks, or non-severe clipping/off-canvas issues. Typical triggers include 20\%--50\% moderate overlap or \texttt{anno\_overlap\_pct} $\geq$ 40\%.
    \item \textbf{3:} Acceptable. The chart is basically readable, with only minor local crowding or mild readable overlap in non-critical areas; key chart text and data content remain unobstructed.
    \item \textbf{4:} Clearly above acceptable. Annotations and chart content are clearly separated, with no visible overlap, clipping, blocking, or intrusion into important chart information.
    \item \textbf{5:} Reading is smooth, with almost no visible clarity issue and no intrusion into important chart information.
\end{itemize}

\medskip
\textbf{4. Annotation Organization Quality (org)}

\textbf{Description:} Evaluate whether annotations are well placed and visually fit the chart.
Focus on placement, grouping, spacing, hierarchy, target attachment, and color use.
This metric penalizes poor organization even when the chart remains readable.

\textbf{Check:} placement, grouping, hierarchy, spacing, whitespace, target attachment, redundancy, disorder, overly complex annotation groups, color contrast, separability, coordination, and support for emphasis.

\textbf{Level:}
\begin{itemize}
    \item \textbf{1:} Clear organization failure, including chaotic placement, clearly wrong grouping, annotations largely outside the chart body, severe detachment from targets, or colors that seriously harm recognition or create annotation-data confusion. This includes cases where more than 50\% of annotation groups contain four or more elements.
    \item \textbf{2:} Clear organization or color problems, including loose or scattered placement, intrusion into title/subtitle/axis/legend areas, placement outside the main plot area or chart body, weak target attachment, redundancy, local imbalance, poor spacing, weak contrast, poor separability, or confusing color use. This includes cases where more than 20\% of annotation groups contain four or more elements.
    \item \textbf{3:} Acceptable. The layout is generally reasonable; annotations are within or visually integrated with the main chart body, close enough to targets, readable, and acceptable in color use.
    \item \textbf{4:} Clearly above acceptable. Grouping, target attachment, spacing, hierarchy, and colors are well coordinated, with no intrusion into essential chart-text areas.
    \item \textbf{5:} Excellent organization. The composition is balanced, natural, and easy to parse, with effective color use and no intrusion into essential chart-text areas.
\end{itemize}

\medskip
\textbf{5. Attention Guidance (attn)}

\textbf{Description:} Evaluate whether the annotations highlight the intended target or target set and create a clear visual focus.

\textbf{Check:} target salience, hierarchy clarity and stability, competing highlights or misleading focus, and search effort needed to find the focus.

\textbf{Level:}
\begin{itemize}
    \item \textbf{1:} Clear failure, including attention drawn to the wrong place, main target severely buried, or strongly misleading focus.
    \item \textbf{2:} Noticeable problems, including weak salience, competing highlights, hierarchy confusion, unstable focus, or substantial search effort.
    \item \textbf{3:} Acceptable. The main target can be found and the guidance basically works, but salience and hierarchy are ordinary; brief search may be needed.
    \item \textbf{4:} Clearly above acceptable. The main target is fairly salient, hierarchy is clear and stable, and the focus can usually be grasped quickly.
    \item \textbf{5:} The main target is extremely clear; visual guidance is efficient and natural, and the focus can be identified almost immediately.
\end{itemize}

\medskip
\textbf{Constraints.}

\texttt{\{stage\_constraints\}}

\medskip
\textbf{Output Format.} Return exactly one JSON object using the following schema, where each metric value is an integer score from 0 to 5. Do not output any extra text.

\begin{verbatim}
{
  "results": [
    {
      "sf": 0,
      "sc": 0,
      "vc": 0,
      "org": 0,
      "attn": 0
    }
  ]
}
\end{verbatim}

\medskip
\textbf{Inputs.}
\begin{itemize}
    \item Instruction: \texttt{\{instruction\}}
    \item Text\_Relation\_Results: \texttt{\{text\_relation\_results\}}
    \item AI image: the chart image generated by the evaluated model
    \item GT image: the reference annotated chart image
\end{itemize}

\end{tcolorbox}

\subsection{Instruction-Level Constraint Instantiations}

The placeholder \texttt{\{stage\_constraints\}} is instantiated according to the instruction level.
All metric scores must be integers in \([0,5]\).
Do not evaluate visual similarity to the GT image. Alternative annotation types, layouts, colors, and placements should not be penalized solely for differing from the GT image, provided that they faithfully satisfy the instruction and are semantically clear and visually effective.
For Operation- and Implementation-level instructions, the GT image is used as a comparison reference, and a substantial mismatch with the GT image caps the corresponding metric at 3.

\section{Generation Prompts and Examples}
\label{app:prompts_examples}

\subsection{Generation Prompt Templates}
\label{app:generation_prompt_templates}

We build each prompt from five parts: a mode-specific role description, common constraints, the unannotated chart code, an instruction-level description, and the annotation instruction.
The common constraints and final request are fixed across all models, instruction levels, and input settings.

\textbf{Mode-specific role descriptions.}
We use different role descriptions for the two input settings.

\begin{tcolorbox}[promptbox,title={Code Input Role}]
\small
You are an expert in chart annotation and Python visualization.
I have created a figure but have not added any annotations yet.
I will provide you with the chart code without annotations, along with annotation instructions.
Your task is to modify the provided code to add annotations based on the instructions.
\end{tcolorbox}

\begin{tcolorbox}[promptbox,title={Code + Image Input Role}]
\small
You are an expert in chart annotation and Python visualization.
I have created a figure but have not added any annotations yet.
I will provide you with the chart code without annotations, along with its corresponding reference image and annotation instructions.
The image is generated by the code and is provided to help with positioning and alignment.
Your task is to modify the provided code to add annotations based on the instructions, using the image for guidance.
\end{tcolorbox}

\textbf{Common constraints and code input.}
After the role description, all prompts include the same constraints.
These constraints require executable, self-contained Python code and discourage unnecessary changes to the base chart.
For multi-plot figures, we also specify the subplot indexing order.
The unannotated chart code is then inserted into the shared template.

\begin{tcolorbox}[promptbox,title={Common Constraints}]
\small
You must follow these constraints:
\begin{enumerate}
    \item The code must be self-contained and executable.
    \item Use Python, and do not change the style of the base chart.
    \item Output code only. Do not include comments or explanations.
\end{enumerate}
\end{tcolorbox}

\begin{tcolorbox}[constraintbox,title={Multi-plot Constraint}]
\small
Layout Constraint for Multi-plots: Subplots are indexed from 0, ordered left to right and top to bottom in row-major order (0, 1, 2...).
\end{tcolorbox}

\begin{tcolorbox}[promptbox,title={Shared Code Block}]
\small
Here is the Python code without annotations:

\texttt{\{code\}}
\end{tcolorbox}

\textbf{Instruction-level descriptions.}
We add a short description before each annotation instruction to specify the expected level of detail.

\begin{tcolorbox}[constraintbox,title={Intent-level Description}]
\small
I will describe the communication goal and the key information to be conveyed, without specifying how the visualization should look.
You should interpret this intent, choose an appropriate annotation design, and generate Matplotlib code accordingly.
\end{tcolorbox}

\begin{tcolorbox}[constraintbox,title={Operation-level Description}]
\small
I will describe the annotation actions and spatial relations using qualitative instructions, such as relative positions, layout relations, and non-numeric style descriptions.
You should convert these instructions into Matplotlib code.
\end{tcolorbox}

\begin{tcolorbox}[constraintbox,title={Implementation-level Description}]
\small
I will provide concrete implementation details, including coordinates, colors, sizes, and other rendering parameters.
You should follow these specifications precisely and produce Matplotlib code that implements them.
\end{tcolorbox}

\textbf{Instruction and final request.}
Finally, we insert the instance-specific annotation instruction and ask the model to return complete Python Matplotlib code.

\begin{tcolorbox}[promptbox,title={Instruction and Final Request}]
\small
\texttt{\{level\_description\}}

Instruction:

\texttt{\{instruction\}}

Based on my instruction above, please write the complete Python Matplotlib code for me.
\end{tcolorbox}

\subsection{Prompt and Evaluation Example}
\label{app:prompt_eval_example}

\begin{figure*}[!htbp]
    \centering
    \includegraphics[width=\textwidth]{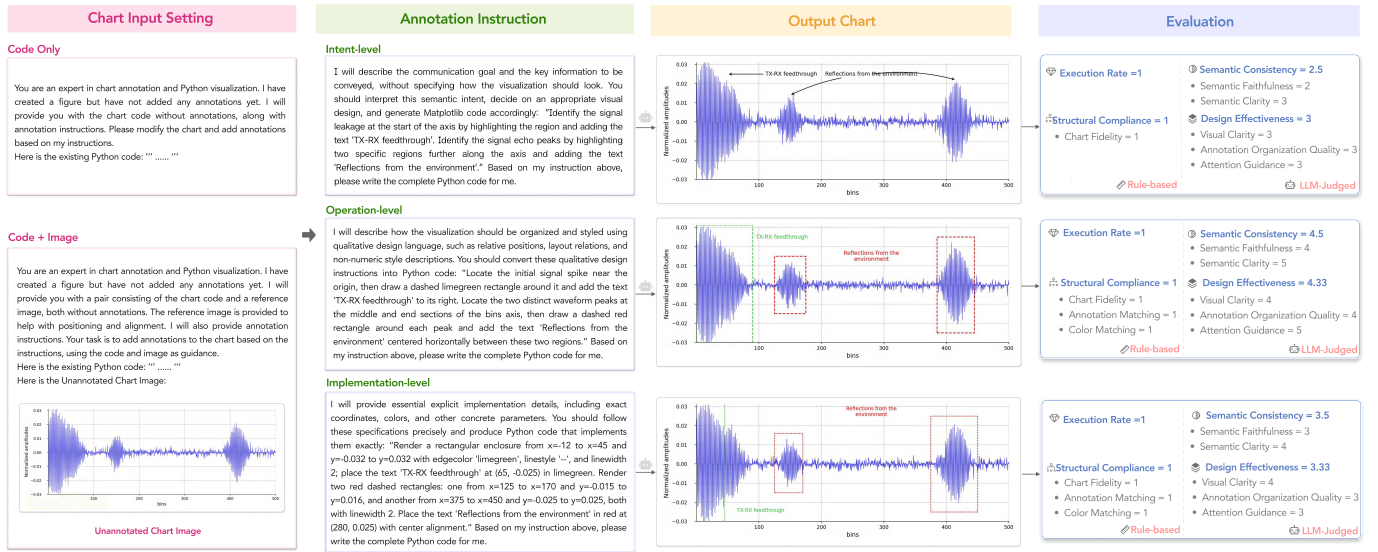}
    \caption{Prompt and evaluation example for one chart instance using Gemini 3.1 Pro Preview.}
    \label{fig:prompt_eval_example}
\end{figure*}

Fig.~\ref{fig:prompt_eval_example} shows one full generation and evaluation example using Gemini 3.1 Pro Preview.
The example includes the two input settings, the shared prompt structure, the three instruction levels, generated charts, the ground-truth annotated chart, and the corresponding rule-based and LLM-judged results.
It also shows how the instruction becomes more specific from Intent to Operation and Implementation.

\section{Model Configurations and Licenses}
\label{app:model_config_license}

\subsection{Model Versions and Inference Settings}
\label{app:model_config}

Tables~\ref{tab:model_versions} and~\ref{tab:model_configs} summarize the model versions and decoding settings.
Proprietary models are identified by their release versions, and open-source models are pinned to the Hugging Face checkpoints listed in Tab.~\ref{tab:model_versions}, which are used throughout all evaluations.
Each row of Tab.~\ref{tab:model_configs} corresponds to the same model as Tab.~\ref{tab:model_versions}, and entries marked with ``--'' indicate settings that are unavailable or left at their defaults as described below.
The maximum output length is set to 16,384 tokens, except for Gemini models, for which we use the default output-length setting to avoid API-side truncation behavior.

For API-based models, \texttt{do\_sample} is not exposed; for
locally deployed models, we set \texttt{do\_sample=False}.
Claude Sonnet 4.6 does not support setting temperature and top-$p$ simultaneously, so only temperature is specified.
We do not explicitly control reasoning-related settings.
Gemini models enable reasoning by default, while GPT-5.4 and Claude Sonnet 4.6 are evaluated under their default reasoning-related settings.

\begin{table}[H]
\caption{Model versions and Hugging Face checkpoints.}
\centering
\small
\setlength{\tabcolsep}{4pt}
\begin{tabular}{ll}
\toprule
\textbf{Model} & \textbf{Version / HF Checkpoint} \\
\midrule
\multicolumn{2}{c}{\textit{Proprietary Models}} \\
\midrule
GPT-5.4 & gpt-5.4 \\
Gemini 3.1 Pro Preview & gemini-3.1-pro-preview \\
Gemini 3 Flash Preview & gemini-3-flash-preview \\
Claude Sonnet 4.6 & claude-sonnet-4-6 \\
\midrule
\multicolumn{2}{c}{\textit{Open-Source Models}} \\
\midrule
Kimi K2.5 & moonshotai/Kimi-K2.5 \\
Gemma 4 31B & google/gemma-4-31B \\
Qwen3.5-397B-A17B & Qwen/Qwen3.5-397B-A17B \\
Qwen3.5-122B-A10B & Qwen/Qwen3.5-122B-A10B \\
Qwen3.5-27B & Qwen/Qwen3.5-27B \\
Qwen3.5-9B & Qwen/Qwen3.5-9B \\
\bottomrule
\end{tabular}
\label{tab:model_versions}
\end{table}

\begin{table}[H]
\caption{Inference settings for all models. Unset values indicate default or unavailable settings.}
\centering
\small
\setlength{\tabcolsep}{4pt}
\resizebox{\columnwidth}{!}{
\begin{tabular}{lcccc}
\toprule
\textbf{Model} 
& \textbf{Do Sample} 
& \textbf{Max Tokens} 
& \textbf{Temp.} 
& \textbf{Top-$p$} \\
\midrule
\multicolumn{5}{c}{\textit{Proprietary Models}} \\
\midrule
GPT-5.4 & -- & 16384 & 0 & 1 \\
Gemini 3.1 Pro Preview & -- & Default & 0 & 1 \\
Gemini 3 Flash Preview & -- & Default & 0 & 1 \\
Claude Sonnet 4.6 & -- & 16384 & 0 & -- \\
\midrule
\multicolumn{5}{c}{\textit{Open-Source Models}} \\
\midrule
Kimi K2.5 & -- & 16384 & -- & -- \\
Gemma 4 31B & False & 16384 & 0 & 1 \\
Qwen3.5-397B-A17B & False & 16384 & 0 & 1 \\
Qwen3.5-122B-A10B & False & 16384 & 0 & 1 \\
Qwen3.5-27B & False & 16384 & 0 & 1 \\
Qwen3.5-9B & False & 16384 & 0 & 1 \\
\bottomrule
\end{tabular}
}
\label{tab:model_configs}
\end{table}

\subsection{Computational Infrastructure and Budget}
\label{app:computational_budget}

The local evaluation uses approximately 450 GPU hours.

\subsection{Model Licenses}
\label{app:model_license}

Proprietary models and Kimi K2.5 are accessed through their official APIs and used under the corresponding API terms, while all locally deployed open-source models (Gemma 4 31B and the Qwen3.5 series) are released under the Apache 2.0 license for both model weights and code.

\section{Chart Image Input Analysis}
\label{app:image_input_analysis}

This section provides additional evidence for the effect of chart image input.
We first report overall significance tests comparing Code + Image Input with Code Input, and then break down the normalized gain in \emph{Design Effectiveness} into its three components.

\subsection{Overall Significance Tests}
\label{app:image_input_significance}

We test whether Code + Image Input improves over Code Input using all matched instances across the ten models.
For \emph{Execution Rate}, we use a one-sided exact McNemar test.
For \emph{Structural Compliance}, \emph{Semantic Consistency}, and \emph{Design Effectiveness}, we use one-sided paired $t$-tests.

Tab.~\ref{tab:image_input_overall_significance} reports the overall results.

\begin{table}[H]
\caption{Overall score changes from Code Input to Code + Image Input.}
\centering
\small
\setlength{\tabcolsep}{2.5pt}
\renewcommand{\arraystretch}{1.12}
\begin{tabularx}{\columnwidth}{>{\raggedright\arraybackslash}Xrrrrr}
\toprule
\textbf{Metric} & \textbf{Code} & \textbf{Code+Image} & \textbf{$\Delta$} & \textbf{$p$} & \textbf{Effect Size} \\
\midrule
Execution Rate
& 0.941 & 0.950 & 0.009 & $<.001$ & $g=0.080$ \\
Structural Compliance
& 0.817 & 0.822 & 0.005 & $<.001$ & $d_z=0.022$ \\
Semantic Consistency
& 3.369 & 3.374 & 0.005 & .209 & $d_z=0.004$ \\
Design Effectiveness
& 3.418 & 3.439 & 0.021 & $<.001$ & $d_z=0.021$ \\
\bottomrule
\end{tabularx}
\label{tab:image_input_overall_significance}
\end{table}

In Tab.~\ref{tab:image_input_overall_significance}, $\Delta$ denotes the mean score change from Code Input to Code + Image Input.
$^{*}$, $^{**}$, and $^{***}$ indicate $p<0.05$, $p<0.01$, and $p<0.001$, respectively.

\subsection{Design Component Gain Analysis}
\label{app:design_component_gain}

\begin{figure}[H]
    \centering
    \includegraphics[width=\columnwidth]{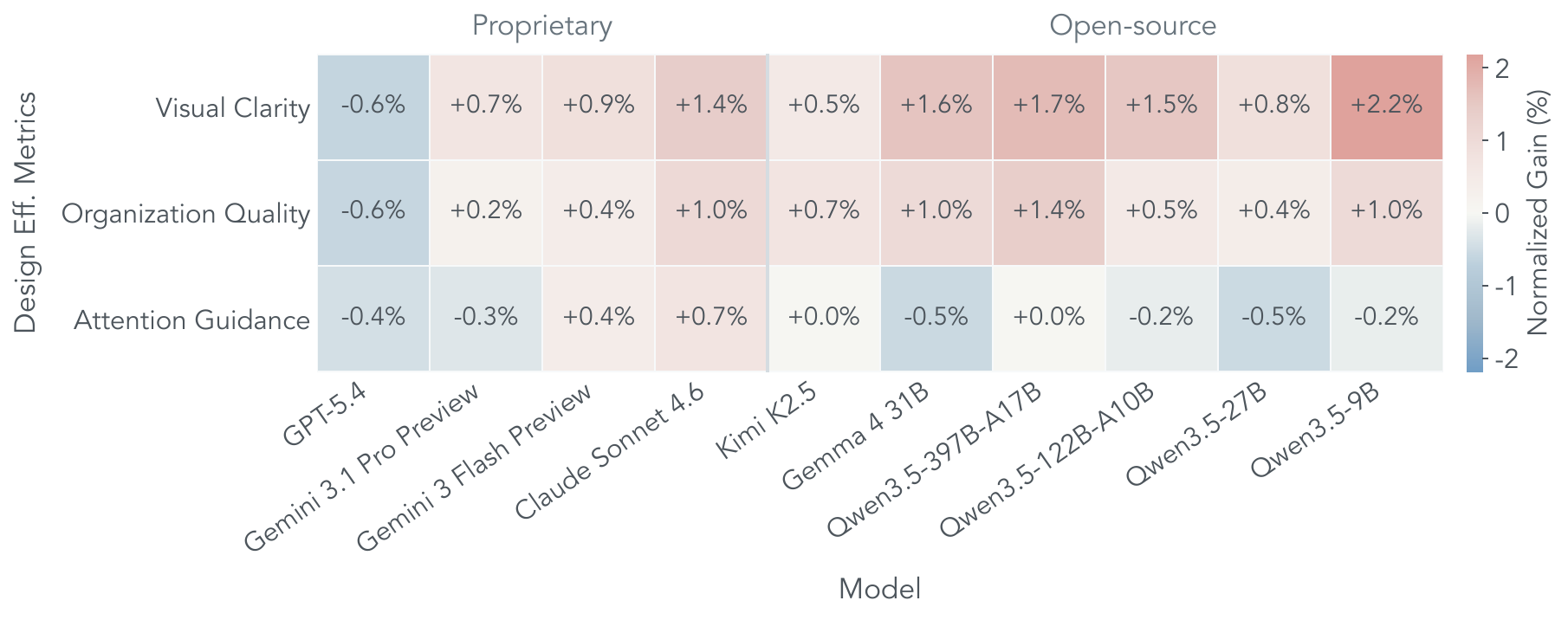}
    \caption{Normalized gains from chart image input on \emph{Design Effectiveness} components. Positive values indicate improvements over Code Input.}
    \label{fig:design_normalized_gain_heatmap}
\end{figure}

Fig.~\ref{fig:design_normalized_gain_heatmap} breaks down the normalized gains from chart image input into the three components of \emph{Design Effectiveness}.
\emph{Visual Clarity} shows the largest and most consistent gains across models, while \emph{Annotation Organization Quality} also improves in many cases.
In contrast, \emph{Attention Guidance} shows limited or negative gains for most models, indicating that chart images help more with local layout and readability than with guiding attention to the intended insight.
The gains are also model-dependent: GPT-5.4 shows negative gains across all three components, whereas several open-source models, especially Qwen variants, benefit more from image input, mainly in \emph{Visual Clarity}.

\section{Image-Only Ablation}
\label{app:image_only_ablation}
\subsection{Image-only Performance}

Tab.~\ref{tab:image_only_results} reports the results under the Image-only setting across three instruction levels.

\begin{table*}[!htbp]
\caption{
Results under the Image-only Input setting across three instruction levels.
Exec., Struct., Sem., and Design denote \emph{Execution Rate}, \emph{Structural Compliance}, \emph{Semantic Consistency}, and \emph{Design Effectiveness}.
\protect\colorbox{gray!12}{Gray} Struct.$^\ast$ columns report only Chart Fidelity for Intent-level instructions and are not directly comparable with \emph{Structural Compliance} at the other levels.
}
\centering
\small
\setlength{\tabcolsep}{3pt}
\resizebox{0.8\textwidth}{!}{
\begin{tabular}{l|c>{\columncolor{gray!12}}ccc|cccc|cccc}
\toprule
Model
& \multicolumn{4}{c|}{Intent-level}
& \multicolumn{4}{c|}{Operation-level}
& \multicolumn{4}{c}{Implementation-level} \\
\cmidrule(lr){2-5}
\cmidrule(lr){6-9}
\cmidrule(lr){10-13}
&
Exec. & Struct.$^\ast$ & Sem. & Design
& Exec. & Struct. & Sem. & Design
& Exec. & Struct. & Sem. & Design \\
\midrule

\rowcolor{groupgreen}
\multicolumn{13}{c}{\textit{Proprietary Models}} \\
\midrule

GPT-5.4
&0.976&0.140&3.000&3.127
&0.965&0.360&2.796&3.175
&0.952&0.395&2.938&3.307\\

Gemini 3.1 Pro Preview
&0.947&0.137&3.080&3.239
&0.941&0.375&2.874&3.273
&0.938&0.409&2.958&3.373\\

Gemini 3 Flash Preview
&0.953&0.113&3.062&3.203
&0.957&0.372&2.882&3.292
&0.955&0.410&3.070&3.469\\

Claude Sonnet 4.6
&0.957&0.053&2.897&3.149
&0.945&0.308&2.604&3.131
&0.950&0.349&2.683&3.211\\

\midrule

\rowcolor{groupgreen}
\multicolumn{13}{c}{\textit{Open-Source Models}} \\
\midrule

Kimi K2.5
&0.942&0.104&2.695&3.028
&0.936&0.337&2.433&3.048
&0.928&0.366&2.548&3.114\\

Gemma 4 31B
&0.860&0.041&2.373&2.631
&0.857&0.292&2.165&2.658
&0.838&0.309&2.188&2.663\\

Qwen3.5-397B-A17B
&0.821&0.053&2.220&2.415
&0.807&0.269&1.970&2.399
&0.820&0.300&2.019&2.474\\

Qwen3.5-122B-A10B
&0.781&0.052&1.998&2.230
&0.764&0.249&1.738&2.151
&0.764&0.279&1.840&2.261\\

Qwen3.5-27B
&0.700&0.055&1.813&2.017
&0.676&0.211&1.593&1.984
&0.677&0.224&1.660&2.032\\

Qwen3.5-9B
&0.646&0.018&1.393&1.663
&0.578&0.167&1.164&1.501
&0.611&0.197&1.287&1.706\\

\bottomrule
\end{tabular}
}
\label{tab:image_only_results}
\end{table*}

To further investigate the contribution of different input modalities, we evaluate an Image-only setting under the same experimental protocol.
We evaluate all 10 models on the full dataset.

Compared with the Code Input setting in Tab.~3 of the main paper, Image-only performance is substantially lower, especially on semantic and design metrics.
Although the relative ranking remains broadly similar, generating annotations from images alone requires additional capabilities beyond annotation generation, including visual chart understanding, recovery of underlying data and structural information, and reconstruction of the original chart.

Among proprietary models, Gemini 3 Flash Preview achieves competitive performance on several Operation- and Implementation-level metrics, while Kimi K2.5 consistently performs best among open-source models.

\subsection{Contribution of Chart Code}

Tab.~\ref{tab:code_gain_over_image_only} reports the absolute improvements introduced by adding chart code.

\begin{table*}[t]
\caption{
Absolute improvements of Code + Image over Image-only.
Positive values indicate additional benefits from providing chart code beyond rendered chart images.
\protect\colorbox{gray!12}{Gray} Struct.$^\ast$ columns report Chart Fidelity gains only for Intent-level instructions.
}
\centering
\small
\setlength{\tabcolsep}{3pt}
\resizebox{0.8\textwidth}{!}{
\begin{tabular}{l|c>{\columncolor{gray!12}}ccc|cccc|cccc}
\toprule
Model
& \multicolumn{4}{c|}{Intent-level}
& \multicolumn{4}{c|}{Operation-level}
& \multicolumn{4}{c}{Implementation-level}\\
\cmidrule(lr){2-5}
\cmidrule(lr){6-9}
\cmidrule(lr){10-13}
&
Exec.&Struct.$^\ast$&Sem.&Design
&
Exec.&Struct.&Sem.&Design
&
Exec.&Struct.&Sem.&Design\\
\midrule

\rowcolor{groupgreen}
\multicolumn{13}{c}{\textit{Proprietary Models}}\\
\midrule

GPT-5.4
&+0.011&+0.716&+0.553&+0.294
&+0.023&+0.481&+0.807&+0.522
&+0.041&+0.502&+1.119&+0.830\\

Gemini 3.1 Pro Preview
&+0.048&+0.780&+0.600&+0.400
&+0.052&+0.501&+0.906&+0.637
&+0.060&+0.507&+1.191&+0.857\\

Gemini 3 Flash Preview
&+0.044&+0.583&+0.538&+0.393
&+0.031&+0.443&+0.761&+0.524
&+0.037&+0.476&+0.936&+0.675\\

Claude Sonnet 4.6
&+0.027&+0.833&+0.615&+0.272
&+0.029&+0.535&+0.894&+0.515
&+0.038&+0.557&+1.337&+0.901\\

\midrule

\textit{Proprietary Average}
&+0.033&+0.728&+0.576&+0.340
&+0.034&+0.490&+0.842&+0.549
&+0.044&+0.510&+1.146&+0.816\\

\midrule
\rowcolor{groupgreen}
\multicolumn{13}{c}{\textit{Open-Source Models}}\\
\midrule

Kimi K2.5
&+0.041&+0.773&+0.615&+0.252
&+0.029&+0.481&+0.900&+0.407
&+0.044&+0.512&+1.330&+0.847\\

Gemma 4 31B
&+0.099&+0.746&+0.786&+0.476
&+0.083&+0.504&+1.055&+0.676
&+0.100&+0.532&+1.493&+1.115\\

Qwen3.5-397B-A17B
&+0.132&+0.780&+0.815&+0.574
&+0.138&+0.512&+1.081&+0.802
&+0.125&+0.538&+1.637&+1.296\\

Qwen3.5-122B-A10B
&+0.148&+0.756&+0.846&+0.593
&+0.154&+0.519&+1.147&+0.861
&+0.157&+0.540&+1.618&+1.338\\

Qwen3.5-27B
&+0.239&+0.773&+1.147&+0.883
&+0.218&+0.535&+1.244&+0.976
&+0.237&+0.597&+1.780&+1.518\\

Qwen3.5-9B
&+0.219&+0.714&+0.892&+0.667
&+0.233&+0.472&+1.074&+0.865
&+0.221&+0.518&+1.554&+1.285\\

\midrule

\textit{Open-Source Average}
&+0.146&+0.757&+0.850&+0.574
&+0.143&+0.504&+1.083&+0.764
&+0.147&+0.539&+1.568&+1.233\\

\midrule

\textbf{Overall Average}
&\textbf{+0.101}&\textbf{+0.745}&\textbf{+0.741}&\textbf{+0.480}
&\textbf{+0.099}&\textbf{+0.498}&\textbf{+0.987}&\textbf{+0.678}
&\textbf{+0.106}&\textbf{+0.528}&\textbf{+1.399}&\textbf{+1.066}\\

\bottomrule
\end{tabular}
}
\label{tab:code_gain_over_image_only}
\end{table*}

\section{Complexity Analysis Details}
\label{app:complexity_details}

\subsection{Chart Complexity Indicators}
\label{app:complexity_definitions}

We define six indicators to analyze how model performance changes with chart and annotation complexity.
The main paper (Sec.~4.2.2) overviews these indicators and splits charts into simple, medium, and complex groups for each; here we detail how each indicator is computed.
These indicators are used only for the complexity analysis and are not part of the main evaluation metrics.

\textbf{Code Token Increment.}
Code Token Increment measures the extra code needed to add annotations.
Following the tokenization procedure in Sec.~3.2 of the main paper, we compute code length with the Llama2 tokenizer and measure the difference between annotated and unannotated chart code.
A larger value means that the annotations require more code changes.

\textbf{Instruction Length.}
Instruction Length measures the average word count of the three instruction levels for each chart.
A larger value means that the chart is paired with longer and more detailed annotation instructions.

\textbf{Annotation Type Count.}
Annotation Type Count measures the number of distinct annotation element types in a chart.
It is computed from the structured annotation elements described in Sec.~\ref{app:annotation_extraction}.
This indicator reflects the variety of annotation forms rather than the total number of annotation elements.

\textbf{Annotation Count.}
Annotation Count measures the total number of added annotation elements.
It is also computed from the structured annotation elements described in Sec.~\ref{app:annotation_extraction}.
Unlike Annotation Type Count, this indicator captures annotation quantity rather than type variety.

\textbf{Annotation Spatial Distribution Entropy.}
Annotation Spatial Distribution Entropy measures how widely annotations are spread over the chart canvas.
We compute it from the pixel-level difference between the annotated chart image and the unannotated chart image.
We first mark pixels whose RGB difference exceeds a fixed threshold as annotation pixels.
We then divide the canvas into a \(4 \times 4\) grid and count annotation pixels in each cell.
Let \(p_i\) denote the proportion of annotation pixels in the \(i\)-th grid cell.
The normalized entropy is:
\[
H_{\mathrm{anno}} =
\frac{
-\sum_{i=1}^{K} p_i \log p_i
}{
\log K
},
\]
where \(K=16\).
A higher value means that annotations are spread across more regions of the chart.

\textbf{Visual Element Occupancy.}
Visual Element Occupancy measures the proportion of the chart canvas occupied by non-background visual content.
We compute it from the unannotated chart image.
For each image, we estimate the background color using the median RGB value of border pixels.
Pixels close to the estimated background color are treated as background pixels.
We compute the whitespace ratio \(R_{\mathrm{white}}\) and define:
\[
R_{\mathrm{occ}} = 1 - R_{\mathrm{white}}.
\]
A larger value indicates that chart content occupies more of the canvas, leaving less empty space for annotations.

\subsection{Complexity Regression Analysis}
\label{app:complexity_regression}

The main paper reports relative score drops from simple to complex cases.
Here, we further run regression analysis to test whether these trends remain after controlling for chart category, input setting, and instruction level.

For each model, metric, and complexity indicator, we fit a categorical regression with the simple group as the reference:
\[
y =
\beta_0
+
\beta_1 \mathbb{I}(\mathrm{medium})
+
\beta_2 \mathbb{I}(\mathrm{complex})
+
\gamma^\top Z
+
\epsilon,
\]
where \(y\) is the metric score, and \(Z\) includes chart category, input setting, and instruction level.
We report \(\beta_2\) as the complex-vs-simple effect.
A negative \(\beta_2\) means that complex cases receive lower scores than simple cases after controlling for these factors.
Significance levels are denoted by \(^{*}p<0.05\), \(^{**}p<0.01\), and \(^{***}p<0.001\).

We run separate regressions for Execution Rate, Structural Compliance, Semantic Consistency, and Design Effectiveness.
The coefficients should be interpreted within each metric, because the metrics have different scales.
Tables~\ref{tab:reg_token}--\ref{tab:reg_occupancy} report the complex-vs-simple coefficients for each complexity indicator.

The regression results are consistent with the main analysis.
Code Token Increment yields the strongest and most consistently significant negative coefficients, with most entries reaching the $p<0.001$ level.
Instruction Length and Annotation Type Count likewise show clear negative coefficients, with the largest magnitudes on Semantic Consistency and Design Effectiveness.
Annotation Type Count has stronger effects than Annotation Count, suggesting that the variety of annotation forms is harder than the number of annotation elements alone.
Visual Element Occupancy shows weaker and less stable effects.
The Execution Rate coefficients are small in absolute terms, reflecting that execution failures are less sensitive to complexity than semantic and design quality.
Open-source models also receive systematically larger negative coefficients than proprietary models across most indicators.

\begin{table}[H]
\caption{Complex-vs-simple coefficients for Code Token Increment.}
\centering
\small
\setlength{\tabcolsep}{2pt}
\renewcommand{\arraystretch}{1.12}
\resizebox{\columnwidth}{!}{
\begin{tabular}{lrrrr}
\toprule
\textbf{Model} & \textbf{Exec.} & \textbf{Struct.} & \textbf{Sem.} & \textbf{Design} \\
\midrule
\multicolumn{5}{l}{\textit{Proprietary models}} \\
GPT-5.4 & -0.002 & -0.068$^{***}$ & -1.096$^{***}$ & -0.995$^{***}$ \\
Gemini 3.1 Pro Preview & -0.005$^{*}$ & -0.075$^{***}$ & -1.022$^{***}$ & -0.869$^{***}$ \\
Gemini 3 Flash Preview & -0.010$^{**}$ & -0.147$^{***}$ & -1.018$^{***}$ & -0.860$^{***}$ \\
Claude Sonnet 4.6 & -0.024$^{***}$ & -0.085$^{***}$ & -1.140$^{***}$ & -1.022$^{***}$ \\
\midrule
\multicolumn{5}{l}{\textit{Open-source models}} \\
Kimi K2.5 & -0.030$^{***}$ & -0.105$^{***}$ & -1.297$^{***}$ & -1.198$^{***}$ \\
Gemma 4 31B & -0.064$^{***}$ & -0.163$^{***}$ & -1.466$^{***}$ & -1.388$^{***}$ \\
Qwen3.5-397B-A17B & -0.036$^{***}$ & -0.126$^{***}$ & -1.349$^{***}$ & -1.302$^{***}$ \\
Qwen3.5-122B-A10B & -0.078$^{***}$ & -0.153$^{***}$ & -1.481$^{***}$ & -1.470$^{***}$ \\
Qwen3.5-27B & -0.075$^{***}$ & -0.157$^{***}$ & -1.483$^{***}$ & -1.455$^{***}$ \\
Qwen3.5-9B & -0.141$^{***}$ & -0.221$^{***}$ & -1.626$^{***}$ & -1.667$^{***}$ \\
\bottomrule
\end{tabular}
}
\label{tab:reg_token}
\end{table}

\begin{table}[H]
\caption{Complex-vs-simple coefficients for Instruction Length.}
\centering
\small
\setlength{\tabcolsep}{2pt}
\renewcommand{\arraystretch}{1.12}
\resizebox{\columnwidth}{!}{
\begin{tabular}{lrrrr}
\toprule
\textbf{Model} & \textbf{Exec.} & \textbf{Struct.} & \textbf{Sem.} & \textbf{Design} \\
\midrule
\multicolumn{5}{l}{\textit{Proprietary models}} \\
GPT-5.4 & -0.002 & -0.053$^{***}$ & -0.834$^{***}$ & -0.864$^{***}$ \\
Gemini 3.1 Pro Preview & -0.005 & -0.049$^{***}$ & -0.701$^{***}$ & -0.684$^{***}$ \\
Gemini 3 Flash Preview & -0.008$^{*}$ & -0.131$^{***}$ & -0.744$^{***}$ & -0.724$^{***}$ \\
Claude Sonnet 4.6 & -0.008 & -0.045$^{***}$ & -0.834$^{***}$ & -0.842$^{***}$ \\
\midrule
\multicolumn{5}{l}{\textit{Open-source models}} \\
Kimi K2.5 & -0.023$^{***}$ & -0.073$^{***}$ & -0.962$^{***}$ & -1.035$^{***}$ \\
Gemma 4 31B & -0.036$^{***}$ & -0.111$^{***}$ & -1.027$^{***}$ & -1.137$^{***}$ \\
Qwen3.5-397B-A17B & -0.028$^{***}$ & -0.080$^{***}$ & -1.019$^{***}$ & -1.145$^{***}$ \\
Qwen3.5-122B-A10B & -0.071$^{***}$ & -0.113$^{***}$ & -1.173$^{***}$ & -1.346$^{***}$ \\
Qwen3.5-27B & -0.059$^{***}$ & -0.111$^{***}$ & -1.152$^{***}$ & -1.293$^{***}$ \\
Qwen3.5-9B & -0.113$^{***}$ & -0.152$^{***}$ & -1.319$^{***}$ & -1.481$^{***}$ \\
\bottomrule
\end{tabular}
}
\label{tab:reg_instruction_length}
\end{table}

\begin{table}[H]
\caption{Complex-vs-simple coefficients for Annotation Type Count.}
\centering
\small
\setlength{\tabcolsep}{2pt}
\renewcommand{\arraystretch}{1.12}
\resizebox{\columnwidth}{!}{
\begin{tabular}{lrrrr}
\toprule
\textbf{Model} & \textbf{Exec.} & \textbf{Struct.} & \textbf{Sem.} & \textbf{Design} \\
\midrule
\multicolumn{5}{l}{\textit{Proprietary models}} \\
GPT-5.4 & -0.002 & -0.066$^{***}$ & -0.870$^{***}$ & -0.692$^{***}$ \\
Gemini 3.1 Pro Preview & -0.005 & -0.058$^{***}$ & -0.847$^{***}$ & -0.641$^{***}$ \\
Gemini 3 Flash Preview & -0.019$^{***}$ & -0.111$^{***}$ & -0.936$^{***}$ & -0.729$^{***}$ \\
Claude Sonnet 4.6 & -0.017$^{**}$ & -0.057$^{***}$ & -0.918$^{***}$ & -0.729$^{***}$ \\
\midrule
\multicolumn{5}{l}{\textit{Open-source models}} \\
Kimi K2.5 & -0.040$^{***}$ & -0.101$^{***}$ & -1.031$^{***}$ & -0.905$^{***}$ \\
Gemma 4 31B & -0.066$^{***}$ & -0.139$^{***}$ & -1.147$^{***}$ & -1.004$^{***}$ \\
Qwen3.5-397B-A17B & -0.076$^{***}$ & -0.158$^{***}$ & -1.231$^{***}$ & -1.065$^{***}$ \\
Qwen3.5-122B-A10B & -0.103$^{***}$ & -0.151$^{***}$ & -1.270$^{***}$ & -1.190$^{***}$ \\
Qwen3.5-27B & -0.138$^{***}$ & -0.182$^{***}$ & -1.339$^{***}$ & -1.233$^{***}$ \\
Qwen3.5-9B & -0.149$^{***}$ & -0.201$^{***}$ & -1.223$^{***}$ & -1.244$^{***}$ \\
\bottomrule
\end{tabular}
}
\label{tab:reg_annotation_type}
\end{table}

\begin{table}[H]
\caption{Complex-vs-simple coefficients for Annotation Count.}
\centering
\small
\setlength{\tabcolsep}{2pt}
\renewcommand{\arraystretch}{1.12}
\resizebox{\columnwidth}{!}{
\begin{tabular}{lrrrr}
\toprule
\textbf{Model} & \textbf{Exec.} & \textbf{Struct.} & \textbf{Sem.} & \textbf{Design} \\
\midrule
\multicolumn{5}{l}{\textit{Proprietary models}} \\
GPT-5.4 & -0.001 & -0.008 & -0.403$^{***}$ & -0.508$^{***}$ \\
Gemini 3.1 Pro Preview & -0.004 & -0.030$^{***}$ & -0.357$^{***}$ & -0.428$^{***}$ \\
Gemini 3 Flash Preview & -0.006 & -0.073$^{***}$ & -0.402$^{***}$ & -0.468$^{***}$ \\
Claude Sonnet 4.6 & -0.007 & -0.025$^{***}$ & -0.397$^{***}$ & -0.443$^{***}$ \\
\midrule
\multicolumn{5}{l}{\textit{Open-source models}} \\
Kimi K2.5 & -0.013$^{*}$ & -0.041$^{***}$ & -0.469$^{***}$ & -0.563$^{***}$ \\
Gemma 4 31B & -0.013 & -0.070$^{***}$ & -0.462$^{***}$ & -0.592$^{***}$ \\
Qwen3.5-397B-A17B & -0.032$^{***}$ & -0.056$^{***}$ & -0.497$^{***}$ & -0.658$^{***}$ \\
Qwen3.5-122B-A10B & -0.053$^{***}$ & -0.066$^{***}$ & -0.554$^{***}$ & -0.742$^{***}$ \\
Qwen3.5-27B & -0.040$^{***}$ & -0.068$^{***}$ & -0.525$^{***}$ & -0.660$^{***}$ \\
Qwen3.5-9B & -0.083$^{***}$ & -0.091$^{***}$ & -0.674$^{***}$ & -0.854$^{***}$ \\
\bottomrule
\end{tabular}
}
\label{tab:reg_annotation_count}
\end{table}

\begin{table}[H]
\caption{Complex-vs-simple coefficients for Annotation Spatial Distribution Entropy.}
\centering
\small
\setlength{\tabcolsep}{2pt}
\renewcommand{\arraystretch}{1.12}
\resizebox{\columnwidth}{!}{
\begin{tabular}{lrrrr}
\toprule
\textbf{Model} & \textbf{Exec.} & \textbf{Struct.} & \textbf{Sem.} & \textbf{Design} \\
\midrule
\multicolumn{5}{l}{\textit{Proprietary models}} \\
GPT-5.4 & -0.006$^{*}$ & -0.040$^{***}$ & -0.253$^{***}$ & -0.295$^{***}$ \\
Gemini 3.1 Pro Preview & -0.007$^{**}$ & -0.033$^{***}$ & -0.245$^{***}$ & -0.267$^{***}$ \\
Gemini 3 Flash Preview & -0.008$^{*}$ & -0.071$^{***}$ & -0.302$^{***}$ & -0.307$^{***}$ \\
Claude Sonnet 4.6 & -0.001 & -0.028$^{***}$ & -0.250$^{***}$ & -0.289$^{***}$ \\
\midrule
\multicolumn{5}{l}{\textit{Open-source models}} \\
Kimi K2.5 & -0.000 & -0.043$^{***}$ & -0.261$^{***}$ & -0.284$^{***}$ \\
Gemma 4 31B & -0.013 & -0.058$^{***}$ & -0.298$^{***}$ & -0.349$^{***}$ \\
Qwen3.5-397B-A17B & -0.024$^{***}$ & -0.063$^{***}$ & -0.330$^{***}$ & -0.409$^{***}$ \\
Qwen3.5-122B-A10B & -0.036$^{***}$ & -0.073$^{***}$ & -0.367$^{***}$ & -0.490$^{***}$ \\
Qwen3.5-27B & -0.024$^{**}$ & -0.059$^{***}$ & -0.371$^{***}$ & -0.407$^{***}$ \\
Qwen3.5-9B & -0.042$^{***}$ & -0.081$^{***}$ & -0.417$^{***}$ & -0.521$^{***}$ \\
\bottomrule
\end{tabular}
}
\label{tab:reg_entropy}
\end{table}

\begin{table}[H]
\caption{Complex-vs-simple coefficients for Visual Element Occupancy.}
\centering
\small
\setlength{\tabcolsep}{2pt}
\renewcommand{\arraystretch}{1.12}
\resizebox{\columnwidth}{!}{
\begin{tabular}{lrrrr}
\toprule
\textbf{Model} & \textbf{Exec.} & \textbf{Struct.} & \textbf{Sem.} & \textbf{Design} \\
\midrule
\multicolumn{5}{l}{\textit{Proprietary models}} \\
GPT-5.4 & 0.005 & -0.020$^{*}$ & -0.220$^{***}$ & -0.227$^{***}$ \\
Gemini 3.1 Pro Preview & -0.001 & -0.001 & -0.275$^{***}$ & -0.304$^{***}$ \\
Gemini 3 Flash Preview & 0.001 & -0.011 & -0.219$^{***}$ & -0.254$^{***}$ \\
Claude Sonnet 4.6 & 0.005 & -0.000 & -0.216$^{***}$ & -0.230$^{***}$ \\
\midrule
\multicolumn{5}{l}{\textit{Open-source models}} \\
Kimi K2.5 & -0.004 & -0.011 & -0.271$^{***}$ & -0.298$^{***}$ \\
Gemma 4 31B & -0.023$^{*}$ & -0.039$^{***}$ & -0.383$^{***}$ & -0.397$^{***}$ \\
Qwen3.5-397B-A17B & 0.035$^{***}$ & 0.013 & -0.174$^{**}$ & -0.214$^{***}$ \\
Qwen3.5-122B-A10B & 0.021 & -0.009 & -0.269$^{***}$ & -0.295$^{***}$ \\
Qwen3.5-27B & 0.017 & -0.001 & -0.189$^{**}$ & -0.235$^{***}$ \\
Qwen3.5-9B & 0.028 & -0.006 & -0.197$^{**}$ & -0.188$^{**}$ \\
\bottomrule
\end{tabular}
}
\label{tab:reg_occupancy}
\end{table}

\subsection{Detailed Complexity Heatmaps}
\label{app:complexity_heatmaps}

\begin{figure}[!htbp]
    \centering
    \includegraphics[width=\columnwidth]{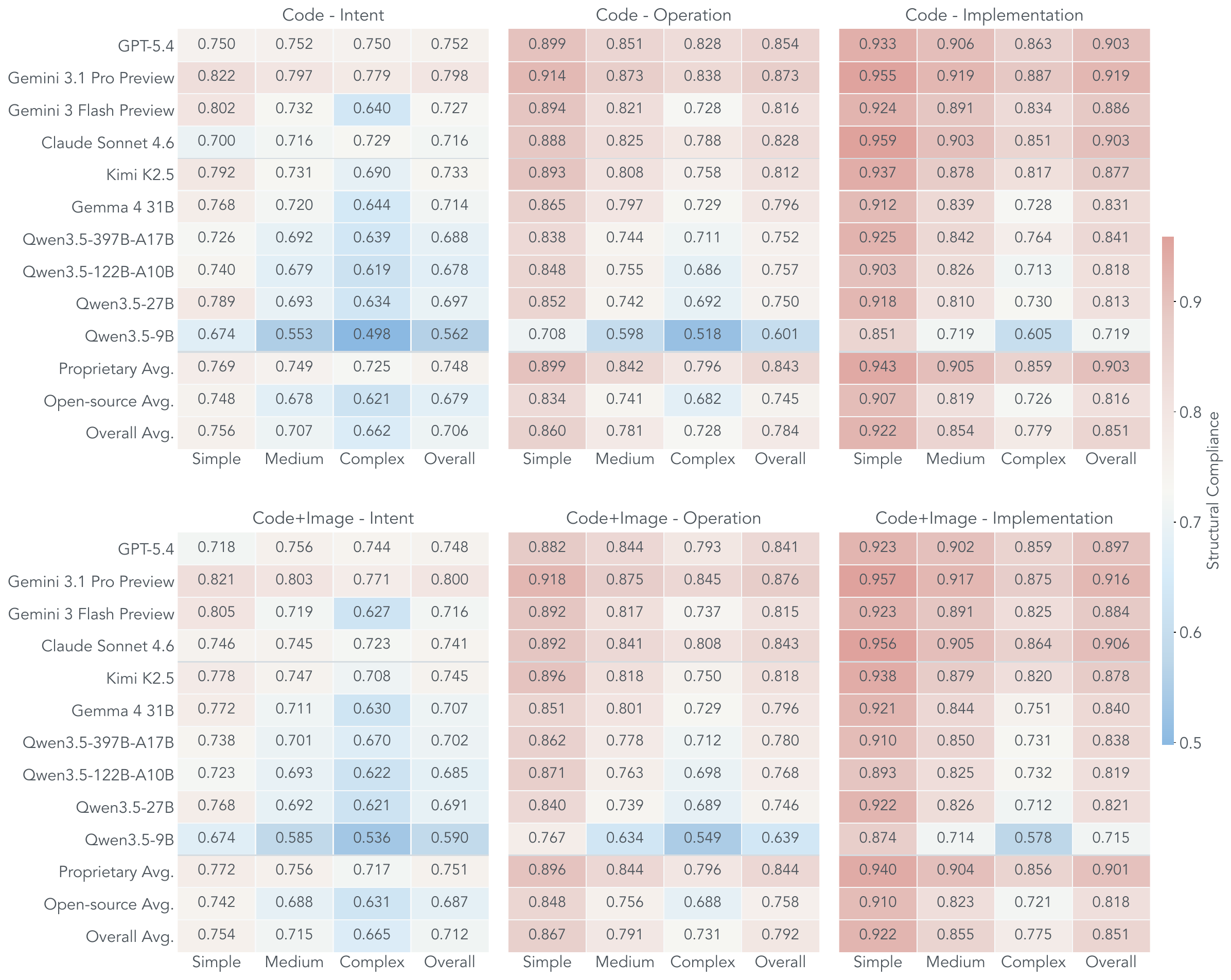}
    \caption{Structural Compliance across Code Token Increment groups.}
    \label{fig:complexity_heatmap_struct}
\end{figure}

\begin{figure}[!htbp]
    \centering
    \includegraphics[width=\columnwidth]{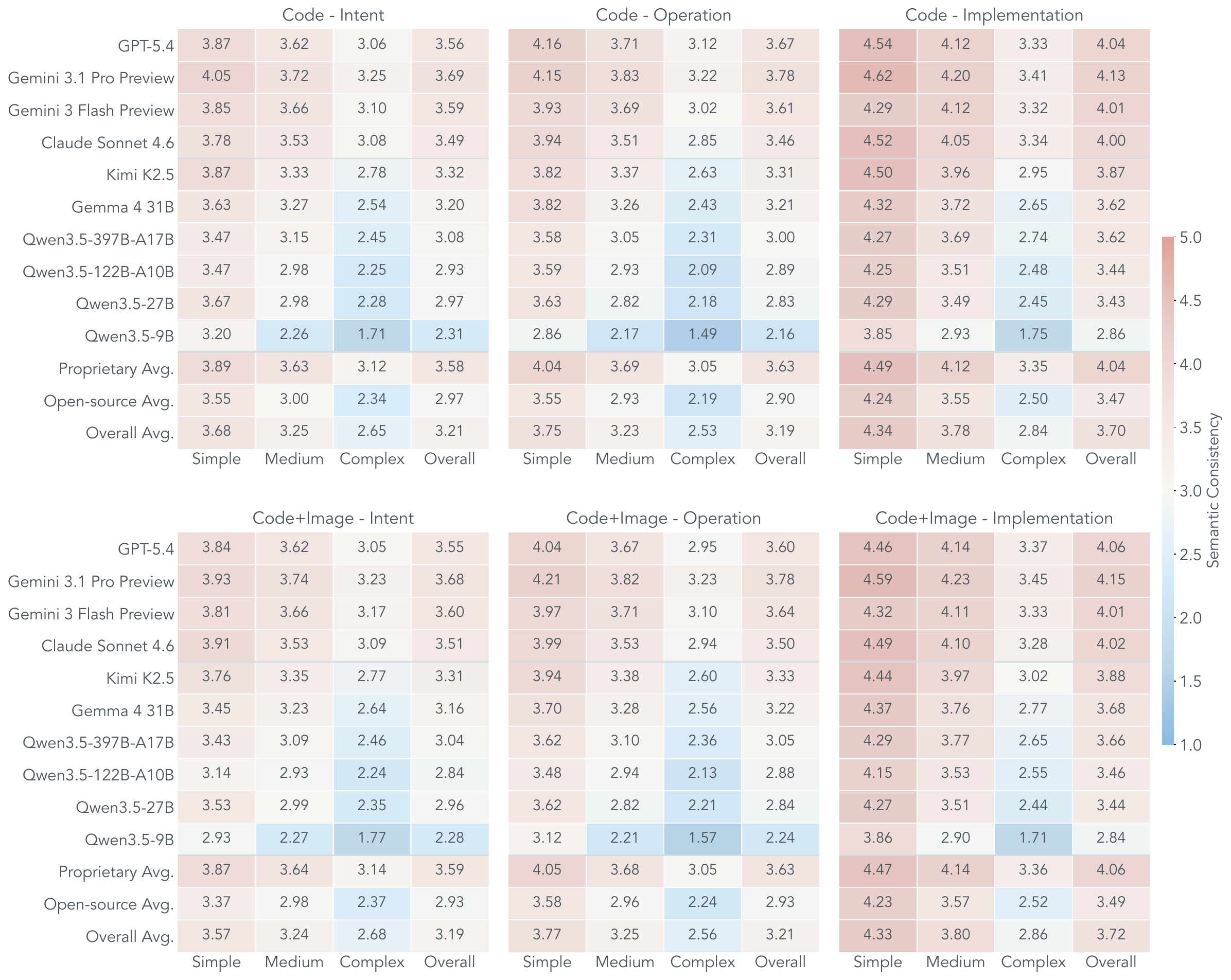}
    \caption{Semantic Consistency across Code Token Increment groups.}
    \label{fig:complexity_heatmap_semantic}
\end{figure}

\begin{figure}[!htbp]
    \centering
    \includegraphics[width=\columnwidth]{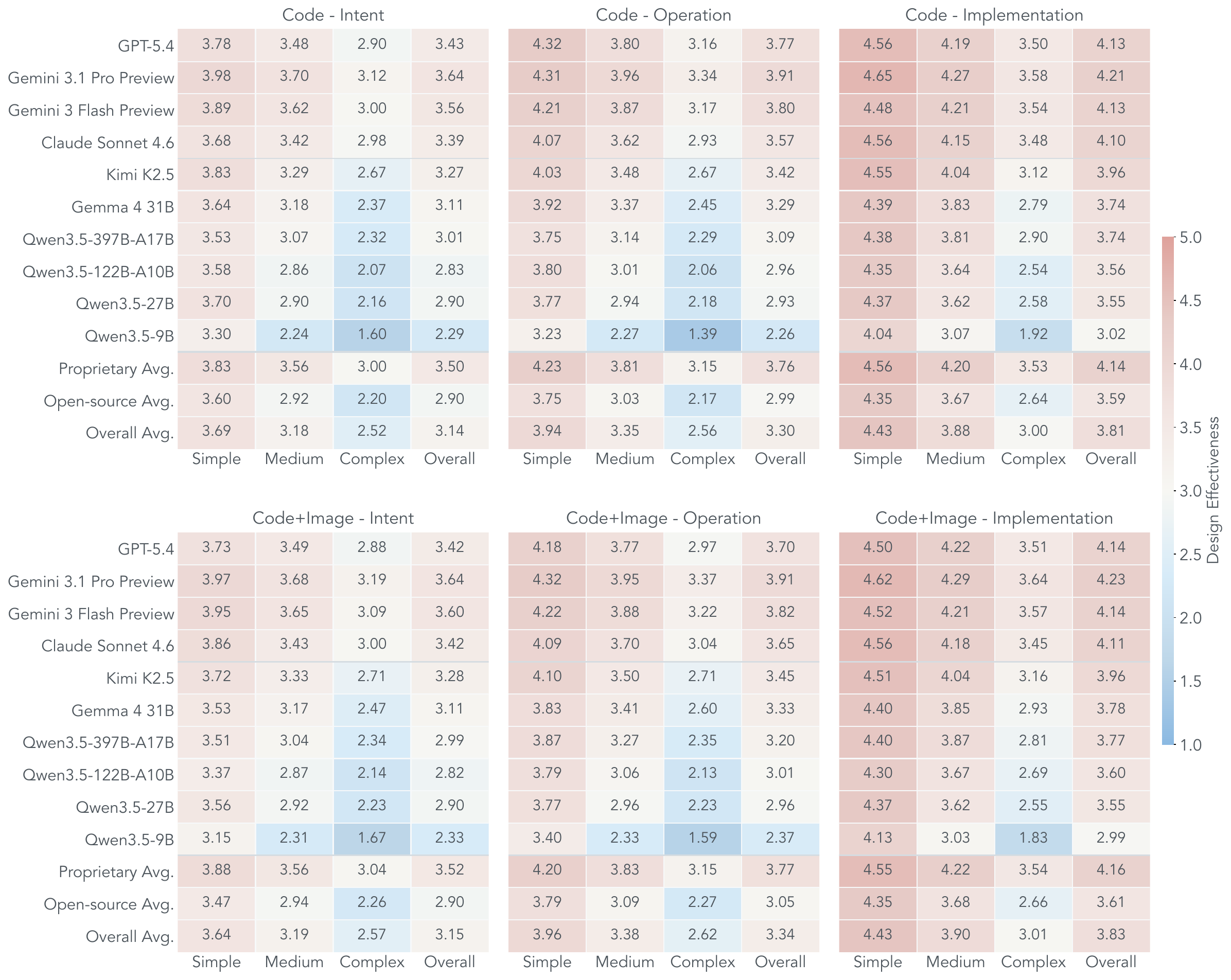}
    \caption{Design Effectiveness across Code Token Increment groups.}
    \label{fig:complexity_heatmap_design}
\end{figure}

We further provide heatmaps grouped by Code Token Increment, the strongest complexity indicator in the main analysis.
Figures~\ref{fig:complexity_heatmap_struct}, \ref{fig:complexity_heatmap_semantic}, and~\ref{fig:complexity_heatmap_design} show Structural Compliance, Semantic Consistency, and Design Effectiveness.
Each heatmap reports model performance across simple, medium, and complex groups under three instruction levels and two input settings.

High-complexity cases remain difficult even under Implementation-level instructions, showing that concrete instructions do not fully remove the difficulty of complex annotation code.

\section{Runtime Error and Chart Fidelity Violation Statistics}
\label{app:runtime_error_statistics}

Tables~\ref{tab:runtime_error_statistics} and~\ref{tab:fidelity_violation_statistics} report runtime error and chart fidelity violation distributions for all models.
Percentages are computed within each model over the corresponding failure cases.
In Tab.~\ref{tab:runtime_error_statistics}, Attr., Value, Type, Syntax, Name, Other, and Index denote AttributeError, ValueError, TypeError, SyntaxError, NameError, uncategorized runtime errors, and IndexError.
In Tab.~\ref{tab:fidelity_violation_statistics}, Layout, Geometry, and Data Mark denote layout, figure geometry, and data mark violations.

\begin{table}[H]
\caption{Runtime error distributions by model. Total denotes the number of failed execution cases.}
\centering
\small
\setlength{\tabcolsep}{3pt}
\resizebox{\columnwidth}{!}{
\begin{tabular}{lrrrrrrrr}
\toprule
\textbf{Model}
& \textbf{Attr.}
& \textbf{Value}
& \textbf{Type}
& \textbf{Syntax}
& \textbf{Name}
& \textbf{Other}
& \textbf{Index}
& \textbf{Total} \\
\midrule
\multicolumn{9}{c}{\textit{Proprietary Models}} \\
\midrule
GPT-5.4 & 24.62\% & 35.38\% & 23.08\% & 10.77\% & 1.54\% & 3.08\% & 1.54\% & 65 \\
Gemini 3.1 Pro Preview & 38.46\% & 7.69\% & 25.64\% & 15.38\% & 12.82\% & 0.00\% & 0.00\% & 39 \\
Gemini 3 Flash Preview & 28.17\% & 23.94\% & 18.31\% & 14.08\% & 2.82\% & 12.68\% & 0.00\% & 71 \\
Claude Sonnet 4.6 & 43.67\% & 8.86\% & 20.89\% & 15.82\% & 1.90\% & 6.33\% & 2.53\% & 158 \\
\midrule
\multicolumn{9}{c}{\textit{Open-Source Models}} \\
\midrule
Kimi K2.5 & 25.36\% & 15.79\% & 16.75\% & 18.18\% & 15.31\% & 4.78\% & 3.83\% & 209 \\
Gemma 4 31B & 37.53\% & 23.11\% & 25.17\% & 8.92\% & 0.23\% & 4.35\% & 0.69\% & 437 \\
Qwen3.5-397B-A17B & 33.86\% & 21.75\% & 18.83\% & 9.64\% & 4.26\% & 8.74\% & 2.91\% & 446 \\
Qwen3.5-122B-A10B & 40.37\% & 24.87\% & 16.18\% & 6.30\% & 3.58\% & 4.94\% & 3.75\% & 587 \\
Qwen3.5-27B & 32.21\% & 31.24\% & 18.52\% & 4.03\% & 1.77\% & 7.57\% & 4.67\% & 621 \\
Qwen3.5-9B & 34.60\% & 22.34\% & 17.38\% & 9.46\% & 3.34\% & 4.97\% & 7.91\% & 1289 \\
\bottomrule
\end{tabular}
}
\label{tab:runtime_error_statistics}
\end{table}

\begin{table}[H]
\caption{Chart fidelity violation distributions by model. Total denotes the number of chart fidelity violation cases.}
\centering
\small
\setlength{\tabcolsep}{4pt}
\resizebox{\columnwidth}{!}{
\begin{tabular}{lrrrr}
\toprule
\textbf{Model}
& \textbf{Layout}
& \textbf{Geometry}
& \textbf{Data Mark}
& \textbf{Total} \\
\midrule
\multicolumn{5}{c}{\textit{Proprietary Models}} \\
\midrule
GPT-5.4 & 58.91\% & 1.18\% & 39.91\% & 679 \\
Gemini 3.1 Pro Preview & 78.35\% & 3.87\% & 17.78\% & 388 \\
Gemini 3 Flash Preview & 32.34\% & 52.86\% & 14.81\% & 1506 \\
Claude Sonnet 4.6 & 62.90\% & 20.04\% & 17.06\% & 469 \\
\midrule
\multicolumn{5}{c}{\textit{Open-Source Models}} \\
\midrule
Kimi K2.5 & 73.47\% & 12.60\% & 13.93\% & 524 \\
Gemma 4 31B & 56.52\% & 26.85\% & 16.62\% & 782 \\
Qwen3.5-397B-A17B & 70.05\% & 12.25\% & 17.70\% & 661 \\
Qwen3.5-122B-A10B & 75.00\% & 7.19\% & 17.81\% & 612 \\
Qwen3.5-27B & 67.07\% & 13.59\% & 19.34\% & 574 \\
Qwen3.5-9B & 70.79\% & 6.59\% & 22.61\% & 743 \\
\bottomrule
\end{tabular}
}
\label{tab:fidelity_violation_statistics}
\end{table}

\section{Representative Error Cases}
\label{app:error_cases}

This section provides additional error examples for both rule-based and LLM-judged evaluation.
We first show cases for \emph{Chart Fidelity}, \emph{Annotation Matching}, and \emph{Color Matching}.
We then show cases for \emph{Semantic Faithfulness}, \emph{Semantic Clarity}, \emph{Visual Clarity}, \emph{Annotation Organization Quality}, and \emph{Attention Guidance}.
The corresponding figures are collected in Sec.~\ref{app:error_case_gallery}.

\subsection{Rule-based Evaluation Cases}
\label{app:rule_based_error_cases}

\textbf{Chart Fidelity.}
Fig.~\ref{fig:error_case_chart_fidelity} shows a chart fidelity failure in an area chart task.
The high-scoring result preserves the base chart structure.
The low-scoring result is executable, but changes the plotting area and area proportions, which disrupts the base coordinate system and layout.

\textbf{Annotation Matching.}
Fig.~\ref{fig:error_case_annotation_matching} shows an annotation matching failure in a line chart task.
The low-scoring result partly follows the x-axis intent, but differs from the reference annotations in target objects, annotation count, and spatial positions.

\textbf{Color Matching.}
Fig.~\ref{fig:error_case_color_matching} shows a color matching failure in a bar chart task.
The reference chart uses orange to highlight only a small set of target bars, while the remaining bars keep the base teal color.
The high-scoring result preserves this contrast.
The low-scoring result adds annotation text and guide lines, but changes almost all bars to orange, making the target bars hard to distinguish from ordinary bars.

\subsection{LLM-judged Evaluation Cases}
\label{app:llm_judged_error_cases}

\textbf{Semantic Faithfulness.}
Fig.~\ref{fig:error_case_semantic_faithfulness} shows a semantic faithfulness failure.
The instruction asks the model to draw two boundary lines separating the outer summer region from the central winter region.
The high-scoring result places the boundaries and region labels consistently with the reference chart.
The low-scoring result draws boundary lines, but reverses the summer and winter meanings.

\textbf{Semantic Clarity.}
Fig.~\ref{fig:error_case_semantic_clarity} shows a semantic clarity failure in a line chart task.
The instruction asks the model to mark the price point corresponding to the Brexit referendum on June 23, 2016.
The high-scoring result anchors the arrow and red marker to the correct point on the line.
The low-scoring result has correct text and a clean layout, but the red marker floats above the curve, making the relation between the event and the data point unclear.

\textbf{Visual Clarity.}
Fig.~\ref{fig:error_case_visual_clarity} shows a visual clarity failure in an event annotation task.
The high-scoring result distributes event labels around the line chart and keeps the reading path clear.
The low-scoring result stacks multiple dates, labels, and arrows in a small region, causing severe overlap.
Although part of the intended content is present, the annotation text is difficult to read.

\textbf{Annotation Organization Quality.}
Fig.~\ref{fig:error_case_annotation_organization} shows an annotation organization failure.
The high-scoring result organizes the subtitle, explanatory annotation, and final data point with a clear visual hierarchy.
The low-scoring result preserves the main content, but the subtitle moves downward and competes with the legend and source area.
The explanatory annotation is also too long and extends beyond the chart area.

\textbf{Attention Guidance.}
Fig.~\ref{fig:error_case_attention_guidance} shows an attention guidance failure in a bar chart task.
The instruction aims to guide attention to the peak snowfall period from 6 p.m. to 11 p.m.
The high-scoring result highlights the target interval with darker bars.
The low-scoring result does not highlight the peak bars clearly, and its labels are far from the key region.
Although the base chart remains readable, the intended focus is weak.

\section{Human Evaluation and LLM-Judge Validation Details}
\label{app:human_evaluation_details}

For each sample, raters were shown the annotation instruction, the unannotated base chart, the generated annotated chart, and the reference annotated chart.
They independently rated each output using the same five 0--5 criteria used by the LLM judge:
\emph{Semantic Faithfulness},
\emph{Semantic Clarity},
\emph{Visual Clarity},
\emph{Annotation Organization Quality},
and
\emph{Attention Guidance}.

\begin{figure}[!htbp]
    \centering
    \includegraphics[width=\columnwidth]{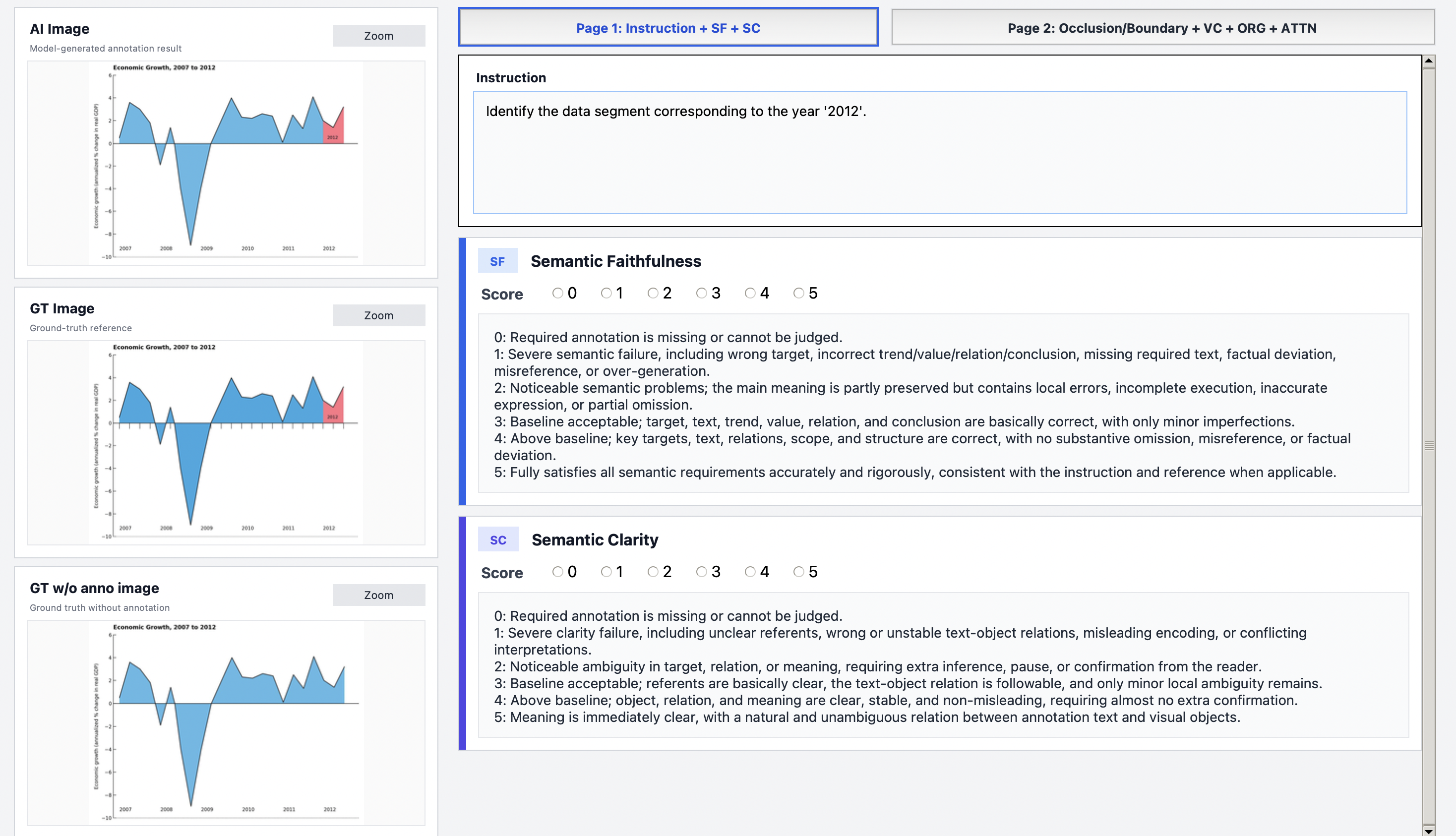}
    \caption{Screenshot of the human rating interface used for judge validation.}
    \label{fig:human_rating_interface}
\end{figure}

\begin{figure*}[!htbp]
    \centering
    \includegraphics[width=\textwidth]{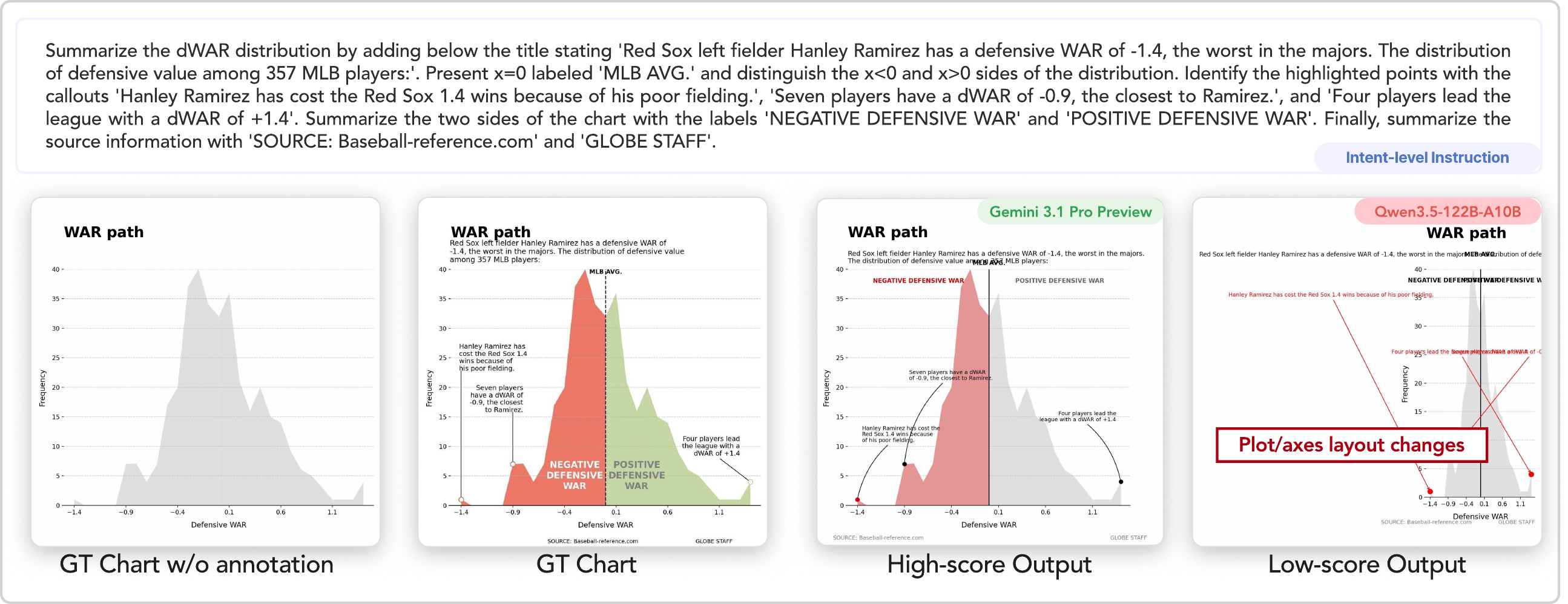}
    \caption{Representative error case for rule-based \emph{Chart Fidelity}. The low-scoring and high-scoring outputs are produced under the same instruction and reference chart.}
    \label{fig:error_case_chart_fidelity}
\end{figure*}

\begin{figure*}[!htbp]
    \centering
    \includegraphics[width=\textwidth]{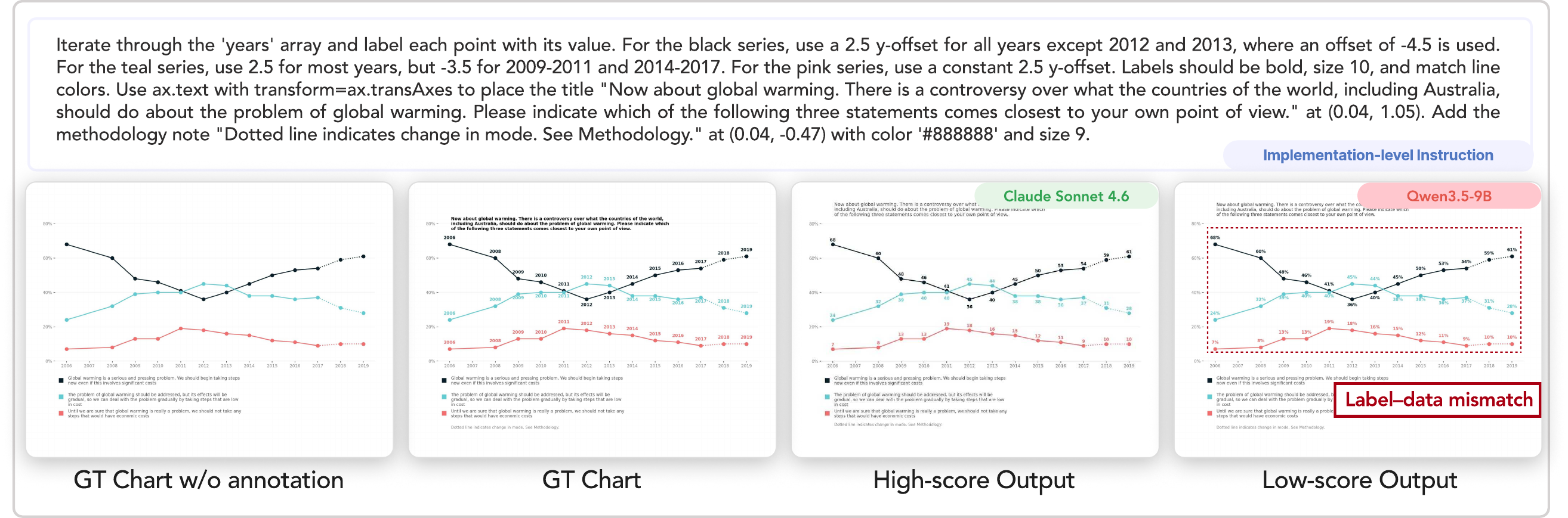}
    \caption{Representative error case for rule-based \emph{Annotation Matching}. The low-scoring and high-scoring outputs are produced under the same instruction and reference chart.}
    \label{fig:error_case_annotation_matching}
\end{figure*}

\begin{figure*}[!htbp]
    \centering
    \includegraphics[width=\textwidth]{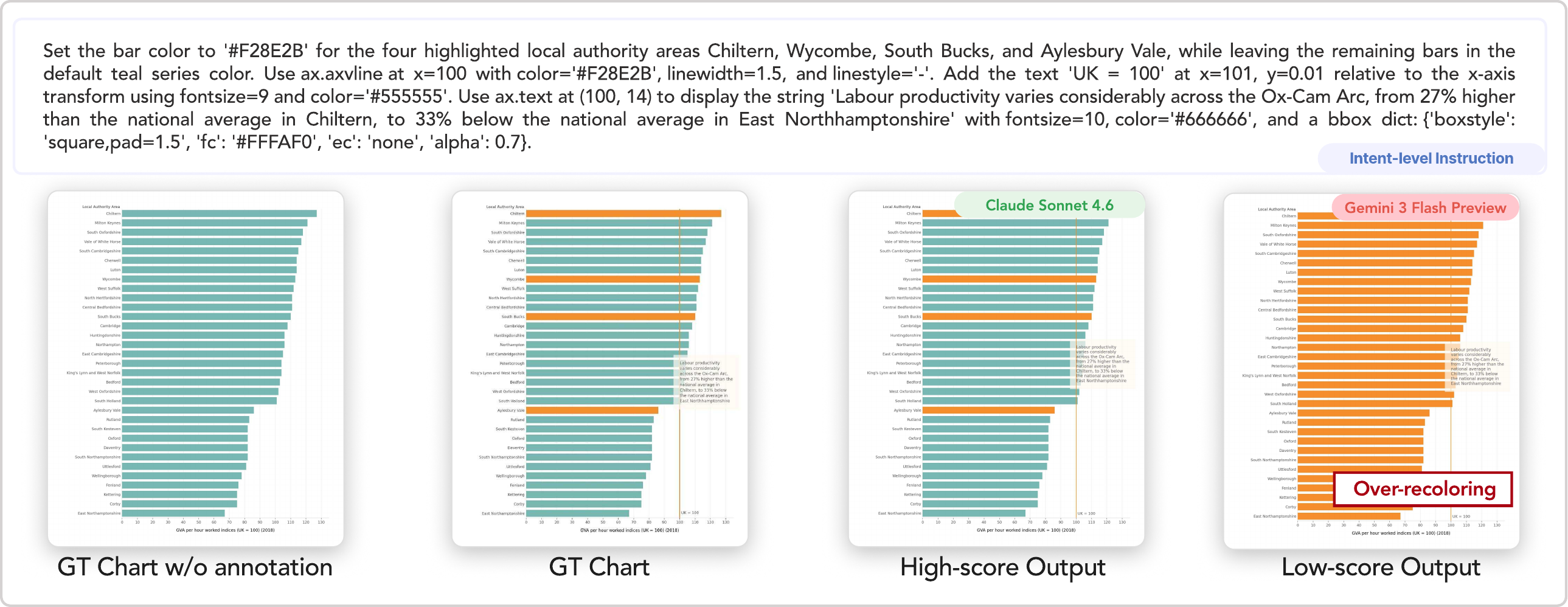}
    \caption{Representative error case for rule-based \emph{Color Matching}. The low-scoring and high-scoring outputs are produced under the same instruction and reference chart.}
    \label{fig:error_case_color_matching}
\end{figure*}

\begin{figure*}[!htbp]
    \centering
    \includegraphics[width=\textwidth]{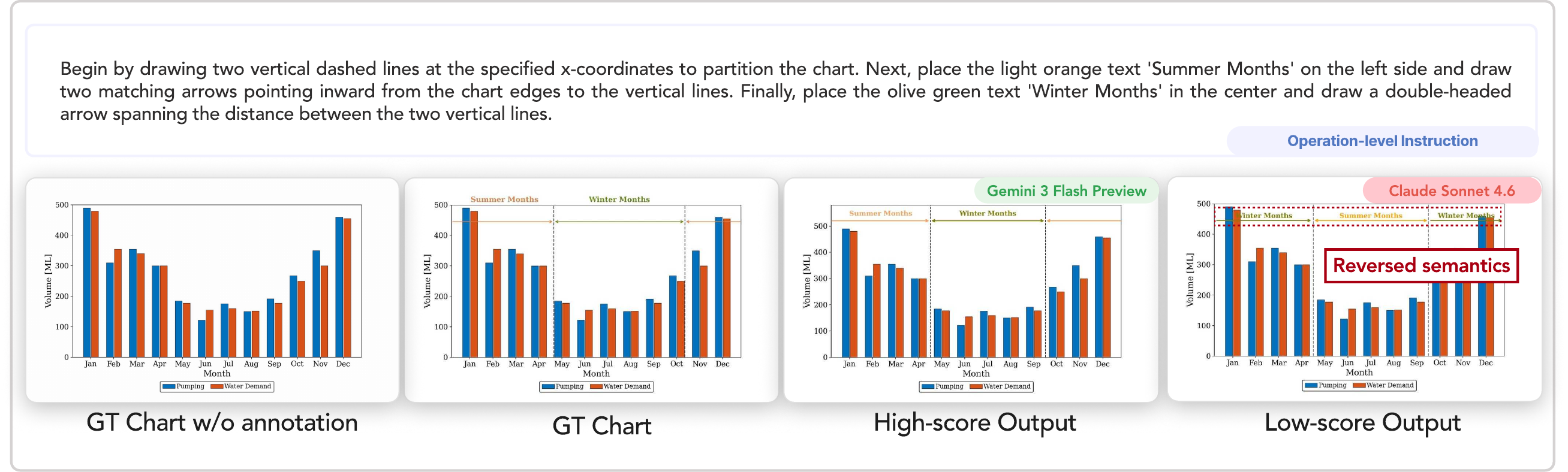}
    \caption{Representative error case for LLM-judged \emph{Semantic Faithfulness}. The low-scoring and high-scoring outputs are produced under the same instruction and reference chart.}
    \label{fig:error_case_semantic_faithfulness}
\end{figure*}

\begin{figure*}[!htbp]
    \centering
    \includegraphics[width=\textwidth]{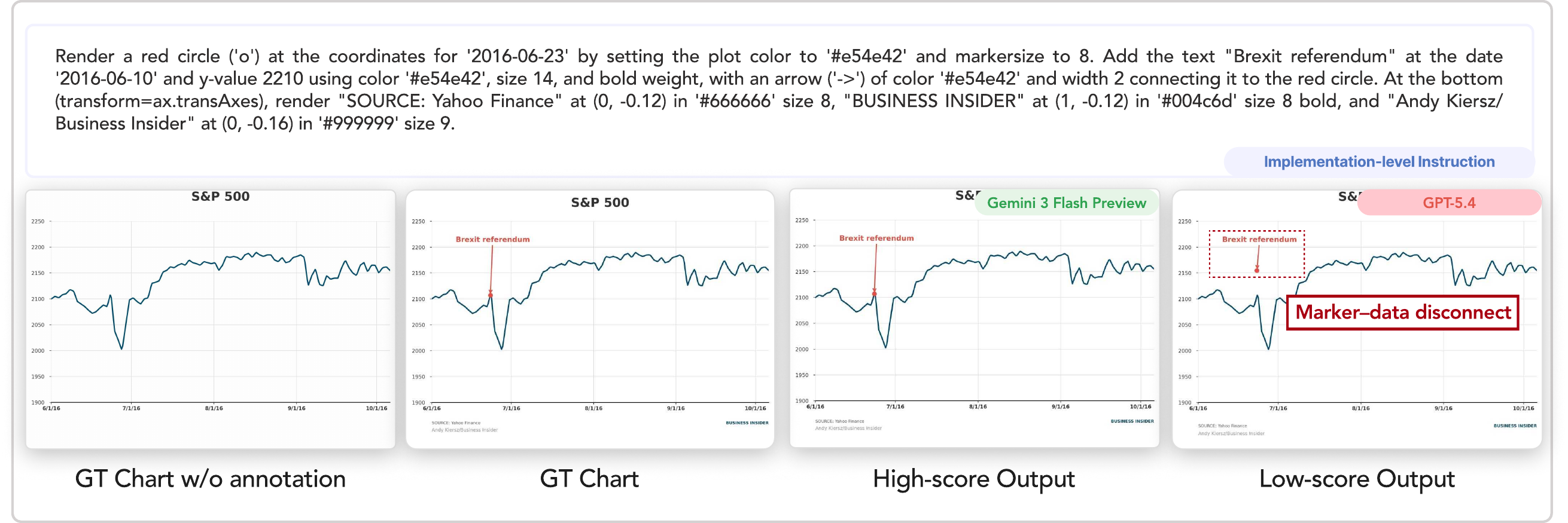}
    \caption{Representative error case for LLM-judged \emph{Semantic Clarity}. The low-scoring and high-scoring outputs are produced under the same instruction and reference chart.}
    \label{fig:error_case_semantic_clarity}
\end{figure*}

\begin{figure*}[!htbp]
    \centering
    \includegraphics[width=\textwidth]{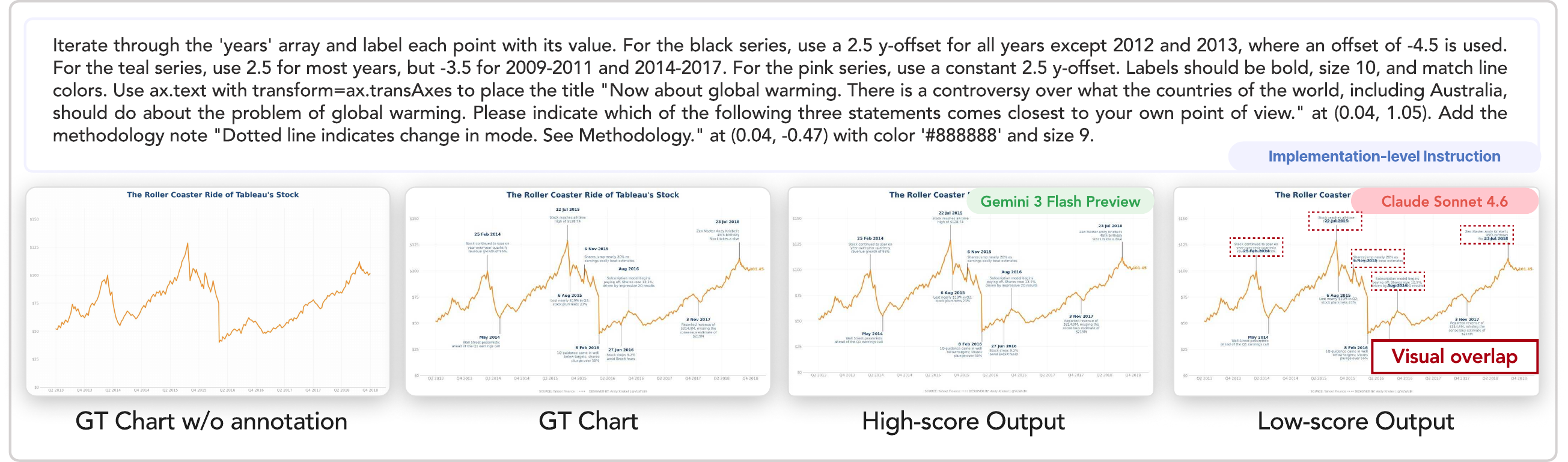}
    \caption{Representative error case for LLM-judged \emph{Visual Clarity}. The low-scoring and high-scoring outputs are produced under the same instruction and reference chart.}
    \label{fig:error_case_visual_clarity}
\end{figure*}

\begin{figure*}[!htbp]
    \centering
    \includegraphics[width=\textwidth]{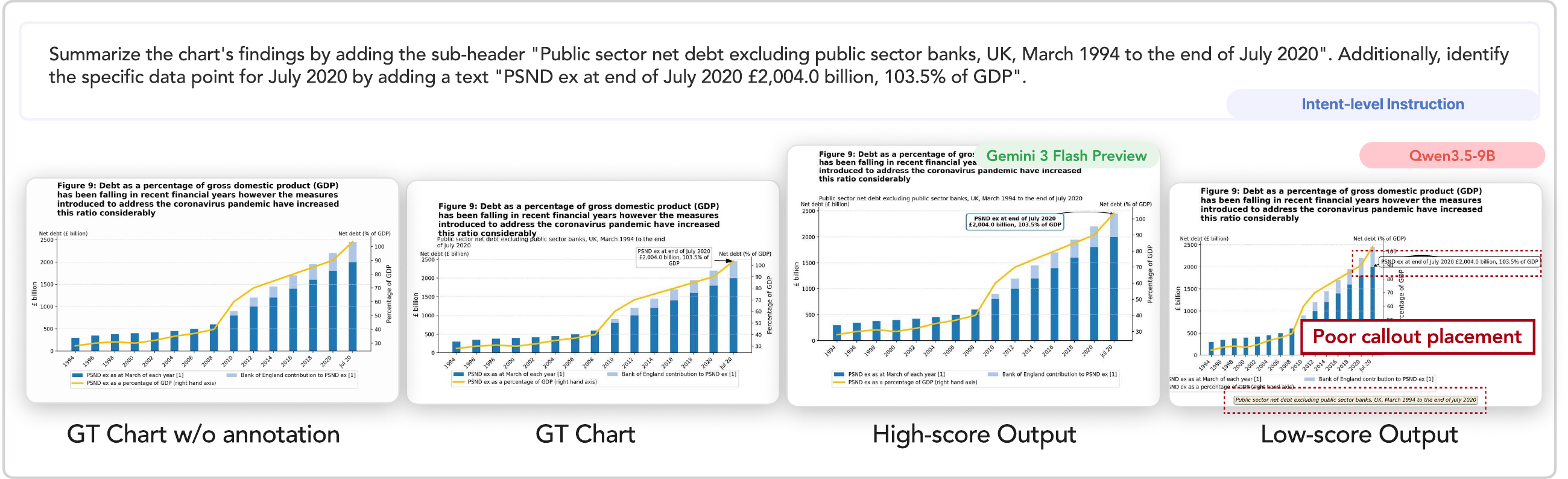}
    \caption{Representative error case for LLM-judged \emph{Annotation Organization Quality}. The low-scoring and high-scoring outputs are produced under the same instruction and reference chart.}
    \label{fig:error_case_annotation_organization}
\end{figure*}

\begin{figure*}[!htbp]
    \centering
    \includegraphics[width=\textwidth]{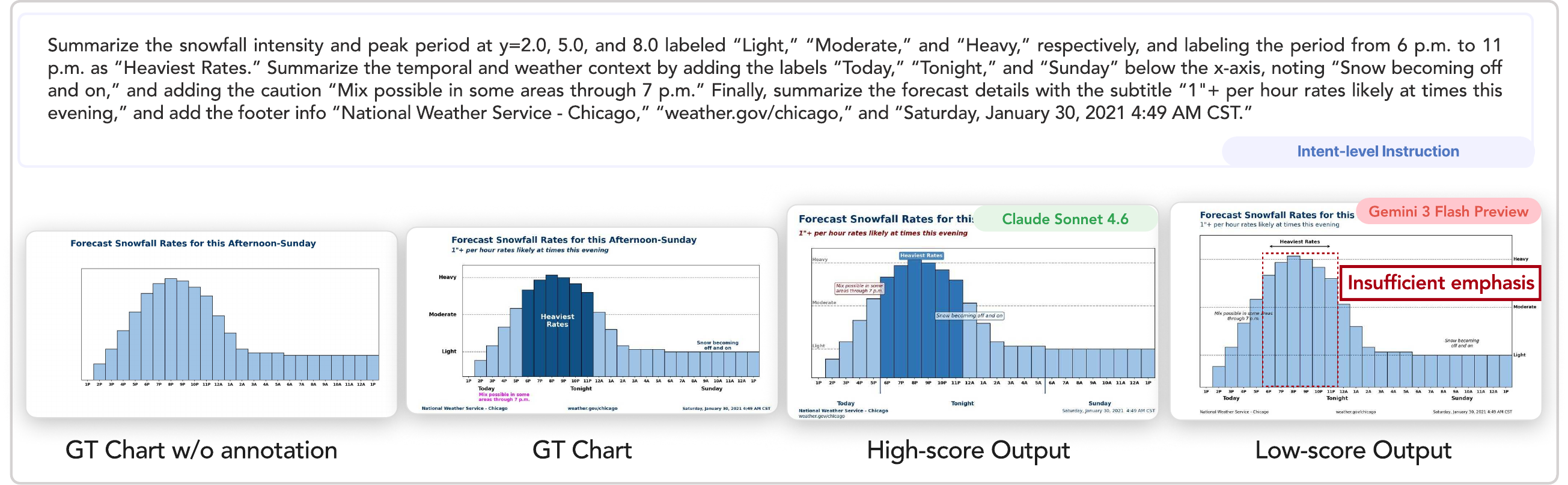}
    \caption{Representative error case for LLM-judged \emph{Attention Guidance}. The low-scoring and high-scoring outputs are produced under the same instruction and reference chart.}
    \label{fig:error_case_attention_guidance}
\end{figure*}

A screenshot of the rating interface is shown in Fig.~\ref{fig:human_rating_interface}.

We provide additional validation of the LLM-based judge from two perspectives:
(1) alignment with human ratings across outputs from models with different capability levels, and
(2) potential judge-model bias through cross-judge consistency analysis.

\subsection{Human Alignment Across Model Capability Levels}
\label{app:human_alignment}

Following the human evaluation protocol described above, we validate whether GPT-5.4 judgments align with expert ratings across outputs from different model capability levels.
We compute Spearman correlation between aggregated human ratings and averaged GPT-5.4 scores to measure human alignment. GPT-5.4 evaluates each sample three times, and ICC(3,1) across repeated judgments is used to measure judging stability.
Tab.~\ref{tab:human_alignment_extended} reports the results.

\begin{table}[!htbp]
\caption{
GPT-5.4 judge validation across output models.
Spearman $\rho$ measures alignment between averaged GPT-5.4 scores and aggregated human ratings.
ICC(3,1) measures stability across three repeated GPT-5.4 runs.
}
\centering
\small
\setlength{\tabcolsep}{5pt}
\resizebox{\columnwidth}{!}{
\begin{tabular}{lrrrrrrr}
\toprule
&
&
\multicolumn{3}{c}{\textbf{Human Alignment: Spearman $\rho$}}
&
\multicolumn{3}{c}{\textbf{Repeated-Run Stability: ICC(3,1)}}\\
\cmidrule(lr){3-5}
\cmidrule(lr){6-8}
\textbf{Output Model}
&
\textbf{$N$}
&
\textbf{Semantic}
&
\textbf{Design}
&
\textbf{Overall}
&
\textbf{Semantic}
&
\textbf{Design}
&
\textbf{Overall}\\
\midrule

Gemini 3.1 Pro Preview
&300
&0.7427
&0.7341
&0.7935
&0.8615
&0.9353
&0.9156\\

Qwen3.5-9B
&150
&0.8027
&0.8602
&0.8678
&0.9121
&0.9453
&0.9412\\

Qwen3.5-27B
&150
&0.7981
&0.8326
&0.8205
&0.9005
&0.9249
&0.9281\\

Qwen Family
&300
&0.8097
&0.8552
&0.8513
&0.9086
&0.9378
&0.9370\\

\midrule

All Samples
&600
&\textbf{0.8194}
&\textbf{0.8378}
&\textbf{0.8593}
&\textbf{0.9109}
&\textbf{0.9474}
&\textbf{0.9427}\\

\bottomrule
\end{tabular}
}
\label{tab:human_alignment_extended}
\end{table}

Tab.~\ref{tab:human_alignment_extended} shows that alignment remains strong on outputs from weaker open-source models.
For Qwen3.5-9B and Qwen3.5-27B outputs, the Overall Spearman correlations are 0.8678 and 0.8205, comparable to or higher than the 0.7935 on Gemini 3.1 Pro Preview outputs, with ICC(3,1) values of 0.9412 and 0.9281.
This indicates that GPT-5.4 provides reliable judgments across different output-quality levels rather than only for outputs from a single strong model.
\subsection{Human Alignment Across Instruction Levels}
\label{app:human_alignment_levels}

We further examine whether GPT-5.4 judgments remain consistent with human ratings across the three instruction levels.
For each level, we compute Spearman correlation between aggregated human ratings and averaged GPT-5.4 scores, together with ICC(3,1) across the three repeated GPT-5.4 judgments.
Tab.~\ref{tab:human_alignment_by_level} reports the results.

\begin{table}[!htbp]
\caption{
GPT-5.4 judge validation across instruction levels.
Spearman $\rho$ measures alignment between averaged GPT-5.4 scores and aggregated human ratings, while ICC(3,1) measures stability across three repeated GPT-5.4 runs.
All Spearman correlations are significant at $p<.001$.
}
\centering
\small
\setlength{\tabcolsep}{4pt}
\resizebox{\columnwidth}{!}{
\begin{tabular}{lrrrrrr}
\toprule
&
&
\multicolumn{2}{c}{\textbf{Human Alignment: Spearman $\rho$}}
&
\multicolumn{3}{c}{\textbf{Repeated-Run Stability: ICC(3,1)}}\\
\cmidrule(lr){3-4}
\cmidrule(lr){5-7}
\textbf{Instruction Level}
&
\textbf{$N$}
&
\textbf{Semantic}
&
\textbf{Design}
&
\textbf{Semantic}
&
\textbf{Design}
&
\textbf{Overall}\\
\midrule
Intent         & 200 & 0.7912 & 0.8333 & 0.9113 & 0.9446 & 0.9441\\
Operation      & 200 & 0.8357 & 0.8362 & 0.9102 & 0.9327 & 0.9333\\
Implementation & 200 & 0.7678 & 0.7312 & 0.8761 & 0.9444 & 0.9265\\
\bottomrule
\end{tabular}
}
\label{tab:human_alignment_by_level}
\end{table}

GPT-5.4 maintains strong human alignment and high repeated-run stability across all three instruction levels.

\subsection{Cross-Judge Consistency and Judge-Model Bias}
\label{app:cross_judge}

We compare three candidate judges:
GPT-5.4, Claude Sonnet 4.6, and Gemini 3.1 Pro Preview.
The judges evaluate outputs from five models:
GPT-5.4, Claude Sonnet 4.6, Gemini 3.1 Pro Preview, Qwen3.5-27B, and Qwen3.5-9B.
For each output model, we randomly sample 300 outputs, resulting in 1,500 evaluated outputs.
All candidate judges use the same evaluation rubric.
We measure pairwise agreement using Spearman correlation and overall consistency among judges using Cronbach's $\alpha$.

\begin{table}[!htbp]
\caption{
Cross-judge consistency among GPT-5.4, Claude Sonnet 4.6, and Gemini 3.1 Pro Preview.
Pairwise agreement is measured using Spearman correlation, while Cronbach's $\alpha$ measures consistency among all three judges.
Each output-model subset contains 300 samples.
}
\centering
\scriptsize
\setlength{\tabcolsep}{4pt}
\resizebox{\columnwidth}{!}{
\begin{tabular}{llrrrr}
\toprule
\textbf{Output Model}
&
\textbf{Metric}
&
\textbf{GPT--Claude $\rho$}
&
\textbf{GPT--Gemini $\rho$}
&
\textbf{Gemini--Claude $\rho$}
&
\textbf{Cronbach's $\alpha$}
\\
\midrule

GPT-5.4
&Semantic Consistency
&0.682&0.603&0.622&0.814\\
&Design Effectiveness
&0.772&0.713&0.733&0.853\\
&Overall
&0.760&0.697&0.731&0.865\\

\midrule

Claude Sonnet 4.6
&Semantic Consistency
&0.744&0.649&0.669&0.852\\
&Design Effectiveness
&0.759&0.715&0.734&0.877\\
&Overall
&0.796&0.741&0.755&0.896\\

\midrule

Gemini 3.1 Pro Preview
&Semantic Consistency
&0.726&0.569&0.546&0.805\\
&Design Effectiveness
&0.752&0.684&0.695&0.844\\
&Overall
&0.775&0.687&0.676&0.860\\

\midrule

Qwen3.5-27B
&Semantic Consistency
&0.806&0.733&0.749&0.890\\
&Design Effectiveness
&0.878&0.813&0.809&0.908\\
&Overall
&0.875&0.803&0.817&0.918\\

\midrule

Qwen3.5-9B
&Semantic Consistency
&0.861&0.813&0.825&0.927\\
&Design Effectiveness
&0.898&0.830&0.840&0.921\\
&Overall
&0.902&0.855&0.852&0.937\\

\midrule

All Samples
&Semantic Consistency
&0.787&0.712&0.725&0.885\\
&Design Effectiveness
&0.832&0.777&0.785&0.898\\
&Overall
&0.842&0.784&0.793&0.913\\

\bottomrule
\end{tabular}
}
\label{tab:cross_judge}
\end{table}

Tab.~\ref{tab:cross_judge} shows strong consistency among the three candidate judges.
Pairwise Spearman correlations range from 0.546 to 0.937 across output models and metrics.
Agreement is lowest on Semantic Consistency for GPT-5.4's own outputs (0.603--0.682), whereas the Overall-score agreement exceeds 0.86 for all five output models.
As reported in the main paper, removing GPT-5.4 from the judge set leaves the ranking of the five output models unchanged.

\section{Detailed Results for Cross-Representation Generalization}
\label{app:representation_generalization}

Tab.~\ref{tab:d3_svg_results} reports the complete model-level results for the D3 and SVG extensions discussed in Sec.~5 of the main paper.
These results complement the aggregated comparison in Fig.~12 of the main paper and provide the performance of each evaluated model across all three instruction levels.
These model-level results echo the two findings summarized in Sec.~5 of the main paper.
First, the key trends established under Python carry over to the new representations: more specific instructions generally yield higher \emph{Semantic Consistency} and \emph{Design Effectiveness}, and the overall model ranking under D3 and SVG remains close to that under Python.
Second, the two representations exhibit different trade-offs across instruction levels, with D3 leading on the rule-based metrics while SVG slightly overtakes D3 on the LLM-judged metrics at the Implementation level.
As in the main evaluation, Intent-level \emph{Structural Compliance} reports \emph{Chart Fidelity} only and is therefore shown separately from the Structural Compliance scores at the Operation and Implementation levels.

\begin{table*}[!htbp]
\caption{Complete model-level evaluation results under D3 (left) and SVG (right) representations. Metrics, notation, and highlighting conventions follow Tab.~3 of the main paper. \protect\colorbox{gray!12}{Gray} Struct.$^\ast$ columns report \emph{Chart Fidelity} only for Intent-level instructions and are not directly comparable with \emph{Structural Compliance} at the other levels.}
\label{tab:d3_svg_results}
\centering
\small
\setlength{\tabcolsep}{2.5pt}
\resizebox{\textwidth}{!}{
\begin{tabular}{
l|
r>{\columncolor{gray!12}}rrr|
rrrr|
rrrr||
r>{\columncolor{gray!12}}rrr|
rrrr|
rrrr
}
\toprule
\textbf{Model}
& \multicolumn{12}{c||}{\textbf{D3}}
& \multicolumn{12}{c}{\textbf{SVG}} \\
\cmidrule(lr){2-13} \cmidrule(lr){14-25}
& \multicolumn{4}{c|}{Intent-level}
& \multicolumn{4}{c|}{Operation-level}
& \multicolumn{4}{c||}{Implementation-level}
& \multicolumn{4}{c|}{Intent-level}
& \multicolumn{4}{c|}{Operation-level}
& \multicolumn{4}{c}{Implementation-level} \\
\cmidrule(lr){2-5} \cmidrule(lr){6-9} \cmidrule(lr){10-13}
\cmidrule(lr){14-17} \cmidrule(lr){18-21} \cmidrule(lr){22-25}
& Exec. & Struct.$^\ast$ & Sem. & Design
& Exec. & Struct. & Sem. & Design
& Exec. & Struct. & Sem. & Design
& Exec. & Struct.$^\ast$ & Sem. & Design
& Exec. & Struct. & Sem. & Design
& Exec. & Struct. & Sem. & Design \\
\midrule
\rowcolor{groupgreen}
\multicolumn{25}{c}{\textit{Proprietary Models}} \\
\midrule

GPT-5.4
& \textbf{1.000} & \textbf{0.827} & 2.929 & \underline{3.125}
& \textbf{0.992} & \textbf{0.752} & 3.008 & 3.258
& 0.983 & 0.771 & 3.471 & 3.555
& \textbf{1.000} & \textbf{0.828} & 2.738 & 3.106
& \textbf{1.000} & 0.691 & 2.767 & 3.175
& \textbf{1.000} & 0.743 & 3.621 & \underline{3.706} \\

Gemini 3.1 Pro Preview
& 0.983 & \underline{0.809} & \textbf{3.084} & \textbf{3.317}
& \underline{0.983} & \underline{0.747} & \textbf{3.246} & \textbf{3.393}
& \underline{0.992} & \textbf{0.807} & \textbf{3.580} & \textbf{3.681}
& \textbf{1.000} & \underline{0.817} & \textbf{3.042} & \textbf{3.320}
& \textbf{1.000} & \textbf{0.728} & \textbf{3.146} & \textbf{3.511}
& \textbf{1.000} & \textbf{0.802} & \textbf{3.696} & \textbf{3.736} \\

Gemini 3 Flash Preview
& \underline{0.992} & 0.783 & \underline{3.017} & 3.078
& \underline{0.983} & 0.737 & \underline{3.122} & \underline{3.359}
& 0.967 & \underline{0.804} & \underline{3.517} & \underline{3.635}
& \underline{0.983} & 0.737 & \underline{2.917} & \underline{3.267}
& \textbf{1.000} & \underline{0.693} & \underline{2.871} & \underline{3.333}
& \underline{0.992} & 0.778 & \underline{3.667} & 3.694 \\

Claude Sonnet 4.6
& \textbf{1.000} & 0.723 & 3.013 & 2.981
& \textbf{0.992} & 0.728 & 3.029 & 3.090
& \textbf{1.000} & 0.788 & 3.517 & 3.503
& \textbf{1.000} & 0.633 & 2.779 & 3.003
& \textbf{1.000} & 0.667 & 2.746 & 3.231
& \textbf{1.000} & \underline{0.801} & 3.646 & 3.625 \\

\midrule
\rowcolor{groupgreen}
\multicolumn{25}{c}{\textit{Open-Source Models}} \\
\midrule

Kimi K2.5
& \textbf{0.983} & 0.773 & \textbf{2.777} & \textbf{2.919}
& \textbf{1.000} & \textbf{0.745} & \textbf{2.842} & \textbf{3.031}
& \textbf{0.992} & \textbf{0.806} & \textbf{3.329} & \textbf{3.403}
& \textbf{0.992} & \textbf{0.792} & \textbf{2.338} & \textbf{2.892}
& \textbf{0.992} & \textbf{0.658} & \textbf{2.367} & \textbf{2.981}
& \textbf{0.992} & \textbf{0.773} & \textbf{3.338} & \underline{3.517} \\

Gemma 4 31B
& \underline{0.958} & \underline{0.824} & \underline{2.722} & \underline{2.829}
& \underline{0.975} & 0.675 & \underline{2.551} & \underline{2.858}
& 0.958 & 0.716 & \underline{3.217} & 3.328
& \underline{0.842} & \underline{0.720} & 2.091 & 2.625
& \underline{0.833} & 0.510 & 2.072 & 2.683
& 0.817 & 0.635 & 3.103 & 3.360 \\

Qwen3.5-397B-A17B
& \underline{0.958} & 0.761 & 2.635 & 2.777
& 0.967 & 0.680 & 2.530 & 2.805
& \underline{0.967} & 0.722 & 3.211 & \underline{3.399}
& \underline{0.842} & 0.685 & \underline{2.197} & \underline{2.724}
& 0.825 & 0.486 & \underline{2.117} & \underline{2.809}
& \underline{0.825} & 0.554 & \underline{3.270} & \textbf{3.562} \\

Qwen3.5-122B-A10B
& 0.950 & \textbf{0.833} & 2.496 & 2.667
& \underline{0.975} & 0.701 & 2.500 & 2.735
& \textbf{0.992} & \underline{0.746} & 3.029 & 3.177
& 0.792 & 0.683 & 2.126 & 2.596
& 0.792 & \underline{0.572} & 2.055 & 2.667
& 0.800 & \underline{0.659} & 3.165 & 3.423 \\

Qwen3.5-27B
& 0.875 & 0.777 & 2.367 & 2.676
& 0.867 & 0.674 & 2.279 & 2.654
& 0.783 & 0.720 & 2.888 & 3.163
& 0.817 & 0.676 & 1.966 & 2.418
& 0.825 & 0.509 & 2.087 & 2.651
& 0.817 & 0.649 & 3.167 & 3.389 \\

Qwen3.5-9B
& 0.875 & 0.753 & 2.010 & 2.333
& 0.875 & \underline{0.705} & 1.967 & 2.406
& 0.850 & 0.736 & 2.392 & 2.716
& 0.733 & 0.644 & 1.442 & 2.153
& 0.733 & 0.499 & 1.446 & 2.226
& 0.800 & 0.637 & 2.657 & 3.061 \\

\bottomrule
\end{tabular}
}
\end{table*}

\section{Use of AI Assistants}
\label{app:ai_assistant_use}

AI assistants were used to support initial chart reconstruction, annotation removal, structured annotation drafting, instruction drafting, coding, figure drafting, and language polishing. All outputs were manually checked and revised by the authors.
The LLM-based judge used in our experiments is part of the proposed evaluation method and is described in Sec.~3.3 of the main paper and Sec.~\ref{app:prompt_templates} of this supplementary document.

\section{Gallery of Representative Error Cases}
\label{app:error_case_gallery}

\end{document}